\documentclass{article} 
\usepackage{iclr2026_conference,times}

\usepackage{amsmath,amsfonts,bm}

\newcommand{\newterm}[1]{{\bf #1}}

\def\eqref#1{equation~\ref{#1}}

\def\1{\bm{1}}

\DeclareMathAlphabet{\mathsfit}{\encodingdefault}{\sfdefault}{m}{sl}
\SetMathAlphabet{\mathsfit}{bold}{\encodingdefault}{\sfdefault}{bx}{n}

\usepackage{hyperref}
\usepackage{url}
\usepackage{multirow}
\usepackage{graphicx}
\usepackage[table,xcdraw]{xcolor}
\usepackage[capitalise]{cleveref}
\usepackage{graphicx}
\usepackage{booktabs}   
\usepackage{wrapfig}
\usepackage{caption}
\usepackage{soul}
\newcommand{\mathcolorbox}[2]{\colorbox{#1}{$\displaystyle #2$}}

\definecolor{mygreen}{HTML}{009900}
\definecolor{lightblue}{HTML}{DAE8FC}
\usepackage{amsmath}
\usepackage{enumitem}
\usepackage{tabularx}

\title{AVERE: Improving Audiovisual Emotion Reasoning with Preference Optimization}

\author{Ashutosh Chaubey, Jiacheng Pang, Maksim Siniukov \& Mohammad Soleymani \\
Institute for Creative Technologies\\
University of Southern California\\
Los Angeles, CA 90007, USA \\
\texttt{achaubey@usc.edu \& soleymani@ict.usc.edu} 
}

\iclrfinalcopy 

\begin{document}
\makeatletter
\fancypagestyle{iclr_arxiv}{
  \fancyhf{} 
  \fancyhead[L]{Accepted as a conference paper at ICLR 2026}
  \renewcommand{\headrulewidth}{0.4pt}   
  \renewcommand{\headrule}{\hrule width\headwidth height\headrulewidth}
  \fancyfoot[C]{\thepage}
}
\pagestyle{iclr_arxiv}
\makeatother

\maketitle

\begin{abstract}
Emotion understanding is essential for building socially intelligent agents. Although recent multimodal large language models (MLLMs) have shown strong performance on this task, two key challenges remain: (i) spurious associations between emotions and irrelevant audiovisual cues (\emph{reasoning errors}) and (ii) hallucination of audiovisual cues (\emph{perception errors}) driven by text priors in the language model backbone. To quantify and understand these issues, we introduce \newterm{EmoReAlM}, a benchmark designed to evaluate MLLMs for cue–emotion associations, hallucinations and modality agreement. We then propose \newterm{AVEm-DPO}, a preference optimization technique that aligns model responses with both audiovisual inputs and emotion-centric queries. Specifically, we construct preferences over (i) responses exhibiting spurious associations or hallucinations and (ii) audiovisual input pairs guided by textual prompts. We also include a regularization term that penalizes reliance on text priors, thereby mitigating modality-specific cue hallucinations. Experimental results on DFEW, RAVDESS and EMER demonstrate that our method significantly improves the performance of the reference baseline models (6-19\% of relative performance) in zero-shot settings. By providing both a rigorous benchmark and a robust optimization framework, this work enables principled evaluation and improvement of MLLMs for emotion understanding and social AI. Code, models and benchmark will be released at \href{https://avere-iclr.github.io/}{avere-iclr.github.io}.
\end{abstract}    
\section{Introduction}
\label{sec:introduction}

Emotion understanding is essential for social AI agents to generate tailored responses and foster meaningful human–machine interactions \citep{chaturvedi_social,Kolomaznik2024,elyoseph2024capacity}. Emotion perception also finds applications in domains such as health \citep{Balcombe2022_hci,Litendahl2025_health} and education \citep{Salloum2025_education}, where appropriately responding to affective states can improve therapeutic alliance and learning outcomes.

Traditional multimodal emotion recognition methods \citep{dfew_mae_dfer,dfew_m3dfel,dfew_s2d} lack interpretability, as they only perform classification without grounding responses in audiovisual cues. Moreover, emotion is a complex and multi-componential construct that extends beyond the basic emotion labels that can be assigned by supervised learning methods \citep{ekman1978facial,scherer2005}. To address these challenges, recent approaches leverage multimodal large language models (MLLMs) to generate detailed emotion descriptions for interpretability \citep{emotion_llama,emotion_qwen} and to output emotion-related keywords that cover a broader spectrum of emotional states \citep{lian2024open_ov_merd_dataset,lian2025affectgpt}.

\begin{figure}
    \centering
    \includegraphics[width=1\linewidth]{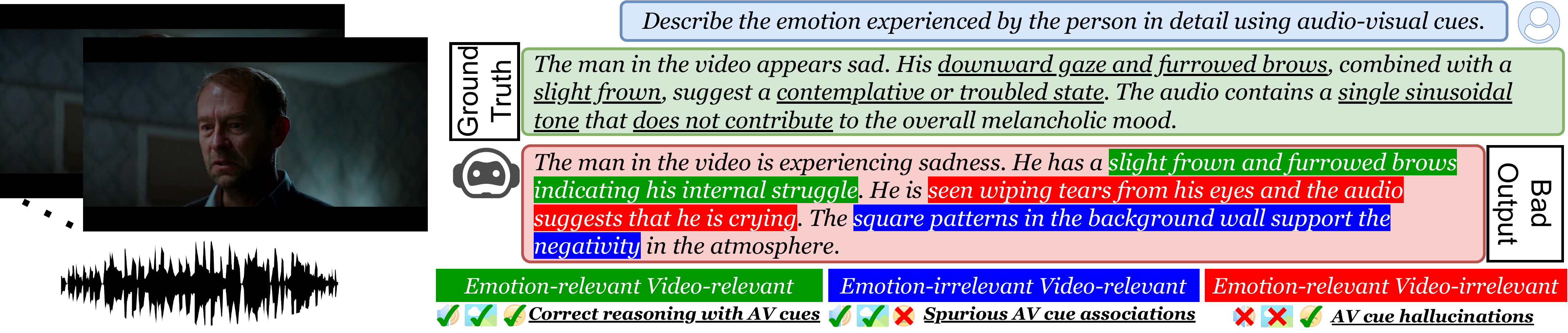}    
    \caption{Existing MLLMs (i) include spurious associations between AV cues and emotions -- \emph{reasoning errors} (\emph{blue highlight}) and (ii) hallucinate AV cues to explain emotions -- \emph{perception errors} (\emph{red highlight}). AV: audiovisual.}
    \label{fig:intro_figure_bottlenecks}
    \vspace{-1.4em}
\end{figure}

However, audiovisual MLLMs are susceptible to \emph{hallucinations}, frequently generating inaccurate or fabricated responses \citep{li2023evaluating_halluc,sahoo-etal-2024-comprehensive}. In the context of emotion understanding, they face two critical bottlenecks, as illustrated in \cref{fig:intro_figure_bottlenecks}. First, these models often ground emotion predictions on irrelevant cues (e.g., attire color, ambient noise) -- \emph{reasoning errors}. Second, they tend to hallucinate additional cues in their responses to justify emotions -- \emph{perception errors}. Such hallucinations are largely driven by text priors in the language model backbone, which bias the model to include cues that commonly co-occur with specific emotions (e.g., associating tears with the sound of crying). The scarcity of high-quality, emotion-specific instruction tuning datasets \citep{emotion_llama,lian2025affectgpt} further aggravates these issues. Addressing these challenges is essential, as they compromise the reliability of MLLM agents in social interactions and complex emotion reasoning scenarios.

\begin{figure}
    \centering
    \includegraphics[width=\linewidth]{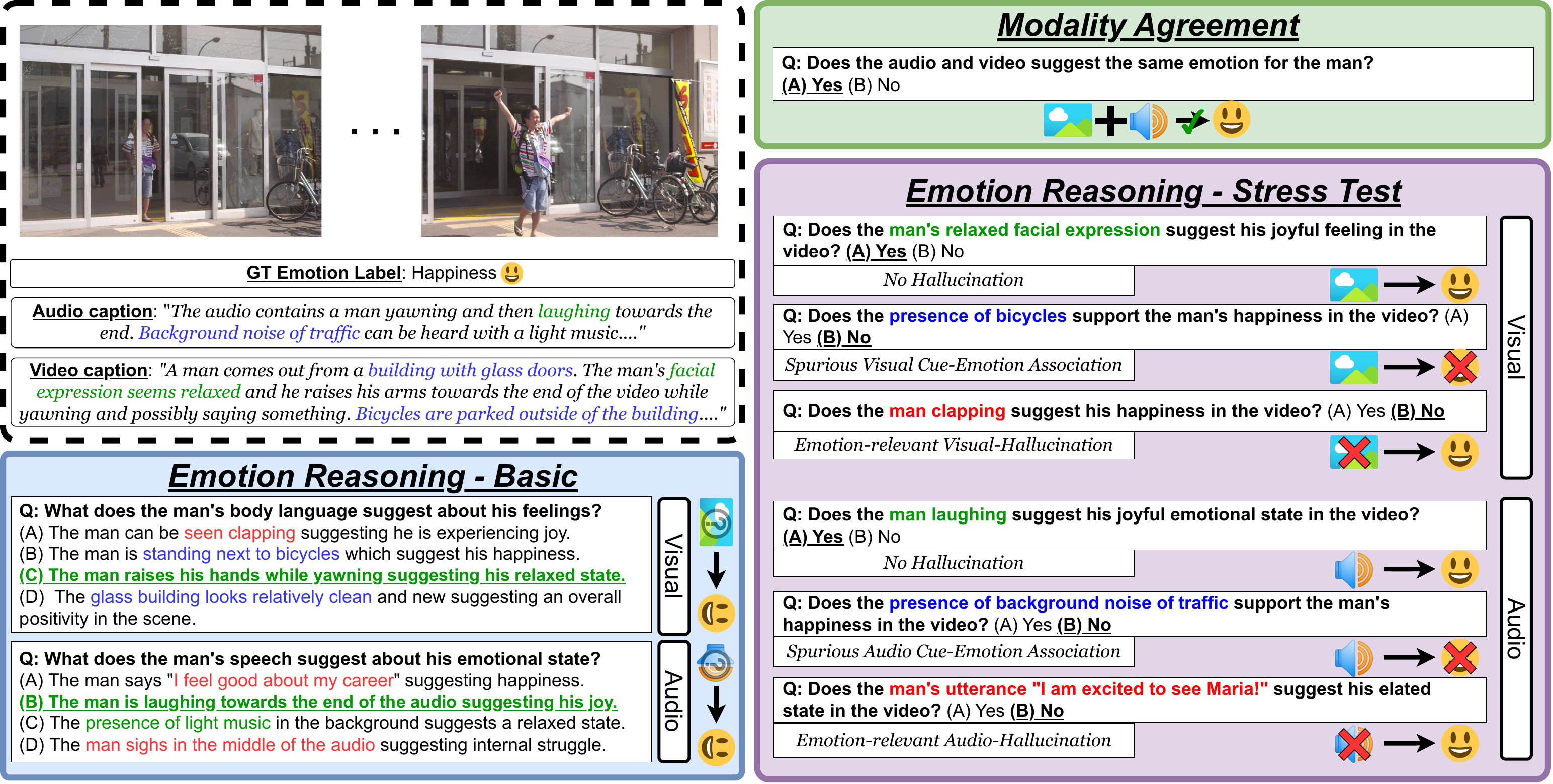}
    \vspace{-1em}
    \caption{\textbf{EmoReAlM Tasks}. In addition to basic emotion reasoning, we include tasks for \emph{Modality Agreement} and \emph{Emotion Reasoning - Stress Test} to test spurious cue-emotion associations and cue hallucinations. \textcolor{red}{Red text} is a hallucinated cue, \textcolor{blue}{blue text} is an emotion-irrelevant cue and \textcolor{mygreen}{green text} is a cue relevant for emotion understanding. Correct choices are \textbf{\underline{underlined}}.} 
    \label{fig:emo_realm_benchmark_samples}
    \vspace{-1em}
\end{figure}

Existing emotion reasoning benchmarks \citep{lian2023explainable_emer,lian2024open_ov_merd_dataset} lack the diverse and complex samples needed to fully evaluate these issues. Additionally, current audiovisual hallucination benchmarks \citep{sung-bin2025_avhbench,leng2025the_curse_of_multi_modality_cmm} predominantly focus on object-level hallucinations in audio or video, rather than on emotion-specific reasoning. Moreover, many existing MLLMs \citep{emotion_llama,lian2025affectgpt} rely on two-stage evaluation pipelines involving an external (often proprietary) LLM such as GPT \citep{openai2024gpt4ocard_gpt4o}, making replication and benchmarking difficult. To address these limitations, we introduce the \newterm{EmoReAlM} benchmark, a comprehensive suite of multiple-choice question–answer (MCQA) tasks designed to evaluate audiovisual emotion reasoning, modality agreement and hallucination-related stress tests (\cref{fig:emo_realm_benchmark_samples}). The MCQA format enables transparent, reproducible and scalable evaluation of MLLMs on emotion-centric tasks without requiring additional LLMs during inference.

Evaluation of recent MLLMs on our benchmark highlights spurious association and hallucination issues outlined in \cref{fig:intro_figure_bottlenecks}. To address these limitations, we propose \newterm{AVEm-DPO} -- a multimodal direct preference optimization (DPO) technique \citep{rafailov2023direct_dpo_orig} to enhance the emotion reasoning capabilities of MLLMs. In particular, we design explicit prompt-based audiovisual input preferences to mitigate hallucinations caused by cross-modal interactions. We also introduce text-prior debiasing, which penalizes policy reward for responses to text-only inputs. Together, these techniques significantly improve the performance of reference MLLMs, outperforming all baselines in zero-shot evaluation on both our benchmark and existing emotion recognition and reasoning datasets. 

To summarize, the main contributions of our work are:
\begin{itemize}[itemsep=0pt, topsep=0pt, leftmargin=10pt]
    \item We introduce the \newterm{EmoReAlM} benchmark with \textbf{4000 human-verified} MCQA samples to evaluate emotion reasoning and emotion-related hallucinations in MLLMs, highlighting bottlenecks such as spurious audiovisual cue associations and hallucinated cues for explaining emotions.
    \item We propose \newterm{AVEm-DPO}, a direct preference optimization technique that enforces explicit prompt-based modality preferences and reduces text-only model biases through a regularizer that penalizes over-reliance on text priors.
    \item We conduct extensive evaluations of existing MLLMs, demonstrating current bottlenecks and showing the superior performance of the proposed DPO-trained models in zero-shot settings.
\end{itemize} 
\section{Related Work}

\textbf{MLLMs for Emotion. }
While general MLLMs \citep{zhang2024llava_video,lin-etal-2024-videollava,zhang2025videollama3frontiermultimodal,xu2025qwen25omnitechnicalreport_qwen25omni,li2025baichuanomni15technicalreport_baichuamomni} show non-trivial emotion recognition ability \citep{emotion_llama}, several studies pursue domain-specific instruction tuning \citep{Xie2024EmoVIT,chaubey2025facellavafacialexpressionattribute_facellava,yang2025omniemotionextendingvideomllm}. EmotionLLaMA\citep{emotion_llama} is an audiovisual LLM for emotion recognition and captioning, finetuned on a limited dataset ($\approx$30k samples). \citet{lian2024open_ov_merd_dataset} introduces open-vocabulary emotion recognition (OV-ER), and AffectGPT \citep{lian2025affectgpt} employs a lightweight audiovisual fusion projector for OV-ER. EmotionQwen \citep{emotion_qwen} improves emotion understanding while preserving general skills via a mixture-of-experts router. \citet{han2025benchmarking_camer_mosear} use modality-specific experts with attention reallocation to handle audiovisual emotion mismatch, and \citet{wen2025listen} leverage retrieval-augmented generation with chain-of-thought for better reasoning. In contrast, we improve reasoning through multimodal preference optimization and text-prior debiasing.

Rigorous evaluation of multimodal emotion reasoning requires diverse, systematic benchmarks. \citet{lian2023explainable_emer} provide detailed descriptions of transcript, audio and visual cues for emotion reasoning, which can support GPT-based evaluation \citep{emotion_llama,han2025benchmarking_camer_mosear}. \citet{xing2025emotionhallucerevaluatingemotionhallucinations} present a holistic benchmark spanning text, image, video and audio hallucinations related to emotions. Our benchmark instead focuses squarely on audiovisual emotion understanding with a standardized pipeline and tasks beyond hallucination, including modality agreement and spurious cue–emotion associations.

\textbf{Preference Optimization. }
Direct preference optimization (DPO) \citep{rafailov2023direct_dpo_orig,liu2025tisdpo} was introduced to align LLMs to human preferences. DPO has also emerged as a leading approach for mitigating hallucinations in vision LLMs \citep{yu2024rlhf_v,wang2024_mdpo,sarkar2025mitigating_halva,huang2025mathcalvistadpo,liu2025miadpo,zhang-etal-2025-direct-hound-dpo}, but its use in audiovisual LLMs remains limited. VistaDPO \citep{huang2025mathcalvistadpo} increases video LLM robustness by building instance-level, temporal-level and object-level preferences of video inputs. \citet{sun2025videosalmonn_o1} apply process DPO for step-wise audiovisual reasoning, while \citet{tang2025video_salmonn2} use multi-round DPO for audiovisual captioning. \citet{luo2025openomni} employ DPO for emotional speech alignment to improve Omni-LLM outputs. \citet{catplus_dpo} construct multimodal preference data via ambiguity scoring, and \citet{lian2025affectgptr1leveragingreinforcementlearning} use group relative policy optimization for AffectGPT. Concurrently, Omni-DPO \citep{chen2025omnidpo} studies audiovisual modality preference. Our method differs by constructing prompt-based audiovisual preference pairs for fine-grained alignment and by introducing text-prior debiasing to reduce hallucinations in MLLMs.

\section{EmoReAlM Benchmark}
\label{sec:benchmark}
\cref{fig:emo_realm_benchmark_samples} shows different tasks present in the proposed \newterm{EmoReAlM} Benchmark.  The goal of this benchmark is to test the reasoning capabilities of MLLMs to judge the \emph{emotion experienced by the character in the given video}, specifically over the following verticals -- (i) \textbf{reasoning the correct emotion} with relevant audiovisual cues (ii) identifying whether the inferred emotion from \textbf{audio and video are in agreement} (iii) testing the \textbf{association of perceived audiovisual cues} with different emotions (\emph{reasoning errors}) and (iv) testing \textbf{audiovisual hallucination due to text-only emotion-related biases} (\emph{perception errors}). 

\subsection{Task Descriptions}
\label{subsec:task_descriptions}
\textbf{Emotion Reasoning – Basic.}
This task evaluates an MLLM’s ability to identify and reason about the emotion experienced by a person in a video by linking appropriate audio (e.g., speech transcription, tone) and visual (e.g., facial expression, body language) cues to specific emotions. To increase difficulty, the ground-truth emotion is not provided in the question. Incorrect options are constructed by modifying the correct answer to include either \textcolor{blue}{emotion-irrelevant cues present in the video} or \textcolor{red}{hallucinated cues} that falsely justify the emotion.

\textbf{Modality Agreement.}
This task assesses whether the audio and visual modalities convey the same emotional state. Unlike AVHBench \citep{sung-bin2025_avhbench}, which focuses on general cross-modal alignment, this task specifically targets agreement in emotional interpretation across modalities.

\textbf{Emotion Reasoning – Stress Test.}
MLLMs are vulnerable to both \emph{reasoning errors} and \emph{perception errors}: the former lead the model to base its responses on irrelevant audiovisual cues present in the input, while the latter cause it to rely on hallucinated cues that are not actually present. This task probes MLLMs for susceptibility to spurious cue-emotion associations (\emph{perception errors}) and hallucinated explanations driven by language model biases (\emph{reasoning errors}). Each question follows the format: \textit{“Does the \{audio/visual cue\} suggest \{emotion\} of the character?”}. For a  modality \texttt{X}, we define three sub-tasks:
(i) No Hallucination — correctly associating an audio/visual cue with the appropriate emotion.
(ii) Spurious \texttt{X} Cue-Emotion Association — linking emotion-irrelevant cues to the correct emotion.
(iii) Emotion-Relevant \texttt{X}-Hallucination — associating the correct emotion with a hallucinated cue that typically co-occurs with it.
For example, in \cref{fig:emo_realm_benchmark_samples}, a man is not clapping (per the visual caption), yet a hallucination-based question associates clapping with happiness—since clapping is commonly linked to positive emotions like joy.

\subsection{Automatic Data Creation}
\label{subsec:benchmark_creation_pipeline}
\begin{figure}
    \centering
    \includegraphics[width=\linewidth]{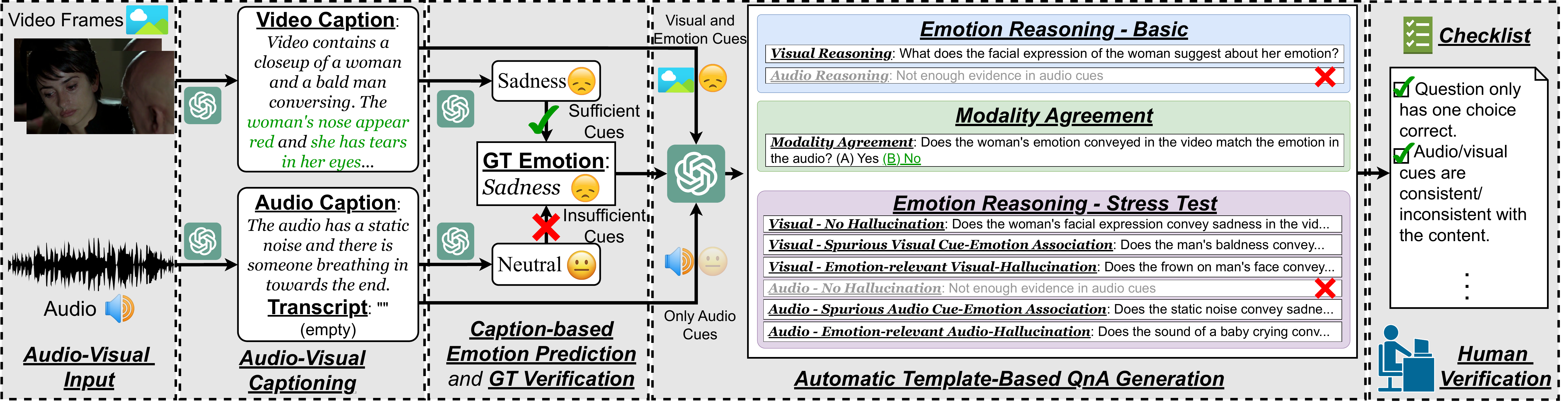}
    \vspace{-1em}
    \caption{\textbf{EmoReAlM Creation Pipeline}. We first disentangle the audiovisual information by separate captioning and verify the cues with text-based emotion prediction to find emotion-relevant cues. Finally, GPT-4o is used to generate MCQA samples that are later verified manually.}
    \label{fig:emo_realm_data_pipeline}
    \vspace{-1.4em}
\end{figure}

\cref{fig:emo_realm_data_pipeline} shows the automatic pipeline used to construct the \emph{EmoReAlM} benchmark. Our approach builds on existing manually labeled audiovisual emotion recognition datasets that provide single-word emotion annotations.
For each video, we first use an MLLM to extract detailed audio and visual captions separately, effectively disentangling the two modalities. These captions describe both emotion-relevant and irrelevant cues. To verify whether either modality reflects an emotion, we prompt an LLM to classify the audio and video captions independently into one of seven categories of neutral, in addition to six basic emotions \cite{EkmanHandbook2005}. Samples are discarded if neither caption yields a valid emotion label.
Given the validated captions and emotion label, we then generate tailored prompts and question templates for each task described in \cref{subsec:task_descriptions}. This modality-wise captioning and emotion verification process ensures the construction of high-quality, verifiable MCQA pairs that reflect meaningful audiovisual cue associations. More details and prompts are present in \cref{sec:appendix_benchmark_details}.

\textbf{Details.}
All videos are sourced from the DFEW dataset \citep{jiang2020dfew}. GPT-4o \citep{openai2024gpt4ocard_gpt4o} is used for caption extraction, emotion classification and question–answer pair generation. 

\subsection{Post-processing and Human Verification}
\label{subsec:post_processing_human_verification}
We employ GPT-4o \citep{openai2024gpt4ocard_gpt4o}, Gemini-2.5 \citep{comanici2025gemini25pushingfrontier_gemini25} and Qwen-2.5 \citep{qwen2025qwen25technicalreport_qwen25} to predict the correct answer to the generated questions just by using question text as input. We remove all the QA pairs for which all the models identified the correct answer just with the text information. Finally, since the QA samples are generated automatically leveraging MLLMs, which can hallucinate themselves, we perform a human verification over the samples generated by recruiting over 470 participants using the crowd-sourcing platform \cite{prolificProlificEasily}. Details are present in \cref{subsec:appendix_human_verification_benchmark}.

\subsection{Benchmark statistics}
\begin{wraptable}[8]{r}{0.45\linewidth}
\centering
\vspace{-\baselineskip} 
\caption{\emph{EmoReAlM} Benchmark Statistics.}
\label{tab:emo_realm_benchmark_statistics}
\vspace{-1em}
\resizebox{\linewidth}{!}{%
\begin{tabular}{ll|cc|c}
\hline\hline
\rowcolor[HTML]{C0C0C0} 
\rowcolor[HTML]{C0C0C0} 
\multicolumn{2}{c|}{\multirow{-1}{*}{\cellcolor[HTML]{C0C0C0}\textbf{Task}}} & \cellcolor[HTML]{C0C0C0}\textbf{\# QA} & \cellcolor[HTML]{C0C0C0}\textbf{\# vid.} & \multirow{-1}{*}{\cellcolor[HTML]{C0C0C0}\textbf{\begin{tabular}[c]{@{}c@{}}Rand. Acc.\end{tabular}}} \\ \hline
 & Audio  & 972 & 784 & 25\% \\
\multirow{-2}{*}{Reasoning Basic} & Visual & 1024 & 883 & 25\% \\
\multicolumn{2}{l|}{Modality Agreement} & 456 & 456 & 50\% \\
 & Audio & 820 & 655 & 50\% \\
\multirow{-2}{*}{Reas. Stress Test} & Visual & 728 & 593 & 50\% \\ \hline
\multicolumn{2}{c|}{\textbf{Total}} & \textbf{4000} & \textbf{2649} & \textbf{} \\\hline\hline
\end{tabular}%
}
\end{wraptable}
\cref{tab:emo_realm_benchmark_statistics} summarizes the data statistics of the proposed \emph{EmoReAlM} Benchmark, which comprises a total of \textbf{4,000 questions} over \textbf{2,649 unique videos}. Samples from the benchmark are present in \cref{subsec:appendix_benchmark_samples}. Importantly, for tasks which always have a fixed set of answer choices (\emph{Emotion Reasoning - Stress Test} and \emph{Modality Agreement} -- \emph{Yes/No}), we ensure that there is a uniform distribution of correct answer texts over the possible answer choice texts. Additionally, we ensure that the distribution of emotion labels over the videos in the benchmark matches the video source dataset (refer to \cref{subsec:appendix_benchmark_statistics} for details). It is also important to note that \emph{EmoReAlM} is only used as a \textbf{test set} to evaluate the reasoning capabilities of MLLMs, and we use a different dataset for preference optimization (refer \cref{subsec:preference_data}).
\section{AVEm-DPO}


Direct preference optimization (DPO) \citep{rafailov2023direct_dpo_orig} aligns LLMs to human preferences, bypassing the need to develop a reward model. In the context of audiovisual LLMs, given a reference model $\pi_{\text{ref}}$, we can reformulate the DPO objective to learn an optimal policy $\pi_\theta$ as the following,
\begin{equation}
\small
\max_{\pi_\theta} \mathbb{E}_{(a,v,x)\sim\mathcal{D},y\sim \pi_\theta(\cdot\mid a,v,x)}\left[
r(a,v,x,y)\right] - \beta \mathbb{D}_{\text{KL}}(\pi_\theta(\cdot \mid a,v,x) \parallel \pi_{\text{ref}}(\cdot \mid a,v,x))
\label{eq:dpo_orig}
\end{equation}
where $(a,v)$ is audiovisual input, $x$ is text prompt, $y$ is text response and $r(a,v,x,y)$ is the reward function for given input-output pair. Optimizing \cref{eq:dpo_orig} to find optimal policy results in the following reward formulation,
\begin{equation}
\small
r(a,v,x,y) = \beta \log \frac{\pi_\theta(y\mid a,v,x)}{\pi_{\text{ref}}(y\mid a,v,x)} + \beta \log Z(a,v,x)
\label{eq:dpo_optimal_reward}
\end{equation}
where $Z(\cdot)$ is the partition function derived in \cite{rafailov2023direct_dpo_orig}. With access to a preference dataset $\mathcal{D}^{\text{pref}}_y$ with samples $(a,v,x,y_w,y_l)$ and using the Bradley-Terry preference model \citep{Bradley1952RankAO_bradleyterry} to model preference of chosen response ($y_w$) over rejected response ($y_l$), the final DPO objective becomes
\begin{equation}
\small
\mathcal{L}_{\text{DPO}}=-\mathbb{E}_{(a,v,x,y_w,y_l)\sim \mathcal{D}^{\text{pref}}}\left[\log \sigma\left(\beta \log \frac{\pi_\theta(y_w \mid a,v,x)}{\pi_{\text{ref}}(y_w \mid a,v,x)} - \beta \log \frac{\pi_\theta(y_l \mid a,v,x)}{\pi_{\text{ref}}(y_l \mid a,v,x)}\right)\right]
\label{eq:naive_dpo_objective}
\end{equation}

\begin{figure}
    \centering
    \includegraphics[width=\linewidth]{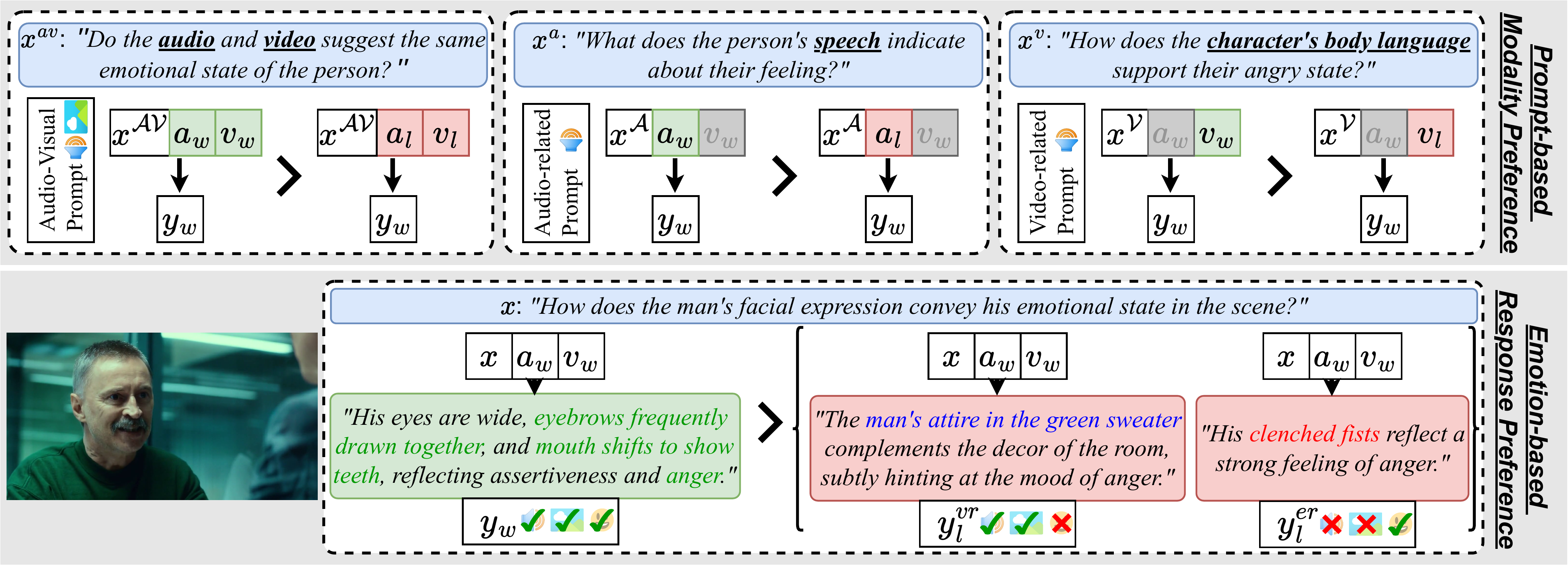}
    \caption{\textbf{Preference pairs in AVEm-DPO}. \textit{(Top)} Fine-grained preference over modality input based on current prompt. \textit{(Bottom)} Each chosen response $y_w$ has two rejected responses -- $y_l^{vr}$ relevant to the video but with spurious emotion association and $y_l^{er}$ irrelevant to the video (hallucinated) but related to the emotion.}
    \label{fig:avem_dpo_preference}
    \vspace{-1.4em}
\end{figure}

\subsection{Multimodal Preference Optimization}
\label{subsec:multimodal_preference}
Naive DPO (\cref{eq:naive_dpo_objective}) applied to MLLMs, when relying only on response preference, often causes the policy model to overfit to the input prompt $x$ while neglecting the multimodal inputs during alignment \citep{wang2024_mdpo,sarkar2025mitigating_halva}. To address this limitation, preference optimization can be extended to incorporate audiovisual inputs as follows: 
\begin{equation}
\small
\begin{aligned}
\mathcal{L}^{av}_{\text{DPO}} &= -\mathbb{E}\big[\log \sigma (u(a_w,v_w,a_l,v_l,x,y_w))\big],\text{ }
u(\cdot) = 
\beta \log \frac{\pi_\theta(y_w \mid a_w,v_w,x)}{\pi_{\text{ref}}(y_w \mid a_w,v_w,x)} 
- \beta \log \frac{\pi_\theta(y_w \mid a_l,v_l,x)}{\pi_{\text{ref}}(y_w \mid a_l,v_l,x)}
\end{aligned}
\label{eq:dpo_av_preference}
\end{equation}
where $(a_w,v_w)$ and $(a_l,v_l)$ denote the chosen and rejected multimodal inputs. This objective ensures that the policy model aligns its response $y_w$ to the correct (chosen) audiovisual input $(a_w,v_w)$.

\textbf{Prompt-based Modality Preference (PMP).} While \cref{eq:dpo_av_preference} enforces preference over $non$-$text$ inputs, in the case of audiovisual (or \emph{``omni"}) LLMs the input prompt $x^{m}$ may relate to both audio and visual modalities, or to only one of them ($m \in \mathcal{M}=\{\mathcal{AV},\mathcal{A},\mathcal{V}\}$). This often leads to cross-modality-induced hallucinations in MLLMs \citep{sung-bin2025_avhbench}, where a response to a prompt concerning one modality $x^{m_1}$ is spuriously influenced by another modality $m_2\in\mathcal{M}\setminus\{m_1\}$.

To mitigate this issue, we construct the preference dataset $\mathcal{D}^{\text{pref}}_{av}$ with fine-grained modality-level preferences conditioned on the input prompt $x^m$, as illustrated in \cref{fig:avem_dpo_preference} \textit{(Top)}. For example, for a query specific to one modality $x^m$ (e.g., visual: \emph{``How does the character's body language support their angry state?"}), we modify only the corresponding input(s) of modality $m$ (i.e. visual) in the rejected pair, thereby enforcing that the model’s response remains grounded in that modality. Thus, our prompt-based modality preference objective becomes,
\begin{equation}
\label{eq:dpo_av_preference_prompt_based}
\mathcal{L}^{av-prompt}_{\text{DPO}} = -\mathbb{E}\big[\log \sigma (u(a_w,v_w,\mathcolorbox{lightblue}{a_l^{\text{PMP}},v_l^{\text{PMP}}},x^m,y_w))\big]
\end{equation}
where $a_l^{\text{PMP}} = {a_w},\text{ iff } m=\mathcal{V}$ and $v_l^{\text{PMP}} = {v_w},\text{ iff } m=\mathcal{A}$. We perform multiple forms of negative sampling for constructing $(a_l,v_l)$ (see \cref{subsec:results_analysis}); however, because our task is emotion reasoning, the best results were achieved when we choose the rejected audiovisual input to be a sample with an emotion different from the chosen input $(a_w,v_w)$.

\textbf{Emotion-based Response Preference.} 
To mitigate spurious cue-emotion associations and hallucinations described in \cref{sec:introduction}, for a given input $(a_w,v_w,x)$ we construct two rejected responses that are variations of the chosen response $y_w$, as illustrated in \cref{fig:avem_dpo_preference}\textit{(Bottom)}. Specifically, $y^{vr}_l$ includes an audio/visual cue that is relevant to the audiovisual input but does not explain the emotion, whereas $y_l^{er}$ introduces audio/visual cues related to the emotion but absent from the audiovisual input (hallucinated). Following \cite{huang2025mathcalvistadpo}, we assign weights to these rejected responses in the DPO loss in \cref{eq:naive_dpo_objective} as,
\begin{equation} 
\tiny 
\mathcal{L}^y_{\text{DPO}}=-\mathbb{E}_{(a_w,v_w,x,y_w,y_l^{vr},y_l^{er})\sim \mathcal{D}^{\text{pref}}_y}\left[\log \sigma\left[\beta\left( \log \frac{\pi_\theta(y_w \mid a_w,v_w,x)}{\pi_{\text{ref}}(y_w \mid a_w,v_w,x)} - \mathcolorbox{lightblue}{\sum_{i \in \{vr,er\}} \beta_{i} \log \frac{\pi_\theta(y_l^{i} \mid a_w,v_w,x)}{\pi_{\text{ref}}(y_l^{i} \mid a_w,v_w,x)}}\right)\right]\right]
\label{eq:dpo_response_preference}
\end{equation}
where $\beta_{er}+\beta_{vr}=1$. This formulation establishes strong contrasts between chosen and rejected responses, encouraging the policy model to ground its outputs in correct and emotion-relevant audiovisual cues. Unlike \cite{huang2025mathcalvistadpo}, however, we do not include completely irrelevant responses as rejections in DPO based on empirical findings in \cref{subsec:appendix_response_preference_ablation}.

\subsection{Text Prior Debiasing (TPD)}
\label{subsec:text_prior_debiasing}

Audiovisual LLMs have strong text priors that cause them to hallucinate and include cues in their response, which usually occur together (e.g., the presence of a crying person accompanied by the sound of crying). To suppress such behaviour, we propose to penalize the reward $r(a,v,x,y)$ derived in \cref{eq:dpo_optimal_reward} to generate the response using only text input as follows,
\begin{equation}
r(a,v,x,y) = \beta \log \frac{\pi_\theta(y\mid a,v,x)}{\pi_{\text{ref}}(y\mid a,v,x)} + \beta \log Z(a,v,x) - \mathcolorbox{lightblue}{\gamma_{\text{TPD}} \log \pi_{\text{text}}(y\mid x)}
\label{eq:dpo_tpd_reward}
\end{equation}
where $\pi_{\text{text}}$ is a trained language model and $\gamma_{\text{TPD}}$ is a hyperparameter. In our experiments, we choose $\pi_{\text{text}}$ to be the language model backbone in $\pi_{\text{ref}}$. This penalty ensures that the responses that are explainable purely by text priors get discounted and responses supported by audio/video get relative credit. Plugging \cref{eq:dpo_tpd_reward} in the Bradley Terry model results in the following objective, 
\begin{equation}
\tiny
\begin{aligned}
\mathcal{L}_{\text{DPO-TPD}}
&= -\mathbb{E}_{(a,v,x,y_w,y_l)\sim \mathcal{D}^{\text{pref}}}\Bigg[
    \log \sigma\Bigg(
        \beta \Bigg(
            \log \frac{\pi_\theta(y_w \mid (a,v,x))}
                      {\pi_{\text{ref}}(y_w \mid (a,v,x))}
            - 
            \log \frac{\pi_\theta(y_l \mid (a,v,x))}
                      {\pi_{\text{ref}}(y_l \mid (a,v,x))}
        \Bigg) \\
&\hspace{3cm}
        - \mathcolorbox{lightblue}{\gamma_{\text{TPD}}
        \Big(
            \log \pi_{\text{text}}(y_w \mid x)
            - \log \pi_{\text{text}}(y_l \mid x)
        \Big)}
    \Bigg)
\Bigg]
\label{eq:dpo_tpd_objective}
\end{aligned}
\end{equation}
where $(a,v)$ denote $(a_w,v_w)$ for simplicity. During training, we stop gradients through $\pi_{\text{text}}$ as it is just used to identify the text priors that a language model has. To maintain the text-only capabilities of the language model backbone, we attach LoRA module \citep{hu2022lora} to it for training.
To accommodate two rejected responses, we perform scaling similar to \cref{eq:dpo_response_preference} on the rejected responses in the TPD term as described in \cref{subsec:appenix_equations_tpd} (\cref{eq:dpo_tpd_objective}) to get the final TPD objective $\mathcal{L}^{y}_{\text{DPO-TPD}}$. The final objective function of \newterm{AVEm-DPO} is as follows,
\begin{equation}
\small
\label{eq:final_avemdpo_objective}
\mathcal{L}_{\text{AVEm-DPO}} = \mathcal{L}^{y}_{\text{DPO-TPD}} + \lambda_{av} \mathcal{L}^{av-prompt}_{\text{DPO}}
\end{equation}
where $\lambda_{av}$ is a hyperparameter. Implementation details are present in \cref{subsec:appendix_implementation_details}.

\subsection{Preference Data}
\label{subsec:preference_data}
For AVEm-DPO training, we construct preference data using a pipeline similar to \cref{fig:emo_realm_data_pipeline}. This preference dataset is different from \emph{EmoReAlM}, which we exclusively use for testing. We use MAFW \citep{liu_mafw_2022} and a subset of MER2025 \citep{lian2025mer} \emph{Track-1 train set} as the source datasets to create preference samples. We prompt Gemini-2.5 \cite{comanici2025gemini25pushingfrontier_gemini25} to generate variations of the correct answers (chosen responses) to the questions where the audiovisual cue is altered to be either a spurious emotion-related video-relevant cue ($y_l^{vr}$) or a hallucinated cue related to the emotion present ($y_l^{er}$). Note that we do not perform any manual verification on the generated data, which still results in a performance gain demonstrating the efficiency of the proposed approach. Details in \cref{subsec:appendix_preference_data}.
\section{Experiments}
\label{subsec:experimental_setup}
\textbf{Datasets \& Metrics.} For EmoReAlM benchmark, we report the average accuracy per task for all the tasks. For tasks with \emph{Yes/No} responses, we additionally report the precision, recall and F1 score following previous multimodal hallucination benchmarks \citep{sung-bin2025_avhbench,li2023evaluating_halluc}. 
\begin{table}[t]
\centering
\caption{Zero-shot performance comparison of different methods on existing audiovisual emotion recognition benchmarks. Mod. are the modalities input to the model with the prompt. A: Audio, V:Video, T: Text Subtitles. $\ddagger$: evaluation without text subtitle input.}
\label{tab:av_emotion_results}
\vspace{-1em}
\resizebox{\textwidth}{!}{%
\begin{tabular}{l|c|cc|cc|c|cccc}
\hline \hline
\rowcolor[HTML]{C0C0C0} 
\multicolumn{1}{c|}{\cellcolor[HTML]{C0C0C0}} & \multicolumn{1}{c|}{\cellcolor[HTML]{C0C0C0}} & \multicolumn{2}{c|}{\cellcolor[HTML]{C0C0C0}\textbf{DFEW}} & \multicolumn{2}{c|}{\cellcolor[HTML]{C0C0C0}\textbf{RAVDESS}} & \multicolumn{1}{c|}{\cellcolor[HTML]{C0C0C0}\textbf{MER2023}} & \multicolumn{4}{c}{\cellcolor[HTML]{C0C0C0}\textbf{EMER}} \\ \cline{3-11} 
\rowcolor[HTML]{C0C0C0} 
\multicolumn{1}{c|}{\multirow{-2}{*}{\cellcolor[HTML]{C0C0C0}\textbf{Model}}} & \multicolumn{1}{c|}{\multirow{-2}{*}{\cellcolor[HTML]{C0C0C0}\textbf{Mod.}}} & \multicolumn{1}{c}{\cellcolor[HTML]{C0C0C0}\textbf{UAR}} & \multicolumn{1}{c|}{\cellcolor[HTML]{C0C0C0}\textbf{WAR}} & \multicolumn{1}{c}{\cellcolor[HTML]{C0C0C0}\textbf{UAR}} & \multicolumn{1}{c|}{\cellcolor[HTML]{C0C0C0}\textbf{WAR}} & \multicolumn{1}{c|}{\cellcolor[HTML]{C0C0C0}\textbf{F1}} & \multicolumn{1}{c}{\cellcolor[HTML]{C0C0C0}\textbf{Clue}} & \multicolumn{1}{c}{\cellcolor[HTML]{C0C0C0}\textbf{Label}} & \multicolumn{1}{c}{\cellcolor[HTML]{C0C0C0}\textbf{Spurious}} & \multicolumn{1}{c}{\cellcolor[HTML]{C0C0C0}\textbf{Halluc.}} \\ \hline
VideoLLaMA 2 & A,V & 43.65 & 48.66 & 41.81 & 31.62 & 50.79 & 3.82 & 3.80 & 4.25 & 4.23  \\
OLA & A,V & 38.17 & 41.73 & 27.45 & 22.11 & 55.82 & 3.80 & 3.33 & 3.93 & 4.22 \\
VITA-1.5 & A,V & 39.31 & 42.56 & 50.67 & 46.88 & 66.94 & 4.77 & 4.72 & 5.16 & 5.70             \\
Qwen-2.5 Omni & A,V & 46.94 & 54.34 & 32.88 & 28.05 & 79.72 & 5.85 & 6.78 & 6.39 & 6.21 \\ 
EmotionLLaMA & A,V,T & 45.59 & 59.37 & 28.20 & 29.24 & 90.36 & 6.03 & 6.99 & 5.89 & 5.26 \\ 
EmotionLLaMA$^\ddagger$ & A,V & 42.72 & 54.06 & 30.36 & 30.45 & 89.05 & 2.76 & 2.78 & 3.44 & 2.36  \\ 
MoSEAR & A,V,T & 44.48 & 56.60 & - & - & 90.27 & - & - & - & - \\ \hline
\textbf{Our base} & A,V & 56.78 & 60.14 & 53.59 & 53.01 & 89.19 & 5.63 & 6.45 & 5.41 & 5.19  \\
+ Naive-DPO &  & 55.67 & 59.90 & 53.63 & 52.94 & 88.59 & 5.81 & 6.30 & 5.96 & 5.48  \\
+ Vista-DPO\textsuperscript{\textdagger} &  & 56.42 & 62.33 & \underline{56.94} & \underline{53.64} & 90.06 & \underline{6.08} & 6.89 & 6.58 & 6.07  \\
\rowcolor[HTML]{DAE8FC}
+ \textbf{AVEm-DPO} &  & \textbf{58.54} & \textbf{64.24} & \textbf{58.66} & \textbf{55.48} & \textbf{92.18} & \textbf{6.37} & \textbf{7.08} & \textbf{7.09} & \textbf{6.75}  \\ \hline
\textbf{EmotionLLaMA$^\star$}  & A,V & 54.89 & 58.26 & 52.59 & 48.12 & 90.01 & 5.78 & 6.21 & 5.36 & 5.23 \\
+ Naive-DPO &  & 54.97 & 58.12 & 52.69 & 49.01 & 89.35 & 5.89 & 6.35 & 5.89 & 5.62 \\
+ Vista-DPO\textsuperscript{\textdagger} &  & 56.28 & 61.58 & 56.42 & 50.96 & 91.19 & 6.05 & 6.56 & 6.85 & 6.31 \\
\rowcolor[HTML]{DAE8FC}
+ \textbf{AVEm-DPO} &  & \underline{57.06} & \underline{62.12} & 56.21 & 51.03 & \underline{91.68} & 6.02 & \underline{6.99} & \underline{7.02} & \underline{6.62} \\ \hline \hline
\end{tabular}%
}
\vspace{-0.3em}
\end{table}
Beyond \emph{EmoReAlM}, we also evaluate on established emotion recognition datasets—DFEW \citep{jiang2020dfew}, RAVDESS \citep{Livingstone2018_ravdess}, MER2023 \citep{lian2023mer2023}—and the emotion reasoning dataset EMER \citep{lian2023explainable_emer}. None of these datasets is used in training to ensure zero-shot evaluation. Following prior work \citep{emotion_llama,han2025benchmarking_camer_mosear}, we report unweighted and weighted average recalls for DFEW and RAVDESS and weighted F1 for MER2023. For emotion reasoning, we adopt GPT-based evaluation \citep{emotion_llama}, comparing generated responses against ground truth. In addition to clue and label overlap, we assess two dimensions: (i) \emph{spurious cue–emotion associations}, where irrelevant cues are linked to emotions, and (ii) \emph{hallucinatory cues}, where non-existent audiovisual cues are fabricated. For all metrics, higher values indicate better performance. Further details are provided in \cref{subsec:appendix_eval_metrics}.

\textbf{Reference models.} We use two audiovisual MLLMs as reference -- EmotionLLaMA \citep{emotion_llama} and our own developed base model. Our model is similar to EmotionLLaMA in architecture with changes to the audio encoder (\emph{whisper-large-v3}\citep{radford2023robust_whisper}) and video encoder (\emph{LanguageBind} \citep{zhu2024languagebind}). For EmotionLLaMA, we remove the text (subtitle) input branch to be consistent with the other baselines and retrain the model on the original dataset without subtitles -- denoted as \newterm{EmotionLLaMA$^\star$} \citep{emotion_llama}. More details in \cref{subsec:appendix_reference_models}. 

\textbf{Baseline Preference Optimization Approaches.} We compare with original \newterm{Naive-DPO} \citep{rafailov2023direct_dpo_orig} using single rejected samples from our DPO data and modified Vista-DPO \citep{huang2025mathcalvistadpo} for audiovisual inputs -- denoted as \newterm{Vista-DPO\textsuperscript{\textdagger}} (\cref{subsec:appendix_baseline_preference_optimization_methods} for details). 

\begin{wraptable}[15]{r}{0.60\linewidth}
\vspace{-2em} 
\centering
\caption{Performance comparison of different methods on the proposed \emph{EmoReAlM} Benchmark. }
\vspace{-1em}
\resizebox{\linewidth}{!}{%
\begin{tabular}{l|c|c|c|c|c}
\hline \hline
\rowcolor[HTML]{C0C0C0} 
\multicolumn{1}{c|}{\cellcolor[HTML]{C0C0C0}} & \multicolumn{2}{c|}{\cellcolor[HTML]{C0C0C0}\textbf{Reas. Basic}} & \multicolumn{1}{c|}{\cellcolor[HTML]{C0C0C0}} & \multicolumn{2}{c}{\cellcolor[HTML]{C0C0C0}\textbf{Reas. - Stress}} \\ \cline{2-3} \cline{5-6}
\rowcolor[HTML]{C0C0C0} 
\multicolumn{1}{c|}{\cellcolor[HTML]{C0C0C0}} & \multicolumn{1}{c|}{\cellcolor[HTML]{C0C0C0}\textbf{Audio}} & \multicolumn{1}{c|}{\cellcolor[HTML]{C0C0C0}\textbf{Visual}} & \multicolumn{1}{c|}{\multirow{-2}{*}{\cellcolor[HTML]{C0C0C0}\textbf{\begin{tabular}[c]{@{}c@{}}Modality \\ Agree.\end{tabular}}}} & \multicolumn{1}{c|}{\cellcolor[HTML]{C0C0C0}\textbf{Audio}} & \multicolumn{1}{c}{\cellcolor[HTML]{C0C0C0}\textbf{Visual}} \\ \cline{2-6}
\rowcolor[HTML]{C0C0C0} 
\multicolumn{1}{c|}{\multirow{-3}{*}{\cellcolor[HTML]{C0C0C0}\textbf{Model}}} & \multicolumn{1}{c|}{\cellcolor[HTML]{C0C0C0}\textbf{Acc.}} & \multicolumn{1}{c|}{\cellcolor[HTML]{C0C0C0}\textbf{Acc.}} & \multicolumn{1}{c|}{\cellcolor[HTML]{C0C0C0}\textbf{F1}} & \multicolumn{1}{c|}{\cellcolor[HTML]{C0C0C0}\textbf{F1}} & \multicolumn{1}{c}{\cellcolor[HTML]{C0C0C0}\textbf{F1}} \\ \hline
VideoLLaMA2 & 63.1 & 66.8 & 52.5 & 53.2 & 58.4 \\
OLA & 63.2 & 60.4 & 42.7 & 56.6 & 54.8 \\
VITA-1.5 & 63.1 & 84.3 & 30.2 & 52.8 & 56.3 \\
Qwen 2.5 Omni & 76.8 & 89.2 & 33.3 & 55.0 & 56.8 \\  \hline
\textbf{Our base} & 69.2 & 85.3 & 34.6 & 50.3 & 59.9 \\
+ Naive-DPO & 71.3 & 85.9 & 41.6 & 54.8 & 65.9  \\
+ Vista-DPO\textsuperscript{\textdagger} & 72.4 & 87.8 & 52.1 & 73.6 & 86.7 \\
\rowcolor[HTML]{DAE8FC}
+ \textbf{AVEm-DPO} & \textbf{77.9} & \textbf{92.5} & \textbf{60.0} & \textbf{80.9} &  \textbf{94.6} \\ \hline
\textbf{Emot.-LLaMA$^\star$} & 64.8 & 84.9 & 33.1 & 46.7 & 63.2 \\
+ Naive-DPO & 67.2 & 85.7 & 42.8 & 52.6& 67.6  \\
+ Vista-DPO\textsuperscript{\textdagger} & 69.0 & 86.9 & 40.9 & 68.6 & 87.3 \\
\rowcolor[HTML]{DAE8FC}
+ \textbf{AVEm-DPO} & \underline{76.5} & \underline{89.9} & \underline{56.8} & \underline{75.4} & \underline{91.7} \\ \hline \hline
\end{tabular}%
}
\label{tab:emo_realm_main_results}
\end{wraptable}
\subsection{Emotion Reasoning and Recognition Results}
\textbf{EmoReAlM Results.} \cref{tab:emo_realm_main_results} presents the performance of different approaches on the proposed \emph{EmoReAlM} benchmark. AVEm-DPO achieves substantial gains over the reference models, demonstrating the effectiveness of multimodal preference optimization and text-prior debiasing. While the baselines perform strongly on basic reasoning tasks, \cref{tab:emo_realm_main_results} shows that they struggle on \emph{Modality Agreement} and \emph{Stress-Test} evaluations (Expanded table in \cref{subsec:appendix_emo_realm_results_detailed_expanded,tab:emo_realm_main_results_detailed}). 

Notably, our preference optimization also surpasses Vista-DPO and Naive-DPO by significant margins. To further examine the bottlenecks in baseline models, \cref{subsec:appendix_emo_realm_stress_detailed} reports results on samples probing spurious audiovisual–emotion correlations and hallucinated cues. For state-of-the-art systems such as Qwen 2.5 Omni \citep{xu2025qwen25omnitechnicalreport_qwen25omni} and VITA-1.5 \cite{fu2025vita}, hallucination emerges as a more severe issue than spurious cue-emotion associations. Moreover, unlike findings from \cite{sung-bin2025_avhbench}, our results indicate that audio and visual hallucinations are equally prevalent in emotion reasoning tasks. Additionally. \cref{tab:emo_realm_main_results_detailed} shows the performance of video-only and audio-only baselines and reveals that multimodal inputs hurt reasoning capabilities.

\begin{wraptable}[10]{r}{0.5\linewidth}
\vspace{-1em}
\centering
\caption{User evaluation on EMER.}
\vspace{-1em}
\label{tab:user_evaluation_emer}
\resizebox{\linewidth}{!}{%
\begin{tabular}{lccc}
\hline\hline
\rowcolor[HTML]{C0C0C0} 
\multicolumn{1}{c}{\cellcolor[HTML]{C0C0C0}\textbf{Model}} & \textbf{Emot.$\uparrow$} & \textbf{Assoc.$\uparrow$} & \textbf{Incons.$\downarrow$} \\ \hline
VideoLLaMA 2 & 9.82\% & 0.75\% & 15.38\% \\
OLA & 9.36\% & 7.46\% & 5.58\% \\
VITA 1.5 & 11.60\% & 17.25\% & 6.04\% \\
Qwen 2.5 Omni & 10.75\% & 18.57\% & 10.13\% \\
EmotionLLaMA & 1.89\% & 11.53\% & 68.61\% \\
\rowcolor[HTML]{DAE8FC}
Our + AVEm-DPO & 54.74\% & 43.35\% & 4.67\% \\ \hline\hline
\end{tabular}%
}
\end{wraptable}
\textbf{Emotion Recognition and Reasoning on Existing Benchmarks.} \cref{tab:av_emotion_results} (expanded in \cref{subsec:appendix_emotion_recognition_results_expanded}) shows the performance on existing emotion benchmarks mentioned before. We can notice that our reference models outperform baselines, showing the efficacy of reference in understanding emotion. Moreover, preference tuning additionally boosts the performance, especially for emotion reasoning on EMER, reducing spurious cue-emotion associations and hallucinations. It is important to note that previous emotion MLLM baselines \citep{emotion_llama,han2025benchmarking_camer_mosear} use text subtitle as additional input. Qualitative comparison to baselines is present in \cref{sec:appendix_qualitative_samples}. While most baselines perform poorly on the out-of-domain RAVDESS dataset, our reference and preference-tuned models perform significantly better, showing their generalizability.

\textbf{User evaluation.} We perform a user evaluation with 40 participants on EMER generations from different models and report results in \cref{tab:user_evaluation_emer}. Participants chose our model the most for emotion description and emotion-cue associations and the least for inconsistencies. (Details in \cref{subsec:appendix_user_evaluation}).

\subsection{Analysis}
\label{subsec:results_analysis}
\begin{wraptable}[12]{r}{0.5\linewidth}
\centering
\vspace{-1em}
\caption{Ablation study over different components of the proposed AVEm-DPO approach. PMP: Prompt-based Modality Preference, ERP: Emotion-based Response Preference, TPD: Text Prior Debiasing. }
\vspace{-1em}
\label{tab:ablation_study}
\resizebox{\linewidth}{!}{%
\begin{tabular}{l|ccc|cc}
\hline\hline
\rowcolor[HTML]{C0C0C0} 
\multicolumn{1}{c|}{\cellcolor[HTML]{C0C0C0}\textbf{Method}} & \textbf{\begin{tabular}[c]{@{}c@{}}Basic.\end{tabular}} & \textbf{\begin{tabular}[c]{@{}c@{}}Agree.\end{tabular}} & \textbf{\begin{tabular}[c]{@{}c@{}}Stress\end{tabular}} &  \textbf{\begin{tabular}[c]{@{}l@{}}Spur.\end{tabular}} & \textbf{\begin{tabular}[c]{@{}l@{}}Hall.\end{tabular}} \\ \hline
Our base & 77.3 & 34.6 & 55.1 & 47.3 & 39.2 \\ \hline
\rowcolor[HTML]{DAE8FC}
+ AVEm-DPO & 85.2 & 60.1 & 87.8 & 92.7 & 97.6 \\
w/o PMP & 81.0 & 54.9 & 79.6 & 86.2 & 88.1 \\
w/o ERP & 81.8 & 56.2 & 79.4 & 84.9 & 88.4 \\
w/o TPD & 83.8 & 58.9 & 78.8 & 87.1 & 77.8 \\ \hline
+ Contr. Dec. & 79.1 & 51.3 & 61.7 & 50.9 & 54.8 \\ \hline\hline
\end{tabular}%
}
\end{wraptable}
\textbf{Ablation Study.} \cref{tab:ablation_study} shows the performance of the preference-tuned model after removing the proposed components of AVEm-DPO. We perform this analysis on \emph{EmoReAlM} and report the average metrics over audio and visual reasoning (\cref{subsec:appendix_ablatlion_detailed_setup} for details). Removal of any of the key components results in a significant performance drop, especially for the reasoning tasks. Moreover, ablating TPD results in a huge performance drop on the hallucination stress test samples, underlining its efficacy in eliminating cue hallucinations in audiovisual emotion reasoning. 

\textbf{Comparison with training-free contrastive decoding.} Similar to VCD \cite{visual_contrastive_decoding_vcd}, we perform contrastive decoding using diffused audiovisual inputs and report results in \cref{tab:ablation_study} (\emph{last row}), showcasing it is significantly worse than AVEm-DPO.

\textbf{Design Choices and Sensitivity to Hyperparams.} \cref{subsec:appendix_modality_preference_ablation} shows that prompt-based modality preference using a different emotion audiovisual (AV) input as $(a_l,v_l)$ works better compared to using random videos or diffused versions of the inputs. \cref{subsec:appendix_response_preference_ablation} shows that using emotion-relevant and video-relevant rejected responses ($y_l^{er}$, $y_l^{vr}$) works better compared to only using one or using a completely irrelevant response. \cref{subsec:appendix_sensitivity_to_hyperparams} detail the sensitivity of AVEm-DPO to various hyperparameters, highlighting the role of various components in eliminating spurious cue-emotion associations and hallucinations.\\

\begin{figure}[t]
  \centering
  \begin{minipage}{0.25\linewidth}
    \centering
    \includegraphics[width=\linewidth]{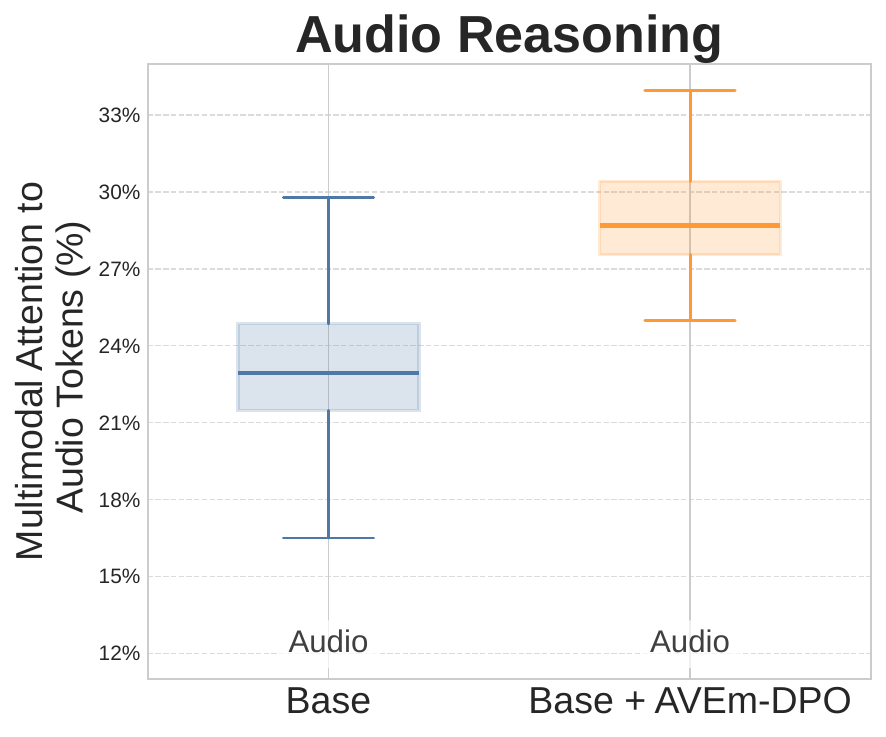}
  \end{minipage}\hfill
  \begin{minipage}{0.25\linewidth}
    \centering
    \includegraphics[width=\linewidth]{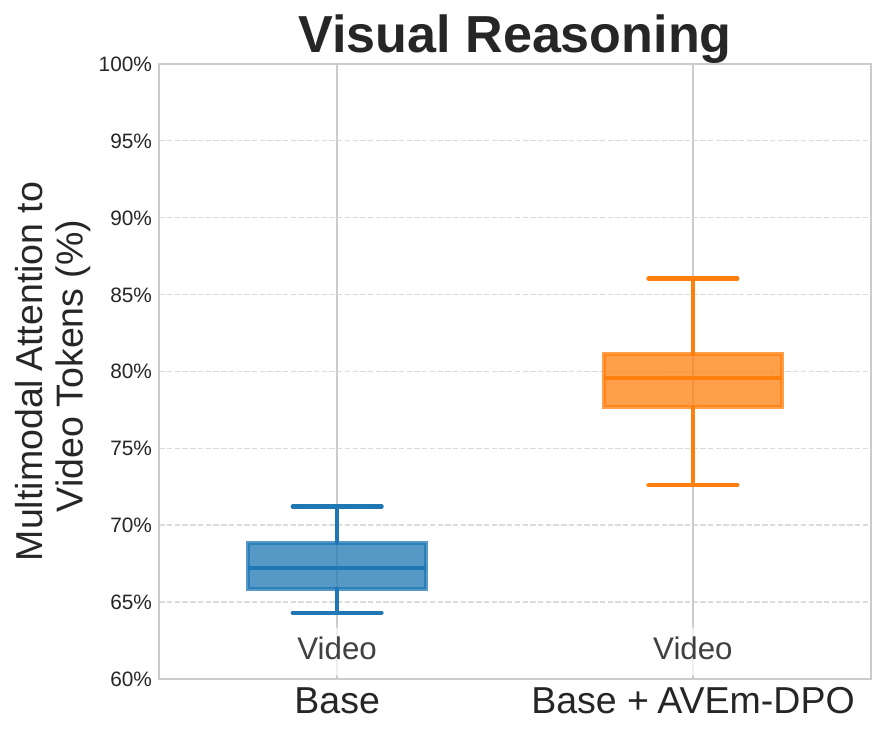}
  \end{minipage}\hfill
  \begin{minipage}{0.25\linewidth}
    \centering
    \includegraphics[width=\linewidth]{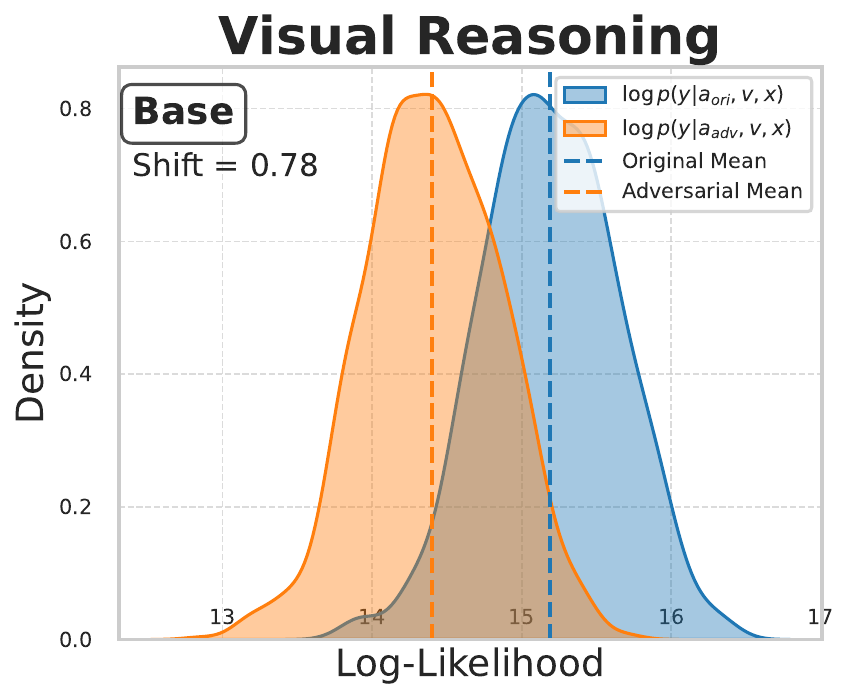} 
  \end{minipage}\hfill
  \begin{minipage}{0.25\linewidth}
    \centering
    \includegraphics[width=\linewidth]{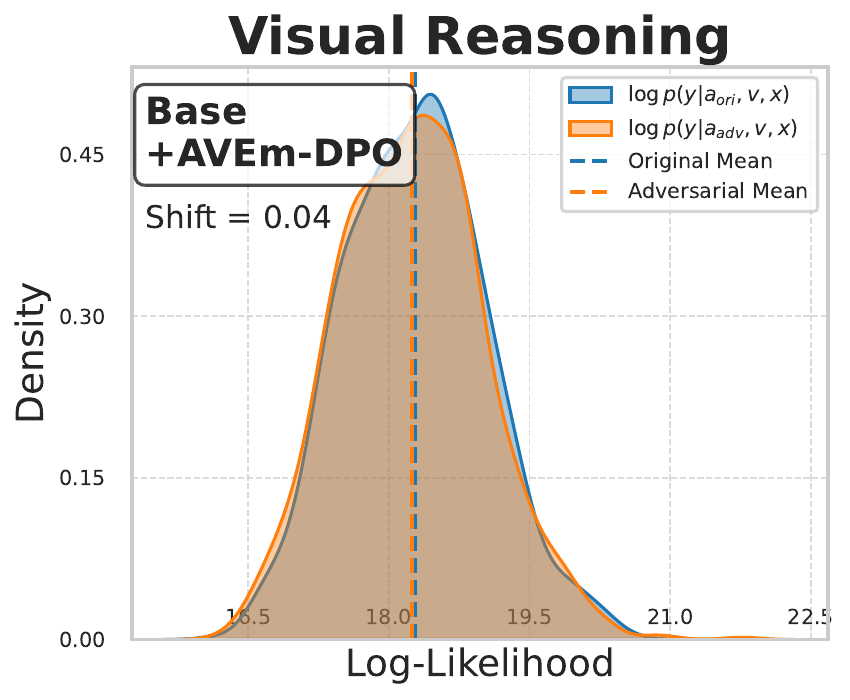} 
  \end{minipage}
  \vspace{-1em}
  \caption{Effect of AVEm-DPO on -- \emph{(Left two plots)} the distribution of attention over video and audio tokens taken as a percentage over the total attention over all multimodal tokens for audio and visual reasoning tasks in \emph{EmoReAlM}; \emph{(Right two plots)} the log-likelihood distribution shift of the correct answer for visual reasoning tasks on corrupting the audio input $a_{ori}$ with adversary $a_{adv}$.}
  \label{fig:attention_analysis}
  \vspace{-1.4em}
\end{figure}

\noindent\textbf{Attention redistribution after AVEm-DPO.} To analyze the effect of preference optimization on model attention, we plot the distribution of aggregate multimodal input attention over audio and visual tokens averaged over all attention heads for audio and visual reasoning tasks in \emph{EmoReAlM} in \cref{fig:attention_analysis} (\emph{left two plots}). We can observe that the attention over relevant modality increases after AVEm-DPO, ensuring consistent model responses grounded on the relevant modality. More attention redistribution experiments are present in \cref{subsec:appendix_attn_redistribution_DPO}.

\textbf{Robustness to adversarial inputs.} As shown in \cref{fig:avem_dpo_adversarial_modality} (\cref{subsec:appendix_reasoning_adversarial_modality}), the model response on a prompt relevant to one modality should not change on changing the input of the irrelevant modality. To test this robustness on visual reasoning tasks, we plot the distribution of log-likelihoods of correct responses for our base and AVEm-DPO models and show the distribution shift using Kernel Density Estimation (KDE) on changing the audio input in \cref{fig:attention_analysis}(\emph{right two plots}). AVEm-DPO trained model results in negligible shifts, showing its robustness. Detailed analysis in \cref{subsec:appendix_reasoning_adversarial_modality}.

\subsection{Validity of Generated Preference Data}

\begin{wraptable}[9]{r}{0.5\linewidth}
\centering
\vspace{-1em}
\caption{Human verification statistics on generated preference data.}
\vspace{-1em}
\label{tab:pref_data_human_verification}
\resizebox{\linewidth}{!}{%
\begin{tabular}{l|ccc}
\hline \hline
\rowcolor[HTML]{C0C0C0} 
\multicolumn{1}{c|}{\cellcolor[HTML]{C0C0C0}\textbf{Response type}} & \textbf{\begin{tabular}[c]{@{}c@{}}\# Total\\ verified\end{tabular}} & \textbf{\begin{tabular}[c]{@{}c@{}}\# Majority\\ correct\end{tabular}} & \textbf{\begin{tabular}[c]{@{}c@{}}\# One or\\ more correct\end{tabular}} \\ \hline
Chosen ($y_w$) & 1000 & 912 & 967 \\
\begin{tabular}[c]{@{}l@{}}Rejected - Video \\ Relevant ($y^{vr}_l$)\end{tabular} & 1000 & 895 & 923 \\
\begin{tabular}[c]{@{}l@{}}Rejected - Emotion\\ Relevant ($y_l^{er}$)\end{tabular} & 1000 & 856 & 912 \\ \hline \hline
\end{tabular}%
}
\end{wraptable}

As mentioned in \cref{subsec:preference_data}, our preference dataset is automatically generated using Gemini 2.5 \citep{comanici2025gemini25pushingfrontier_gemini25}. Performing human verification on the entire training data is too costly. Therefore, to show the validity of the generated preference tuning data, we perform human verification on a subset of 1000 random samples from the generated data with the help of 90 participants recruited through Prolific \citep{prolificProlificEasily}. Each generated sample is verified by three or more annotators. As shown in \cref{tab:pref_data_human_verification}, for the different categories of preference responses mentioned in \cref{subsec:multimodal_preference} -- chosen ($y_w$), video-relevant rejected ($y_l^{vr}$), and emotion-relevant rejected ($y_l^{er}$) -- we report the number of samples in which the majority of annotators found the generated responses correct. These results validate our automatically generated preference data.

\section{Limitations and Future Work}
\label{sec:limitations}
The proposed EmoReAlM benchmark is derived from the DFEW \citep{jiang2020dfew} dataset, leveraging its emotion labels, and hence, it may inherit its cultural biases. Additionally, since our benchmark and training data are derived from existing emotion recognition datasets with short videos ($\sim$ 2-10 seconds), long video emotion understanding and reasoning remain an open topic that can be addressed in future work.

Although the proposed AVEm-DPO significantly improves the reference model's performance, a few limitations remain. Similar to other baselines, our model trained with AVEm-DPO performs poorly on the recognition for \emph{disgust} (an ambiguous emotion \citep{Hendel2023-ow-disgust}) as shown in \cref{subsec:appendix_emotion_recognition_results_expanded,tab:appendix_dfew_results_class_wise}. We attribute this to the limited amount of training samples available for this emotion class. Moreover, a closer look at the performance on the subtasks of the \emph{Emotion Reasoning - Stress Test} task of EmoReAlM (\cref{subsec:appendix_emo_realm_stress_test_results_detailed,tab:emo_realm_hallucination_types}) reveals that there is still room for improvement to mitigate spurious audio cue-emotion associations. 

\section{Conclusion}
This work addresses the bottlenecks of emotion reasoning in MLLMs, with two major contributions -- \emph{EmoReAlM} Benchmark for evaluating emotion reasoning over a complex and diverse set of tasks and \emph{AVEm-DPO} preference optimization technique to mitigate bottlenecks of MLLMs such as spurious audiovisual cue-emotion associations and audiovisual cue hallucinations. The proposed method outperforms open-source baselines on the proposed and existing emotion understanding benchmarks under a zero-shot setting. Moreover, a detailed ablation study with analysis of attention redistribution and log-likelihood shift upon preference tuning supports the efficacy of the proposed prompt-based modality preference and text-prior debiasing approaches.

\textsc{Ethics Statement}

This work builds upon publicly available audiovisual datasets for research purposes, specifically DFEW for benchmark creation (\cref{subsec:benchmark_creation_pipeline}) and MAFW/MER2025 for preference optimization (\cref{subsec:preference_data}). We did not collect new audiovisual data, ensuring no additional privacy risks. All data usage complies with the licensing terms of the original datasets. To mitigate potential harms, the released \emph{EmoReAlM} benchmark will only contain automatically generated and human-verified question–answer pairs; users must independently obtain the underlying videos from the original sources under appropriate licenses. 
For human verification (\cref{subsec:post_processing_human_verification}) and user studies (\cref{tab:user_evaluation_emer}), participants were recruited via Prolific and compensated at fair rates commensurate with task requirements and participant location, aligning with ethical standards for crowd work. We ensured informed consent, anonymity and the right to withdraw at any point.
The proposed methods aim to improve reliability in emotion reasoning by reducing hallucinations and spurious cue associations in multimodal large language models. However, emotion recognition and inference from audiovisual data can carry risks of misinterpretation, bias reinforcement, or misuse in surveillance and high-stakes applications. Moreover, users of the proposed method are advised to read the limitations of the proposed approach mentioned in \cref{sec:limitations} to avoid potential safety concerns. We emphasize that our benchmark and models are intended strictly for academic research, with the goal of advancing robust, interpretable and socially responsible AI. We caution against deployment in sensitive real-world contexts (e.g., healthcare, hiring, law enforcement) without careful domain-specific validation and safeguards.


\textsc{Reproducibility Statement}

To ensure reproducibility and transparency, we provide additional details about data creation and experiments in the Appendix. All the prompts used for data creation are present in \cref{subsec:appendix_prompts_benchmark}. Implementation details for the proposed method, along with hyperparameter settings, are provided in \cref{subsec:appendix_reference_models,subsec:appendix_implementation_details}, while the details about the baseline approaches are present in \cref{subsec:appendix_baseline_implementation_details,subsec:appendix_baseline_preference_optimization_methods}. Details about human verification of the benchmark and user evaluation are present in \cref{subsec:appendix_human_verification_benchmark,subsec:appendix_user_evaluation}. Evaluation metrics are detailed in \cref{subsec:appendix_eval_metrics}. We also provide the detailed setup for our ablations in \cref{subsec:appendix_ablatlion_detailed_setup}. Our benchmark, code and model weights will be made publicly available upon acceptance to ensure reproducibility and ease of use for the proposed work. Code, models and benchmark will be released at \href{https://avere-iclr.github.io/}{avere-iclr.github.io}.

\subsubsection*{Acknowledgements}

Research was sponsored by the Army Research Office and was accomplished under Cooperative Agreement Number W911NF-25-2-0040. Work was also in part supported by the National Science Foundation under Grant IIS-2211550 and the National Institute of Mental Health of the National Institutes of Health under Award Number R61MH135407. The views and conclusions contained in this document are those of the authors and should not be interpreted as representing the official policies, either expressed or implied, of the Army Research Office, NSF, NIH, or the U.S. Government. The U.S. Government is authorized to reproduce and distribute reprints for Government purposes notwithstanding any copyright notation herein.

\bibliography{references}

@inproceedings{zhangbertscore,
  title={BERTScore: Evaluating Text Generation with BERT},
  author={Zhang, Tianyi and Kishore, Varsha and Wu, Felix and Weinberger, Kilian Q and Artzi, Yoav},
  booktitle={International Conference on Learning Representations}
}

@inproceedings{reimers-2019-sentence-bert,
    title = "Sentence-BERT: Sentence Embeddings using Siamese BERT-Networks",
    author = "Reimers, Nils and Gurevych, Iryna",
    booktitle = "Proceedings of the 2019 Conference on Empirical Methods in Natural Language Processing",
    month = "11",
    year = "2019",
    publisher = "Association for Computational Linguistics",
    url = "https://arxiv.org/abs/1908.10084",
}

@ARTICLE{Hendel2023-ow-disgust,
  title    = "Exploration of visual factors in the disgust-anger confusion: the
              importance of the mouth",
  author   = "Hendel, Emalie and Gallant, Ad{\`e}le and Mazerolle, Marie-Pier
              and Cyr, Sabah-Izayah and Roy-Charland, Annie",
  journal  = "Cogn. Emot.",
  volume   =  37,
  number   =  4,
  pages    = "835--851",
  month    =  may,
  year     =  2023,
  language = "en"
}

@incollection{EkmanHandbook2005,
	title = {Basic {Emotions}},
	booktitle = {Handbook of {Cognition} and {Emotion}},
	publisher = {John Wiley \& Sons, Ltd},
	author = {Ekman, Paul},
	year = {2005},
	doi = {10.1002/0470013494.ch3},
	pages = {45--60},
}

@article{scherer2005,
	title = {What are emotions? {And} how can they be measured?},
	volume = {44},
	doi = {10.1177/0539018405058216},
	number = {4},
	journal = {Social Science Information},
	author = {Scherer, Klaus R},
	year = {2005},
	pages = {695--729},
}

@article{lian2023explainable_emer,
  title={Explainable multimodal emotion recognition},
  author={Lian, Zheng and Sun, Haiyang and Sun, Licai and Gu, Hao and Wen, Zhuofan and Zhang, Siyuan and Chen, Shun and Xu, Mingyu and Xu, Ke and Chen, Kang and others},
  journal={arXiv preprint arXiv:2306.15401},
  year={2023}
}

@inproceedings{emotion_llama,
  author = {Cheng, Zebang and Cheng, Zhi-Qi and He, Jun-Yan and Wang, Kai and Lin, Yuxiang and Lian, Zheng and Peng, Xiaojiang and Hauptmann, Alexander},
  booktitle = {Advances in Neural Information Processing Systems},
  editor = {A. Globerson and L. Mackey and D. Belgrave and A. Fan and U. Paquet and J. Tomczak and C. Zhang},
  pages = {110805--110853},
  publisher = {Curran Associates, Inc.},
  title = {Emotion-LLaMA: Multimodal Emotion Recognition and Reasoning with Instruction Tuning},
  url = {https://proceedings.neurips.cc/paper_files/paper/2024/file/c7f43ada17acc234f568dc66da527418-Paper-Conference.pdf},
  volume = {37},
  year = {2024}
}

@article{fu2025vita,
  title={Vita-1.5: Towards gpt-4o level real-time vision and speech interaction},
  author={Fu, Chaoyou and Lin, Haojia and Wang, Xiong and Zhang, Yi-Fan and Shen, Yunhang and Liu, Xiaoyu and Cao, Haoyu and Long, Zuwei and Gao, Heting and Li, Ke and others},
  journal={arXiv preprint arXiv:2501.01957},
  year={2025}
}

@misc{emotion_qwen,
      title={Emotion-Qwen: A Unified Framework for Emotion and Vision Understanding}, 
      author={Dawei Huang and Qing Li and Chuan Yan and Zebang Cheng and Zihao Han and Yurong Huang and Xiang Li and Bin Li and Xiaohui Wang and Zheng Lian and Zhi-Qi Cheng and Xiaojiang Peng},
      year={2025},
      eprint={2505.06685},
      archivePrefix={arXiv},
      primaryClass={cs.MM},
      url={https://arxiv.org/abs/2505.06685}, 
}

@inproceedings{
sung-bin2025_avhbench,
title={{AVHB}ench: A Cross-Modal Hallucination Benchmark for Audio-Visual Large Language Models},
author={Kim Sung-Bin and Oh Hyun-Bin and JungMok Lee and Arda Senocak and Joon Son Chung and Tae-Hyun Oh},
booktitle={The Thirteenth International Conference on Learning Representations},
year={2025},
url={https://openreview.net/forum?id=jTEKTdI3K9}
}

@article{han2025benchmarking_camer_mosear,
  title={Benchmarking and Bridging Emotion Conflicts for Multimodal Emotion Reasoning},
  author={Han, Zhiyuan and Zhu, Beier and Xu, Yanlong and Song, Peipei and Yang, Xun},
  journal={arXiv preprint arXiv:2508.01181},
  year={2025}
}

@inproceedings{jiang2020dfew,
  title={Dfew: A large-scale database for recognizing dynamic facial expressions in the wild},
  author={Jiang, Xingxun and Zong, Yuan and Zheng, Wenming and Tang, Chuangao and Xia, Wanchuang and Lu, Cheng and Liu, Jiateng},
  booktitle={Proceedings of the 28th ACM International Conference on Multimedia},
  pages={2881--2889},
  year={2020}
}

@misc{openai2024gpt4ocard_gpt4o,
      title={GPT-4o System Card}, 
      author={OpenAI and others},
      year={2024},
      eprint={2410.21276},
      archivePrefix={arXiv},
      primaryClass={cs.CL},
      url={https://arxiv.org/abs/2410.21276}, 
}

@misc{comanici2025gemini25pushingfrontier_gemini25,
      title={Gemini 2.5: Pushing the Frontier with Advanced Reasoning, Multimodality, Long Context, and Next Generation Agentic Capabilities}, 
      author={Gemini-Team and others},
      year={2025},
      eprint={2507.06261},
      archivePrefix={arXiv},
      primaryClass={cs.CL},
      url={https://arxiv.org/abs/2507.06261}, 
}

@misc{qwen2025qwen25technicalreport_qwen25,
      title={Qwen2.5 Technical Report}, 
      author={Qwen-Team and others},
      year={2025},
      eprint={2412.15115},
      archivePrefix={arXiv},
      primaryClass={cs.CL},
      url={https://arxiv.org/abs/2412.15115}, 
}

@inproceedings{Xie2024EmoVIT,
  title={EmoVIT: Revolutionizing Emotion Insights with Visual Instruction Tuning},
  author={Hongxia Xie and Chu-Jun Peng and Yu-Wen Tseng and Hung-Jen Chen and Chan-Feng Hsu and Hong-Han Shuai and Wen-Huang Cheng},
  booktitle={Proceedings of the IEEE/CVF Conference on Computer Vision and Pattern Recognition (CVPR)},
  year={2024}
}

@inproceedings{lin-etal-2024-videollava,
    title = "Video-{LL}a{VA}: Learning United Visual Representation by Alignment Before Projection",
    author = "Lin, Bin  and
      Ye, Yang  and
      Zhu, Bin  and
      Cui, Jiaxi  and
      Ning, Munan  and
      Jin, Peng  and
      Yuan, Li",
    editor = "Al-Onaizan, Yaser  and
      Bansal, Mohit  and
      Chen, Yun-Nung",
    booktitle = "Proceedings of the 2024 Conference on Empirical Methods in Natural Language Processing",
    month = nov,
    year = "2024",
    address = "Miami, Florida, USA",
    publisher = "Association for Computational Linguistics",
    url = "https://aclanthology.org/2024.emnlp-main.342/",
    doi = "10.18653/v1/2024.emnlp-main.342",
    pages = "5971--5984"
}

@misc{zhang2024llava_video,
    title={Video Instruction Tuning With Synthetic Data}, 
    author={Yuanhan Zhang and Jinming Wu and Wei Li and Bo Li and Zejun Ma and Ziwei Liu and Chunyuan Li},
    year={2024},
    eprint={2410.02713},
    archivePrefix={arXiv},
    primaryClass={cs.CV},
    url={https://arxiv.org/abs/2410.02713}, 
}

@misc{xu2025qwen25omnitechnicalreport_qwen25omni,
      title={Qwen2.5-Omni Technical Report}, 
      author={Jin Xu and Zhifang Guo and Jinzheng He and Hangrui Hu and Ting He and Shuai Bai and Keqin Chen and Jialin Wang and Yang Fan and Kai Dang and Bin Zhang and Xiong Wang and Yunfei Chu and Junyang Lin},
      year={2025},
      eprint={2503.20215},
      archivePrefix={arXiv},
      primaryClass={cs.CL},
      url={https://arxiv.org/abs/2503.20215}, 
}

@misc{zhang2025videollama3frontiermultimodal,
      title={VideoLLaMA 3: Frontier Multimodal Foundation Models for Image and Video Understanding}, 
      author={Boqiang Zhang and Kehan Li and Zesen Cheng and Zhiqiang Hu and Yuqian Yuan and Guanzheng Chen and Sicong Leng and Yuming Jiang and Hang Zhang and Xin Li and Peng Jin and Wenqi Zhang and Fan Wang and Lidong Bing and Deli Zhao},
      year={2025},
      eprint={2501.13106},
      archivePrefix={arXiv},
      primaryClass={cs.CV},
      url={https://arxiv.org/abs/2501.13106}, 
}

@article{damonlpsg2023videollama,
  title = {Video-LLaMA: An Instruction-tuned Audio-Visual Language Model for Video Understanding},
  author = {Zhang, Hang and Li, Xin and Bing, Lidong},
  journal = {arXiv preprint arXiv:2306.02858},
  year = {2023},
  url = {https://arxiv.org/abs/2306.02858}
}

@misc{li2025baichuanomni15technicalreport_baichuamomni,
      title={Baichuan-Omni-1.5 Technical Report}, 
      author={Yadong Li and team},
      year={2025},
      eprint={2501.15368},
      archivePrefix={arXiv},
      primaryClass={cs.CL},
      url={https://arxiv.org/abs/2501.15368}, 
}

@article{lian2025affectgpt,
  title={AffectGPT: A New Dataset, Model, and Benchmark for Emotion Understanding with Multimodal Large Language Models},
  author={Lian, Zheng and Chen, Haoyu and Chen, Lan and Sun, Haiyang and Sun, Licai and Ren, Yong and Cheng, Zebang and Liu, Bin and Liu, Rui and Peng, Xiaojiang and others},
  journal={ICML},
  year={2025}
}

@article{lian2024open_ov_merd_dataset,
  title={Open-vocabulary Multimodal Emotion Recognition: Dataset, Metric, and Benchmark},
  author={Lian, Zheng and Sun, Haiyang and Sun, Licai and Chen, Lan and Chen, Haoyu and Gu, Hao and Wen, Zhuofan and Chen, Shun and Zhang, Siyuan and Yao, Hailiang and others},
  journal={ICML},
  year={2024}
}

@misc{chaubey2025facellavafacialexpressionattribute_facellava,
      title={Face-LLaVA: Facial Expression and Attribute Understanding through Instruction Tuning}, 
      author={Ashutosh Chaubey and Xulang Guan and Mohammad Soleymani},
      year={2025},
      eprint={2504.07198},
      archivePrefix={arXiv},
      primaryClass={cs.CV},
      url={https://arxiv.org/abs/2504.07198}, 
}

@misc{lian2025affectgptr1leveragingreinforcementlearning,
      title={AffectGPT-R1: Leveraging Reinforcement Learning for Open-Vocabulary Emotion Recognition}, 
      author={Zheng Lian},
      year={2025},
      eprint={2508.01318},
      archivePrefix={arXiv},
      primaryClass={cs.HC},
      url={https://arxiv.org/abs/2508.01318}, 
}

@misc{yang2025omniemotionextendingvideomllm,
      title={Omni-Emotion: Extending Video MLLM with Detailed Face and Audio Modeling for Multimodal Emotion Analysis}, 
      author={Qize Yang and Detao Bai and Yi-Xing Peng and Xihan Wei},
      year={2025},
      eprint={2501.09502},
      archivePrefix={arXiv},
      primaryClass={cs.CV},
      url={https://arxiv.org/abs/2501.09502}, 
}

@inproceedings{wen2025listen,
  title={Listen, Watch, and Learn to Feel: Retrieval-Augmented Emotion Reasoning for Compound Emotion Generation},
  author={Wen, Zhuofan and Lian, Zheng and Chen, Shun and Yao, Hailiang and Yang, Longjiang and Liu, Bin and Tao, Jianhua},
  booktitle={Findings of the Association for Computational Linguistics: ACL 2025},
  pages={11313--11327},
  year={2025}
}

@misc{xing2025emotionhallucerevaluatingemotionhallucinations,
      title={EmotionHallucer: Evaluating Emotion Hallucinations in Multimodal Large Language Models}, 
      author={Bohao Xing and Xin Liu and Guoying Zhao and Chengyu Liu and Xiaolan Fu and Heikki Kälviäinen},
      year={2025},
      eprint={2505.11405},
      archivePrefix={arXiv},
      primaryClass={cs.CV},
      url={https://arxiv.org/abs/2505.11405}, 
}

@inproceedings{rafailov2023direct_dpo_orig,
title={Direct Preference Optimization: Your Language Model is Secretly a Reward Model},
author={Rafael Rafailov and Archit Sharma and Eric Mitchell and Christopher D Manning and Stefano Ermon and Chelsea Finn},
booktitle={Thirty-seventh Conference on Neural Information Processing Systems},
year={2023},
url={https://openreview.net/forum?id=HPuSIXJaa9}
}

@inproceedings{liu2025tisdpo,
title={{TIS}-{DPO}: Token-level Importance Sampling for Direct Preference Optimization With Estimated Weights},
author={Aiwei Liu and Haoping Bai and Zhiyun Lu and Yanchao Sun and Xiang Kong and Xiaoming Simon Wang and Jiulong Shan and Albin Madappally Jose and Xiaojiang Liu and Lijie Wen and Philip S. Yu and Meng Cao},
booktitle={The Thirteenth International Conference on Learning Representations},
year={2025},
url={https://openreview.net/forum?id=oF6e2WwxX0}
}

@inproceedings{wang2024_mdpo,
  title={mDPO: Conditional Preference Optimization for Multimodal Large Language Models},
  author={Wang, Fei and Zhou, Wenxuan and Huang, James Y and Xu, Nan and Zhang, Sheng and Poon, Hoifung and Chen, Muhao},
  journal={Proceedings of EMNLP 2024},
  year={2024}
}

@inproceedings{
huang2025mathcalvistadpo,
title={Vista DPO: Video Hierarchical Spatial-Temporal Direct Preference Optimization for Large Video Models},
author={Haojian Huang and Haodong Chen and Shengqiong Wu and Meng Luo and Jinlan Fu and Xinya Du and Hanwang Zhang and Hao Fei},
booktitle={Forty-second International Conference on Machine Learning},
year={2025},
url={https://openreview.net/forum?id=O2jukIZR50}
}

@inproceedings{
liu2025miadpo,
title={{MIA}-{DPO}: Multi-Image Augmented Direct Preference Optimization For Large Vision-Language Models},
author={Ziyu Liu and Yuhang Zang and Xiaoyi Dong and Pan Zhang and Yuhang Cao and Haodong Duan and Conghui He and Yuanjun Xiong and Dahua Lin and Jiaqi Wang},
booktitle={The Thirteenth International Conference on Learning Representations},
year={2025},
url={https://openreview.net/forum?id=f7WBRSuf9l}
}

@article{Bradley1952RankAO_bradleyterry,
  title={Rank Analysis of Incomplete Block Designs: I. The Method of Paired Comparisons},
  author={Ralph Allan Bradley and Milton E. Terry},
  journal={Biometrika},
  year={1952},
  volume={39},
  pages={324},
  url={https://api.semanticscholar.org/CorpusID:125209808}
}

@inproceedings{
sarkar2025mitigating_halva,
title={Mitigating Object Hallucination in {MLLM}s via Data-augmented Phrase-level Alignment},
author={Pritam Sarkar and Sayna Ebrahimi and Ali Etemad and Ahmad Beirami and Sercan O Arik and Tomas Pfister},
booktitle={The Thirteenth International Conference on Learning Representations},
year={2025},
url={https://openreview.net/forum?id=yG1fW8igzP}
}

@article{Kolomaznik2024,
  title = {The role of socio-emotional attributes in enhancing human-AI collaboration},
  volume = {15},
  ISSN = {1664-1078},
  url = {http://dx.doi.org/10.3389/fpsyg.2024.1369957},
  DOI = {10.3389/fpsyg.2024.1369957},
  journal = {Frontiers in Psychology},
  publisher = {Frontiers Media SA},
  author = {Kolomaznik,  Michal and Petrik,  Vladimir and Slama,  Michal and Jurik,  Vojtech},
  year = {2024},
  month = oct 
}

@article{chaturvedi_social,
title = {Social companionship with artificial intelligence: Recent trends and future avenues},
journal = {Technological Forecasting and Social Change},
volume = {193},
pages = {122634},
year = {2023},
issn = {0040-1625},
doi = {https://doi.org/10.1016/j.techfore.2023.122634},
url = {https://www.sciencedirect.com/science/article/pii/S0040162523003190},
author = {Rijul Chaturvedi and Sanjeev Verma and Ronnie Das and Yogesh K. Dwivedi}
}

@article{elyoseph2024capacity,
  author    = {Elyoseph, Zohar and Refoua, Efrat and Asraf, Keren and Lvovsky, Michael and Shimoni, Yair and Hadar-Shoval, Dalit},
  title     = {Capacity of Generative AI to Interpret Human Emotions From Visual and Textual Data: Pilot Evaluation Study},
  journal   = {JMIR Mental Health},
  volume    = {11},
  pages     = {e54369},
  year      = {2024},
  month     = {Feb},
  doi       = {10.2196/54369},
  pmid      = {38319707},
  pmcid     = {PMC10879976}
}

@article{Salloum2025_education,
  title = {Emotion recognition for enhanced learning: using AI to detect students’ emotions and adjust teaching methods},
  volume = {12},
  ISSN = {2196-7091},
  url = {http://dx.doi.org/10.1186/s40561-025-00374-5},
  DOI = {10.1186/s40561-025-00374-5},
  number = {1},
  journal = {Smart Learning Environments},
  publisher = {Springer Science and Business Media LLC},
  author = {Salloum,  Said A. and Alomari,  Khaled Mohammad and Alfaisal,  Aseel M. and Aljanada,  Rose A. and Basiouni,  Azza},
  year = {2025},
  month = feb 
}

@article{Litendahl2025_health,
  title = {Healthcare Professionals’ Perceptions of Emotional Intelligence in Remote Counselling—A Descriptive Qualitative Study},
  volume = {12},
  ISSN = {2054-1058},
  url = {http://dx.doi.org/10.1002/nop2.70218},
  DOI = {10.1002/nop2.70218},
  number = {4},
  journal = {Nursing Open},
  publisher = {Wiley},
  author = {Litendahl,  Maija and Kaihlaniemi,  Juulia and Autio,  Olli and K\"{a}hk\"{o}nen,  Outi and Oikarinen,  Anne},
  year = {2025},
  month = apr 
}

@article{Balcombe2022_hci,
  title = {Human-Computer Interaction in Digital Mental Health},
  volume = {9},
  ISSN = {2227-9709},
  url = {http://dx.doi.org/10.3390/informatics9010014},
  DOI = {10.3390/informatics9010014},
  number = {1},
  journal = {Informatics},
  publisher = {MDPI AG},
  author = {Balcombe,  Luke and De Leo,  Diego},
  year = {2022},
  month = feb,
  pages = {14}
}

@ARTICLE{dfew_s2d,
  author={Chen, Yin and Li, Jia and Shan, Shiguang and Wang, Meng and Hong, Richang},
  journal={IEEE Transactions on Affective Computing}, 
  title={From Static to Dynamic: Adapting Landmark-Aware Image Models for Facial Expression Recognition in Videos}, 
  year={2024},
  volume={},
  number={},
  pages={1-15},
  doi={10.1109/TAFFC.2024.3453443}}

@INPROCEEDINGS{dfew_m3dfel,
  author={Wang, Hanyang and Li, Bo and Wu, Shuang and Shen, Siyuan and Liu, Feng and Ding, Shouhong and Zhou, Aimin},
  booktitle={2023 IEEE/CVF Conference on Computer Vision and Pattern Recognition (CVPR)}, 
  title={Rethinking the Learning Paradigm for Dynamic Facial Expression Recognition}, 
  year={2023},
  volume={},
  number={},
  pages={17958-17968},
  doi={10.1109/CVPR52729.2023.01722}}

@inproceedings{dfew_mae_dfer,
    author = {Sun, Licai and Lian, Zheng and Liu, Bin and Tao, Jianhua},
    title = {MAE-DFER: Efficient Masked Autoencoder for Self-Supervised Dynamic Facial Expression Recognition},
    year = {2023},
    booktitle = {Proceedings of the 31st ACM International Conference on Multimedia},
    pages = {6110–6121}
}

@book{ekman1978facial,
  author    = {Ekman, Paul and Friesen, Wallace V.},
  title     = {Facial Action Coding System: A Technique for the Measurement of Facial Movement},
  publisher = {Consulting Psychologists Press},
  year      = {1978},
  edition   = {1st},
  address = {Palo Alto, CA}
}

@inproceedings{radford2023robust_whisper,
  title={Robust speech recognition via large-scale weak supervision},
  author={Radford, Alec and Kim, Jong Wook and Xu, Tao and Brockman, Greg and McLeavey, Christine and Sutskever, Ilya},
  booktitle={International conference on machine learning},
  pages={28492--28518},
  year={2023},
  organization={PMLR}
}

@inproceedings{
li2023evaluating_halluc,
title={Evaluating Object Hallucination in Large Vision-Language Models},
author={Yifan Li and Yifan Du and Kun Zhou and Jinpeng Wang and Xin Zhao and Ji-Rong Wen},
booktitle={The 2023 Conference on Empirical Methods in Natural Language Processing},
year={2023},
url={https://openreview.net/forum?id=xozJw0kZXF}
}

@inproceedings{sahoo-etal-2024-comprehensive,
    title = "A Comprehensive Survey of Hallucination in Large Language, Image, Video and Audio Foundation Models",
    author = "Sahoo, Pranab  and
      Meharia, Prabhash  and
      Ghosh, Akash  and
      Saha, Sriparna  and
      Jain, Vinija  and
      Chadha, Aman",
    editor = "Al-Onaizan, Yaser  and
      Bansal, Mohit  and
      Chen, Yun-Nung",
    booktitle = "Findings of the Association for Computational Linguistics: EMNLP 2024",
    month = nov,
    year = "2024",
    address = "Miami, Florida, USA",
    publisher = "Association for Computational Linguistics",
    url = "https://aclanthology.org/2024.findings-emnlp.685/",
    doi = "10.18653/v1/2024.findings-emnlp.685",
    pages = "11709--11724"
}

@inproceedings{hu2022lora,
title={Lo{RA}: Low-Rank Adaptation of Large Language Models},
author={Edward J Hu and yelong shen and Phillip Wallis and Zeyuan Allen-Zhu and Yuanzhi Li and Shean Wang and Lu Wang and Weizhu Chen},
booktitle={International Conference on Learning Representations},
year={2022},
url={https://openreview.net/forum?id=nZeVKeeFYf9}
}

@inbook{liu_mafw_2022,
	author = {Liu, Yuanyuan and Dai, Wei and Feng, Chuanxu and Wang, Wenbin and Yin, Guanghao and Zeng, Jiabei and Shan, Shiguang},
	title = {MAFW: A Large-scale, Multi-modal, Compound Affective Database for Dynamic Facial Expression Recognition in the Wild},
	year = {2022},
	isbn = {978-1-4503-9203-7},
	publisher = {ACM},
	address = {New York, NY, USA},
	url = {https://doi.org/10.1145/3503161.3548190},
	booktitle = {Proceedings of the 30th ACM International Conference on Multimedia (MM’22)},
	numpages = {9}
}

@article{lian2025mer,
  title={Mer 2025: When affective computing meets large language models},
  author={Lian, Zheng and Liu, Rui and Xu, Kele and Liu, Bin and Liu, Xuefei and Zhang, Yazhou and Liu, Xin and Li, Yong and Cheng, Zebang and Zuo, Haolin and others},
  journal={arXiv preprint arXiv:2504.19423},
  year={2025}
}

@article{Livingstone2018_ravdess,
  title = {The Ryerson Audio-Visual Database of Emotional Speech and Song (RAVDESS): A dynamic,  multimodal set of facial and vocal expressions in North American English},
  volume = {13},
  ISSN = {1932-6203},
  url = {http://dx.doi.org/10.1371/journal.pone.0196391},
  DOI = {10.1371/journal.pone.0196391},
  number = {5},
  journal = {PLOS ONE},
  publisher = {Public Library of Science (PLoS)},
  author = {Livingstone,  Steven R. and Russo,  Frank A.},
  editor = {Najbauer,  Joseph},
  year = {2018},
  month = may,
  pages = {e0196391}
}

@inproceedings{lian2023mer2023,
  title={Mer 2023: Multi-label learning, modality robustness, and semi-supervised learning},
  author={Lian, Zheng and Sun, Haiyang and Sun, Licai and Chen, Kang and Xu, Mingyu and Wang, Kexin and Xu, Ke and He, Yu and Li, Ying and Zhao, Jinming and others},
  booktitle={Proceedings of the 31st ACM international conference on multimedia},
  pages={9610--9614},
  year={2023}
}

@INPROCEEDINGS{visual_contrastive_decoding_vcd,
  author={Leng, Sicong and Zhang, Hang and Chen, Guanzheng and Li, Xin and Lu, Shijian and Miao, Chunyan and Bing, Lidong},
  booktitle={2024 IEEE/CVF Conference on Computer Vision and Pattern Recognition (CVPR)}, 
  title={Mitigating Object Hallucinations in Large Vision-Language Models through Visual Contrastive Decoding}, 
  year={2024},
  volume={},
  number={},
  pages={13872-13882},
  doi={10.1109/CVPR52733.2024.01316}}

@article{tang2025video_salmonn2,
  title={video-SALMONN 2: Captioning-Enhanced Audio-Visual Large Language Models},
  author={Tang, Changli and Li, Yixuan and Yang, Yudong and Zhuang, Jimin and Sun, Guangzhi and Li, Wei and Ma, Zejun and Zhang, Chao},
  journal={arXiv preprint arXiv:2506.15220},
  year={2025}
}

@ARTICLE{catplus_dpo,
  author={Ye, Qilang and Yu, Zitong and Shao, Rui and Cui, Yawen and Kang, Xiangui and Liu, Xin and Torr, Philip and Cao, Xiaochun},
  journal={IEEE Transactions on Pattern Analysis and Machine Intelligence}, 
  title={CAT+: Investigating and Enhancing Audio-Visual Understanding in Large Language Models}, 
  year={2025},
  volume={47},
  number={10},
  pages={8674-8690},
  doi={10.1109/TPAMI.2025.3582389}}

@inproceedings{
sun2025videosalmonn_o1,
title={video-{SALMONN}-o1: Reasoning-enhanced Audio-visual Large Language Model},
author={Guangzhi Sun and Yudong Yang and Jimin Zhuang and Changli Tang and Yixuan Li and Wei Li and Zejun MA and Chao Zhang},
booktitle={Forty-second International Conference on Machine Learning},
year={2025},
url={https://openreview.net/forum?id=y62fhuA69I}
}

@article{luo2025openomni,
  title={OpenOmni: Advancing Open-Source Omnimodal Large Language Models with Progressive Multimodal Alignment and Real-Time Self-Aware Emotional Speech Synthesis},
  author={Luo, Run and Lin, Ting-En and Zhang, Haonan and Wu, Yuchuan and Liu, Xiong and Yang, Min and Li, Yongbin and Chen, Longze and Li, Jiaming and Zhang, Lei and others},
  journal={arXiv preprint arXiv:2501.04561},
  year={2025}
}

@article{chen2025omnidpo,
  title={OmniDPO: A Preference Optimization Framework to Address Omni-Modal Hallucination},
  author={Chen, Junzhe and Zhang, Tianshu and Huang, Shiyu and Niu, Yuwei and Sun, Chao and Zhang, Rongzhou and Zhou, Guanyu and Wen, Lijie and Hu, Xuming},
  journal={arXiv preprint arXiv:2509.00723},
  year={2025}
}

@inproceedings{yu2024rlhf_v,
  title={RLHF-V: Towards trustworthy mllms via behavior alignment from fine-grained correctional human feedback},
  author={Yu, Tianyu and Yao, Yuan and Zhang, Haoye and He, Taiwen and Han, Yifeng and Cui, Ganqu and Hu, Jinyi and Liu, Zhiyuan and Zheng, Hai-Tao and Sun, Maosong and others},
  booktitle={Proceedings of the IEEE/CVF Conference on Computer Vision and Pattern Recognition},
  pages={13807--13816},
  year={2024}
}

@inproceedings{
zhu2024languagebind,
title={LanguageBind: Extending Video-Language Pretraining to N-modality by Language-based Semantic Alignment},
author={Bin Zhu and Bin Lin and Munan Ning and Yang Yan and Jiaxi Cui and WANG HongFa and Yatian Pang and Wenhao Jiang and Junwu Zhang and Zongwei Li and Cai Wan Zhang and Zhifeng Li and Wei Liu and Li Yuan},
booktitle={The Twelfth International Conference on Learning Representations},
year={2024},
url={https://openreview.net/forum?id=QmZKc7UZCy}
}

@INPROCEEDINGS{librispeech,
  author={Panayotov, Vassil and Chen, Guoguo and Povey, Daniel and Khudanpur, Sanjeev},
  booktitle={2015 IEEE International Conference on Acoustics, Speech and Signal Processing (ICASSP)}, 
  title={Librispeech: An ASR corpus based on public domain audio books}, 
  year={2015},
  volume={},
  number={},
  pages={5206-5210},
  doi={10.1109/ICASSP.2015.7178964}}

@inproceedings{
jin2024speechcraft,
title={SpeechCraft: A Fine-Grained Expressive Speech Dataset with Natural Language Description},
author={Zeyu Jin and Jia Jia and Qixin Wang and Kehan Li and Shuoyi Zhou and Songtao Zhou and Xiaoyu Qin and Zhiyong Wu},
booktitle={ACM Multimedia 2024},
year={2024},
url={https://openreview.net/forum?id=rjAY1DGUWC}
}

@misc{
leng2025the_curse_of_multi_modality_cmm,
title={The Curse of Multi-Modalities: Evaluating Hallucinations of Large Multimodal Models across Language, Visual, and Audio},
author={Sicong Leng and Yun Xing and Zesen Cheng and Yang Zhou and Hang Zhang and Xin Li and Deli Zhao and Shijian Lu and Chunyan Miao and Lidong Bing},
year={2025},
url={https://openreview.net/forum?id=VeSsiD0DP9}
}

@misc{prolificProlificEasily,
	author = {Prolific},
	title = {{P}rolific | {E}asily collect high-quality data from real people --- prolific.com},
	howpublished = {\url{https://www.prolific.com/}},
	year = {},
	note = {[Accessed 23-09-2025]},
}

@misc{qualtricsQualtricsExperience,
	author = {Qualtrics},
	title = {{Q}ualtrics {X}{M} - {E}xperience {M}anagement {S}oftware --- qualtrics.com},
	howpublished = {\url{https://www.qualtrics.com/}},
	year = {},
	note = {[Accessed 23-09-2025]},
}

@article{goel2025audioflamingo3,
  title={Audio flamingo 3: Advancing audio intelligence with fully open large audio language models},
  author={Goel, Arushi and Ghosh, Sreyan and Kim, Jaehyeon and Kumar, Sonal and Kong, Zhifeng and Lee, Sang-gil and Yang, Chao-Han Huck and Duraiswami, Ramani and Manocha, Dinesh and Valle, Rafael and others},
  journal={arXiv preprint arXiv:2507.08128},
  year={2025}
}

@article{Qwen2-Audio,
  title={Qwen2-Audio Technical Report},
  author={Chu, Yunfei and Xu, Jin and Yang, Qian and Wei, Haojie and Wei, Xipin and Guo,  Zhifang and Leng, Yichong and Lv, Yuanjun and He, Jinzheng and Lin, Junyang and Zhou, Chang and Zhou, Jingren},
  journal={arXiv preprint arXiv:2407.10759},
  year={2024}
}

@article{ding2025kimiaudio,
  title={Kimi-audio technical report},
  author={Ding, Ding and Ju, Zeqian and Leng, Yichong and Liu, Songxiang and Liu, Tong and Shang, Zeyu and Shen, Kai and Song, Wei and Tan, Xu and Tang, Heyi and others},
  journal={arXiv preprint arXiv:2504.18425},
  year={2025}
}

@misc{xu2025qwen25omnitechnicalreport,
      title={Qwen2.5-Omni Technical Report}, 
      author={Jin Xu and Zhifang Guo and Jinzheng He and Hangrui Hu and Ting He and Shuai Bai and Keqin Chen and Jialin Wang and Yang Fan and Kai Dang and Bin Zhang and Xiong Wang and Yunfei Chu and Junyang Lin},
      year={2025},
      eprint={2503.20215},
      archivePrefix={arXiv},
      primaryClass={cs.CL},
      url={https://arxiv.org/abs/2503.20215}, 
}

@misc{su2023pandagptmodelinstructionfollow,
      title={PandaGPT: One Model To Instruction-Follow Them All}, 
      author={Yixuan Su and Tian Lan and Huayang Li and Jialu Xu and Yan Wang and Deng Cai},
      year={2023},
      eprint={2305.16355},
      archivePrefix={arXiv},
      primaryClass={cs.CL},
      url={https://arxiv.org/abs/2305.16355}, 
}

@misc{girdhar2023imagebindembeddingspacebind,
      title={ImageBind: One Embedding Space To Bind Them All}, 
      author={Rohit Girdhar and Alaaeldin El-Nouby and Zhuang Liu and Mannat Singh and Kalyan Vasudev Alwala and Armand Joulin and Ishan Misra},
      year={2023},
      eprint={2305.05665},
      archivePrefix={arXiv},
      primaryClass={cs.CV},
      url={https://arxiv.org/abs/2305.05665}, 
}

@misc{han2025onellmframeworkalignmodalities,
      title={OneLLM: One Framework to Align All Modalities with Language}, 
      author={Jiaming Han and Kaixiong Gong and Yiyuan Zhang and Jiaqi Wang and Kaipeng Zhang and Dahua Lin and Yu Qiao and Peng Gao and Xiangyu Yue},
      year={2025},
      eprint={2312.03700},
      archivePrefix={arXiv},
      primaryClass={cs.CV},
      url={https://arxiv.org/abs/2312.03700}, 
}

@misc{wang2025internvl35advancingopensourcemultimodal,
      title={InternVL3.5: Advancing Open-Source Multimodal Models in Versatility, Reasoning, and Efficiency}, 
      author={Weiyun Wang and Zhangwei Gao and Lixin Gu and Hengjun Pu and Long Cui and Xingguang Wei and Zhaoyang Liu and Linglin Jing and Shenglong Ye and Jie Shao and Zhaokai Wang and Zhe Chen and Hongjie Zhang and Ganlin Yang and Haomin Wang and Qi Wei and Jinhui Yin and Wenhao Li and Erfei Cui and Guanzhou Chen and Zichen Ding and Changyao Tian and Zhenyu Wu and Jingjing Xie and Zehao Li and Bowen Yang and Yuchen Duan and Xuehui Wang and Zhi Hou and Haoran Hao and Tianyi Zhang and Songze Li and Xiangyu Zhao and Haodong Duan and Nianchen Deng and Bin Fu and Yinan He and Yi Wang and Conghui He and Botian Shi and Junjun He and Yingtong Xiong and Han Lv and Lijun Wu and Wenqi Shao and Kaipeng Zhang and Huipeng Deng and Biqing Qi and Jiaye Ge and Qipeng Guo and Wenwei Zhang and Songyang Zhang and Maosong Cao and Junyao Lin and Kexian Tang and Jianfei Gao and Haian Huang and Yuzhe Gu and Chengqi Lyu and Huanze Tang and Rui Wang and Haijun Lv and Wanli Ouyang and Limin Wang and Min Dou and Xizhou Zhu and Tong Lu and Dahua Lin and Jifeng Dai and Weijie Su and Bowen Zhou and Kai Chen and Yu Qiao and Wenhai Wang and Gen Luo},
      year={2025},
      eprint={2508.18265},
      archivePrefix={arXiv},
      primaryClass={cs.CV},
      url={https://arxiv.org/abs/2508.18265}, 
}

@misc{bai2025qwen25vltechnicalreport,
      title={Qwen2.5-VL Technical Report}, 
      author={Shuai Bai and Keqin Chen and Xuejing Liu and Jialin Wang and Wenbin Ge and Sibo Song and Kai Dang and Peng Wang and Shijie Wang and Jun Tang and Humen Zhong and Yuanzhi Zhu and Mingkun Yang and Zhaohai Li and Jianqiang Wan and Pengfei Wang and Wei Ding and Zheren Fu and Yiheng Xu and Jiabo Ye and Xi Zhang and Tianbao Xie and Zesen Cheng and Hang Zhang and Zhibo Yang and Haiyang Xu and Junyang Lin},
      year={2025},
      eprint={2502.13923},
      archivePrefix={arXiv},
      primaryClass={cs.CV},
      url={https://arxiv.org/abs/2502.13923}, 
}

@inproceedings{zhang-etal-2025-direct-hound-dpo,
    title = "Direct Preference Optimization of Video Large Multimodal Models from Language Model Reward",
    author = "Zhang, Ruohong  and
      Gui, Liangke  and
      Sun, Zhiqing  and
      Feng, Yihao  and
      Xu, Keyang  and
      Zhang, Yuanhan  and
      Fu, Di  and
      Li, Chunyuan  and
      Hauptmann, Alexander G  and
      Bisk, Yonatan  and
      Yang, Yiming",
    editor = "Chiruzzo, Luis  and
      Ritter, Alan  and
      Wang, Lu",
    booktitle = "Proceedings of the 2025 Conference of the Nations of the Americas Chapter of the Association for Computational Linguistics: Human Language Technologies (Volume 1: Long Papers)",
    month = apr,
    year = "2025",
    address = "Albuquerque, New Mexico",
    publisher = "Association for Computational Linguistics",
    url = "https://aclanthology.org/2025.naacl-long.30/",
    doi = "10.18653/v1/2025.naacl-long.30",
    pages = "694--717",
    ISBN = "979-8-89176-189-6"
}
\bibliographystyle{iclr2026_conference}

\clearpage
\appendix

{
        \centering
        \huge
        \vspace{0.5em} \textsc{Appendix} \\
        \vspace{1.0em}
}


{\noindent \Large \textsc{Table of Contents}
}

\begin{itemize}
    \item {\textsc{LLM Usage}} \dotfill \ref{sec:appendix_llm_usage}
    \item {\textsc{Benchmark Details}} \dotfill \ref{sec:appendix_benchmark_details}
    \begin{itemize}
        \item {\small \textsc{Prompts used in Benchmark Creation}} \dotfill \ref{subsec:appendix_prompts_benchmark}
        \item {\small \textsc{Human Verification}} \dotfill \ref{subsec:appendix_human_verification_benchmark}
        \item {\small \textsc{Benchmark Statistics}} \dotfill \ref{subsec:appendix_benchmark_statistics}
        \item {\small \textsc{Frame Sampling Rate for Automatic Visual Captioning}} \dotfill \ref{subsec:frame_sampling_rate}
        \item {\small \textsc{Benchmark Samples}} \dotfill \ref{subsec:appendix_benchmark_samples}
    \end{itemize}
    \item {\textsc{Methodological Details}} \dotfill \ref{sec:appendix_method_details}
    \begin{itemize}
        \item {\small \textsc{Text-Prior Debiasing}} \dotfill \ref{subsec:appenix_equations_tpd}
        \item {\small \textsc{Preference Data}} \dotfill \ref{subsec:appendix_preference_data}
        \item {\small \textsc{Implementation Details}} \dotfill \ref{subsec:appendix_implementation_details}
    \end{itemize}
    \item {\textsc{Experimental Details}} \dotfill \ref{sec:appendix_experimental_details}
    \begin{itemize}
        \item {\small \textsc{Evaluation Metrics}} \dotfill \ref{subsec:appendix_eval_metrics}
        \item {\small \textsc{Reference Models}} \dotfill \ref{subsec:appendix_reference_models}
        \item {\small \textsc{Baseline Preference Optimization Techniques}} \dotfill \ref{subsec:appendix_baseline_preference_optimization_methods}
        \item {\small \textsc{Baseline Implementations}} \dotfill \ref{subsec:appendix_baseline_implementation_details}
        \item {\small \textsc{Experimental Setup for Ablation Study}} \dotfill \ref{subsec:appendix_ablatlion_detailed_setup}
    \end{itemize}
    \item {\textsc{Detailed Results}} \dotfill \ref{sec:appendix_detailed_results}
    \begin{itemize}
        \item {\small \textsc{EmoReAlM Results - Expanded}} \dotfill \ref{subsec:appendix_emo_realm_results_detailed_expanded}
        \item {\small \textsc{EmoReAlM Results on different Stress Test Subtasks}} \dotfill \ref{subsec:appendix_emo_realm_stress_detailed}
        \item {\small \textsc{Emotion Recognition Results - Expanded}} \dotfill \ref{subsec:appendix_emotion_recognition_results_expanded}
        \item {\small \textsc{User Evaluation}} \dotfill \ref{subsec:appendix_user_evaluation}
        \item {\small \textsc{Modality Preference Ablation}} \dotfill \ref{subsec:appendix_modality_preference_ablation}
        \item {\small \textsc{Response Preference Ablation}} \dotfill \ref{subsec:appendix_response_preference_ablation}
        \item {\small \textsc{Sensitivity to Hyperparameters}} \dotfill \ref{subsec:appendix_sensitivity_to_hyperparams}
        \item {\small \textsc{Attention Redistribution after Preference Optimization}} \dotfill \ref{subsec:appendix_attn_redistribution_DPO}
        \item {\small \textsc{Reasoning with Adversarial Modality Inputs}} \dotfill \ref{subsec:appendix_reasoning_adversarial_modality}
        \item {\small \textsc{Effect of Individual Modalities for Emotion Prediction}} \dotfill \ref{subsec:individual_modality_effect}
    \end{itemize}
    \item {\textsc{Qualitative Samples}} \dotfill \ref{sec:appendix_qualitative_samples}
    \item {\textsc{Prompt Pool}} \dotfill \ref{sec:appendix_prompt_pool}
\end{itemize}

\section{LLM Usage}
\label{sec:appendix_llm_usage}
We used GPT-5 to polish the text we added to the paper for grammar and consistency checks. We verify the grammar changes suggested by GPT to ensure its validity. No significant part of the text in the paper is written by any LLM. Apart from polishing the paper, we use LLMs for data annotation and automatic evaluation as mentioned in \cref{subsec:benchmark_creation_pipeline,subsec:preference_data,subsec:appendix_preference_data,subsec:appendix_reference_models,subsec:appendix_eval_metrics}. 

\section{Benchmark Details}
\label{sec:appendix_benchmark_details}

\subsection{Prompts used in Benchmark Creation}
\label{subsec:appendix_prompts_benchmark}
In this section, we detail the prompts that are used in various parts of the benchmark creation pipeline mentioned in \cref{sec:benchmark,fig:emo_realm_data_pipeline}. Note that the text prompts themselves are present at the end of the document in \cref{sec:appendix_prompt_pool}. 

\paragraph{Audio and Video Captioning.} \cref{fig:prompt_audio_caption,fig:prompt_video_caption} contains the prompts used to caption the audio and visual content separately for a given video as described in \cref{subsec:benchmark_creation_pipeline,fig:emo_realm_data_pipeline}. For visual captioning, we sample eight uniform frames from the video and pass those to GPT-4o. For audio captioning, we only pass the audio as a WAV file to GPT-4o-audio. 

\begin{figure}[!h]
    \centering
    \includegraphics[width=\linewidth]{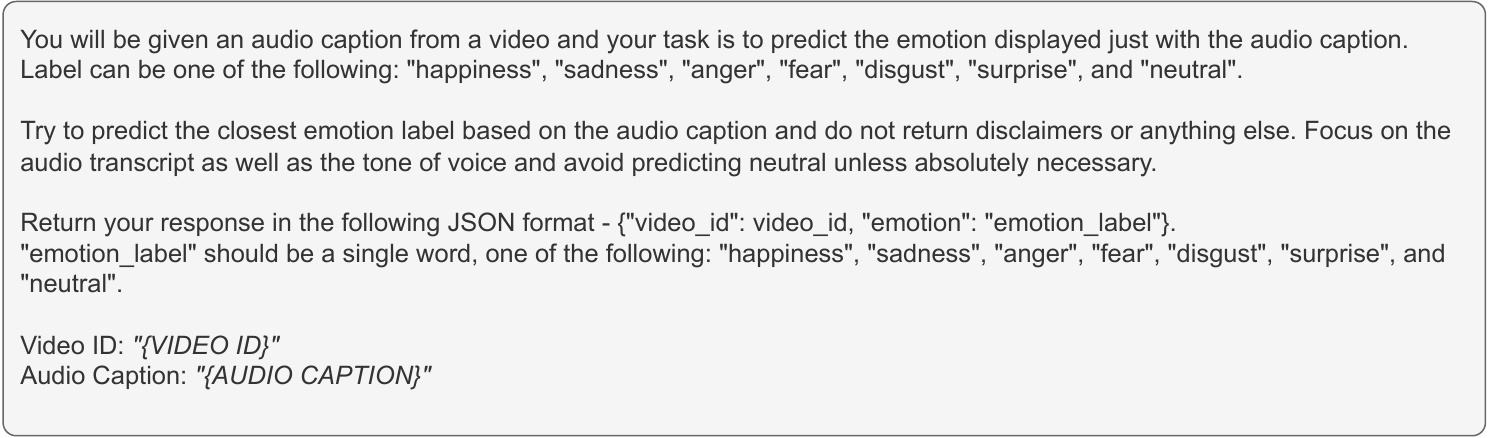}
    \caption{\textbf{Audio emotion prediction prompt} -- used to predict the emotion into one of the 7 basic categories from only the audio caption.}
    \label{fig:prompt_audio_emotion_prediction}
\end{figure}

\begin{figure}[!h]
    \centering
    \includegraphics[width=\linewidth]{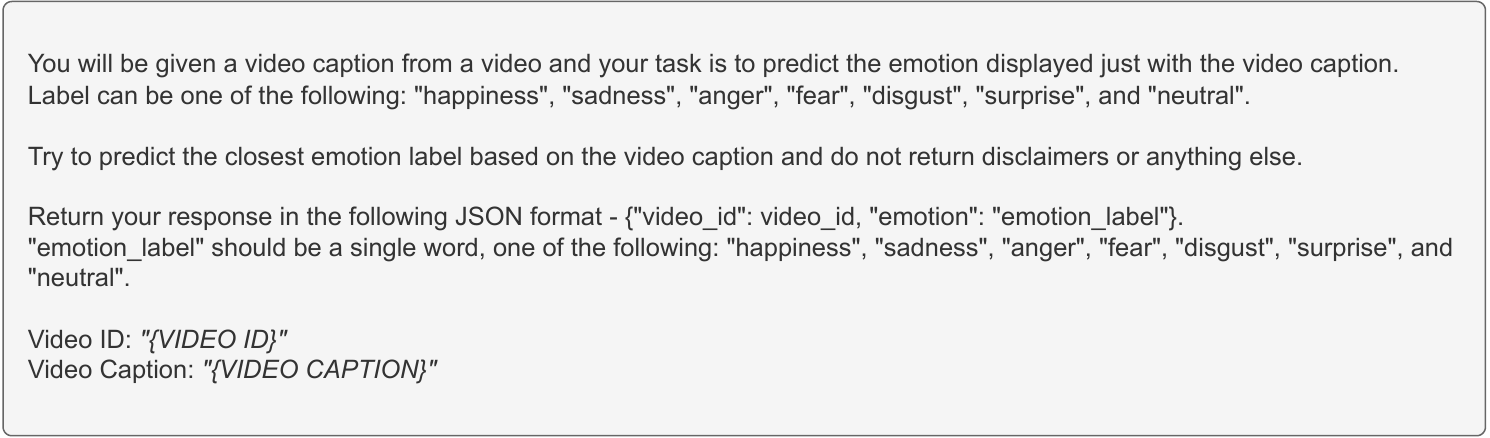}
    \caption{\textbf{Video emotion prediction prompt} --  used to predict the emotion into one of the 7 basic categories from only the video caption}
    \label{fig:prompt_video_emotion_prediction}
\end{figure}

\paragraph{Emotion prediction from audio and video captions separately.} \cref{fig:prompt_audio_emotion_prediction,fig:prompt_video_emotion_prediction} contain prompts used to predict the emotion (out of the seven basic categories) just using the audio and video captions separately. If the ground truth emotion label cannot be predicted by both the audio and video captions, then we do not proceed with such a video for the subsequent data pipeline.

\begin{figure}[!h]
    \centering
    \includegraphics[width=\linewidth]{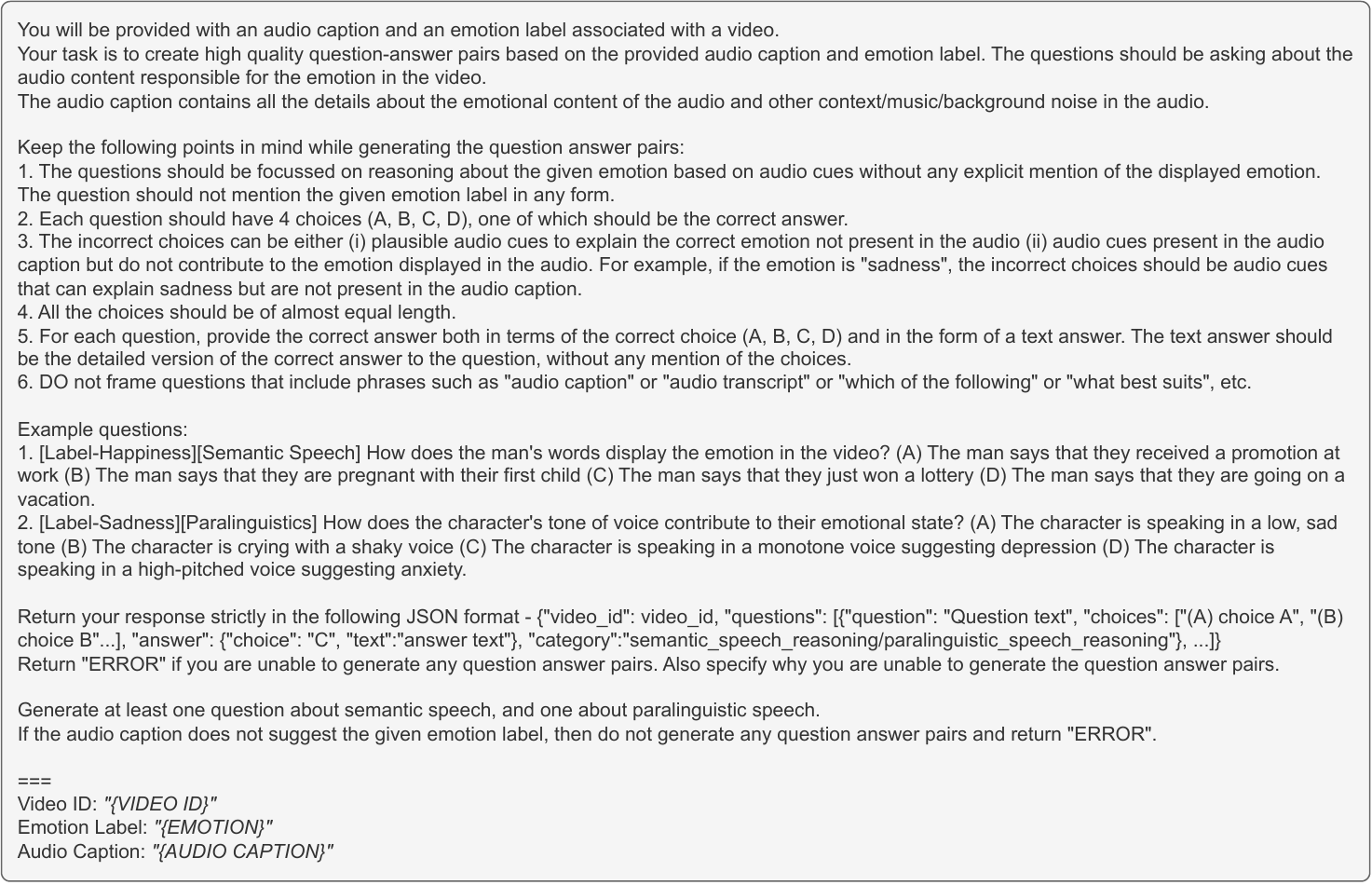}
    \caption{\textbf{EmoReAlM Basic Reasoning Prompt - Audio} --  used to generate questions which ask about the audio cues that suggest the emotion of the person in the video.}
    \label{fig:prompt_emorealm_reasoning_audio}
\end{figure}

\begin{figure}[!h]
    \centering
    \includegraphics[width=\linewidth]{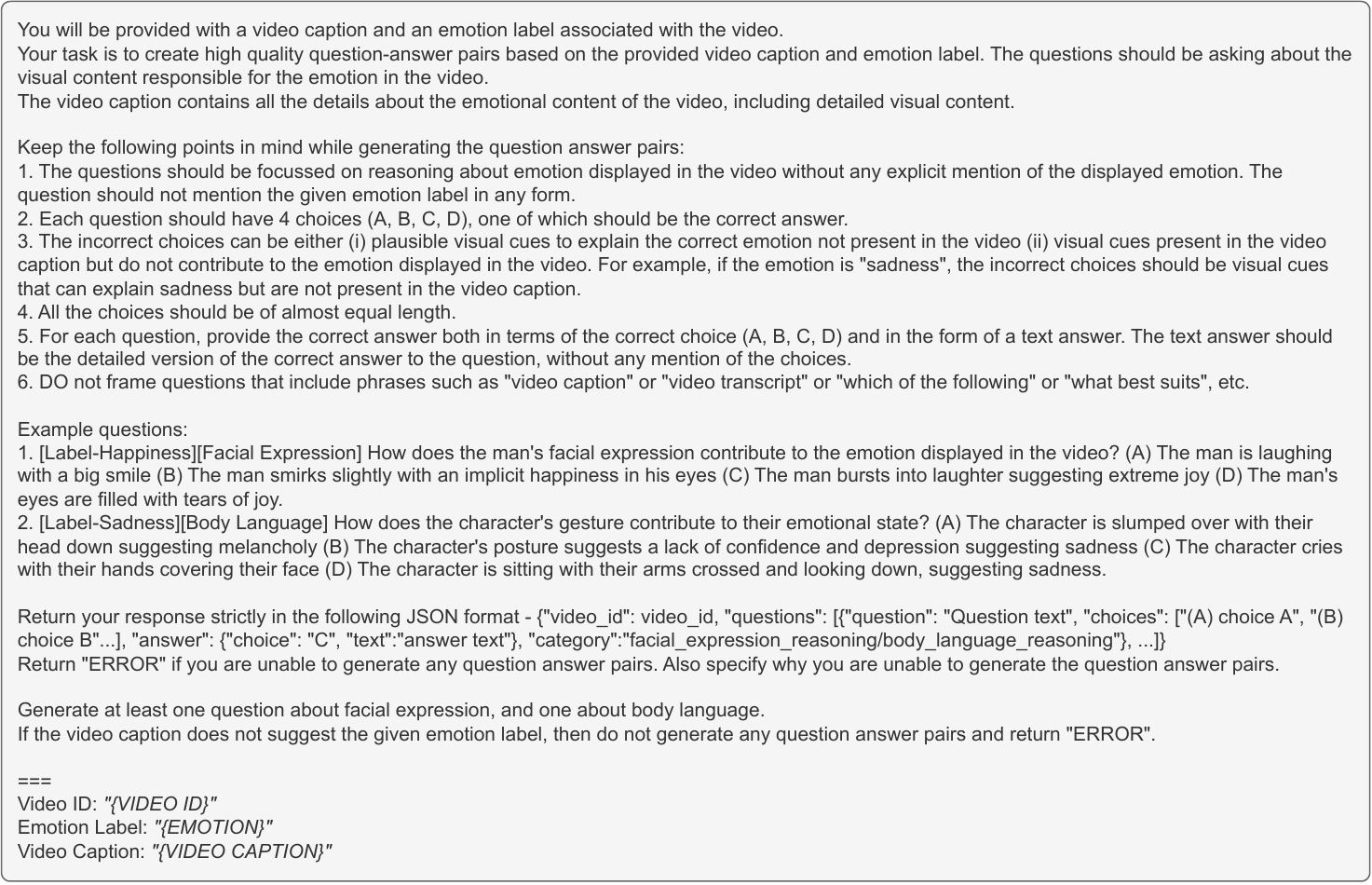}
    \caption{\textbf{EmoReAlM Basic Reasoning Prompt - Visual} --  used to generate questions which ask about the visual cues that suggest the emotion of the person in the video.}
    \label{fig:prompt_emorealm_reasoning_visual}
\end{figure}

\paragraph{EmoReAlM QA Generation.} \cref{fig:prompt_emorealm_reasoning_audio,fig:prompt_emorealm_reasoning_visual} contains the prompts to generate questions related to \emph{Emotion Reasoning - Basic} as described in \cref{subsec:task_descriptions} for audio and visual reasoning respectively. We use the ground truth emotion label already present in the source emotion recognition dataset, as well as the audio/video captions, to generate the question answers. Note that audio and visual reasoning samples are only generated for those samples in which emotion was predicted correctly from the audio and visual captions, respectively (using prompts in \cref{fig:prompt_audio_emotion_prediction,fig:prompt_video_emotion_prediction}).

\begin{figure}[!h]
    \centering
    \includegraphics[width=\linewidth]{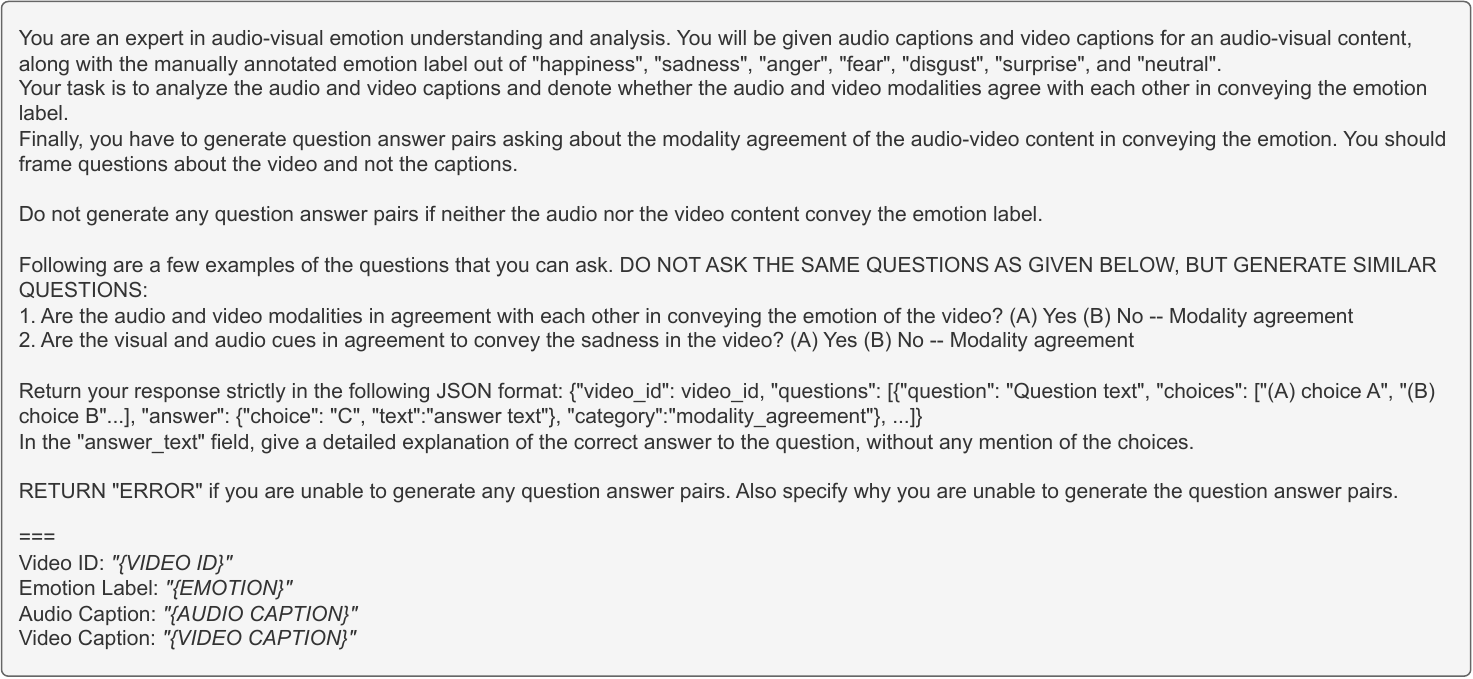}
    \caption{\textbf{EmoReAlM Modality Agreement Prompt} --  used to generate questions which ask whether the audio and video in the audiovisual input align with each other to express the emotion in the video.}
    \label{fig:prompt_emorealm_modality_agreement}
\end{figure}

We use prompt in \cref{fig:prompt_emorealm_modality_agreement} to generate questions related to \emph{Modality Agreement} (\cref{subsec:task_descriptions}) by passing the audio captions, video captions and the ground truth emotion label present in the source dataset. We also verify the answers to the generated questions using the ground truth emotion label present for the video and the emotions predicted using only audio and video captions. If both the audio and the video caption predict the ground truth emotion label from the captions (using prompts in \cref{fig:prompt_audio_emotion_prediction,fig:prompt_video_emotion_prediction}), then the correct answer should be \emph{``Yes"}, else it should be \emph{``No"}.

\begin{figure}[!h]
    \centering
    \includegraphics[width=\linewidth]{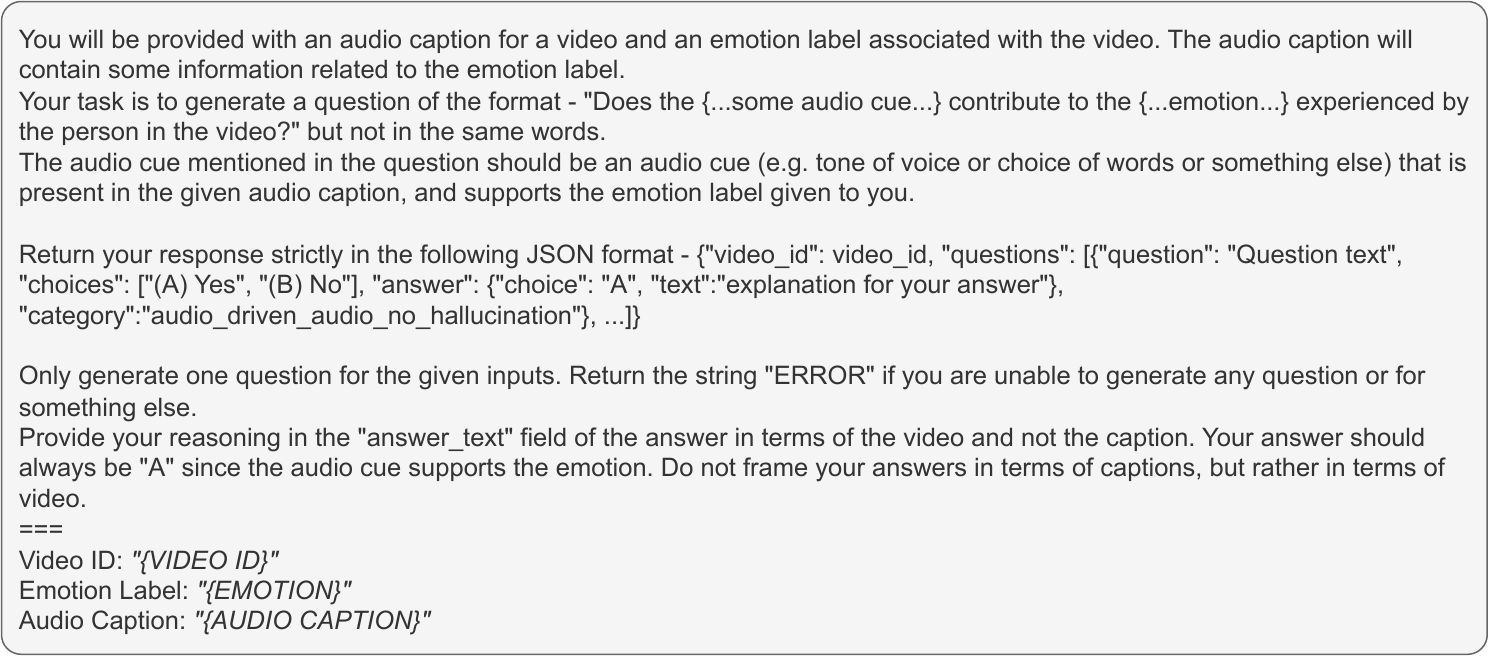}
    \caption{\textbf{EmoReAlM Stress Test Prompt - Audio - No Hallucination} --  used to generate questions where the audio cue mentioned in the question is present in the audiovisual input and supports the emotion of the person in the video.}
    \label{fig:prompt_emorealm_audio_no_hallucination}
\end{figure}

\begin{figure}[!h]
    \centering
    \includegraphics[width=\linewidth]{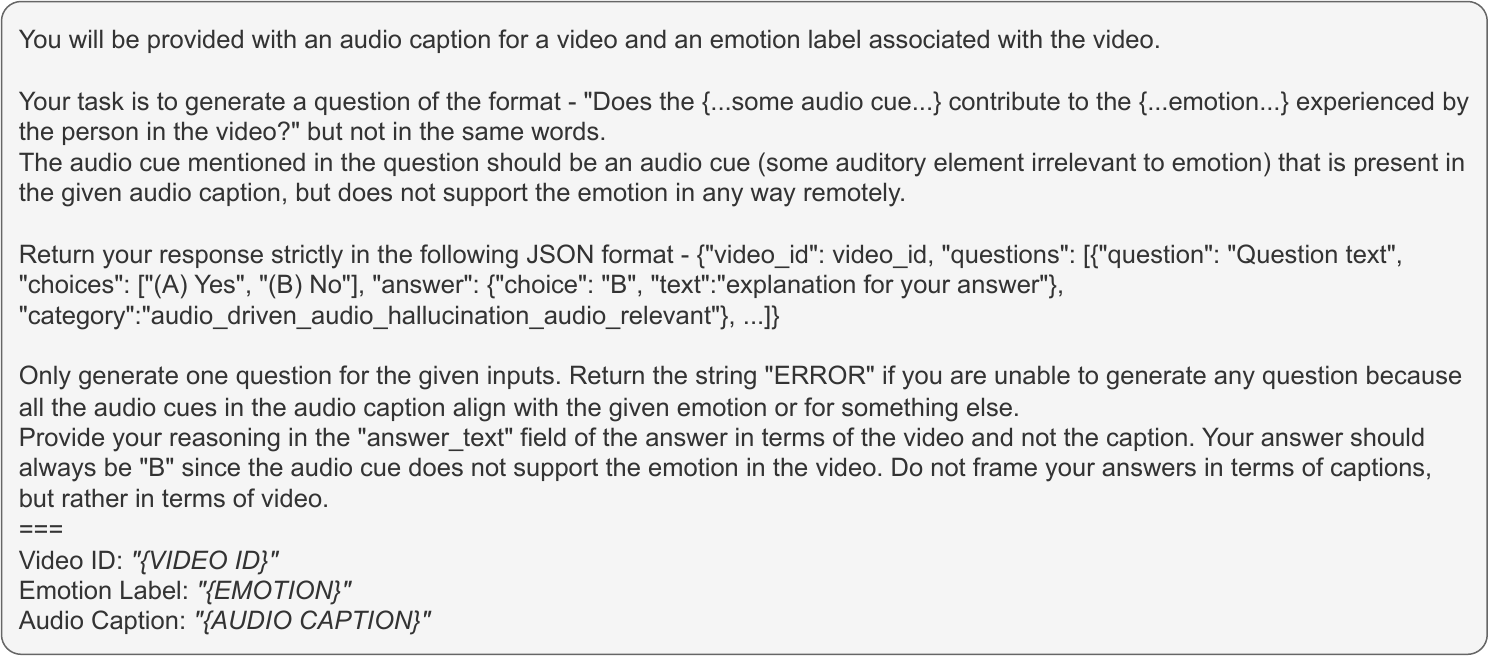}
    \caption{\textbf{EmoReAlM Stress Test Prompt - Audio - Spurious Associations} --  used to generate questions where the audio cue mentioned in the question is present in the audiovisual input and but it is spuriously related to the emotion of the person in the video.}
    \label{fig:prompt_emorealm_audio_spurious}
\end{figure}

\begin{figure}[!h]
    \centering
    \includegraphics[width=\linewidth]{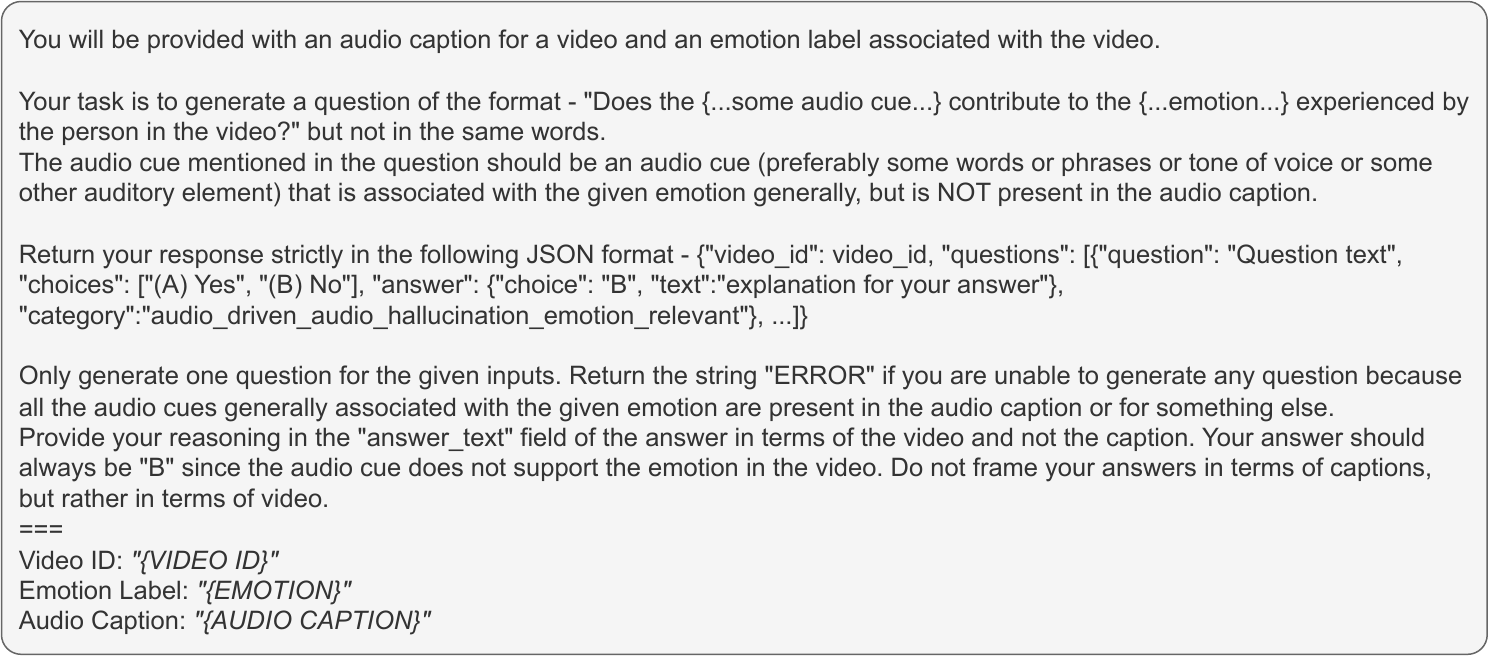}
    \caption{\textbf{EmoReAlM Stress Test Prompt - Audio - Hallucination} --  used to generate questions where the audio cue mentioned in the question is hallucinated (not present in the audiovisual input) and but it usually explains the emotion experienced by the person in the video.}
    \label{fig:prompt_emorealm_audio_hallucination}
\end{figure}

\begin{figure}[!h]
    \centering
    \includegraphics[width=\linewidth]{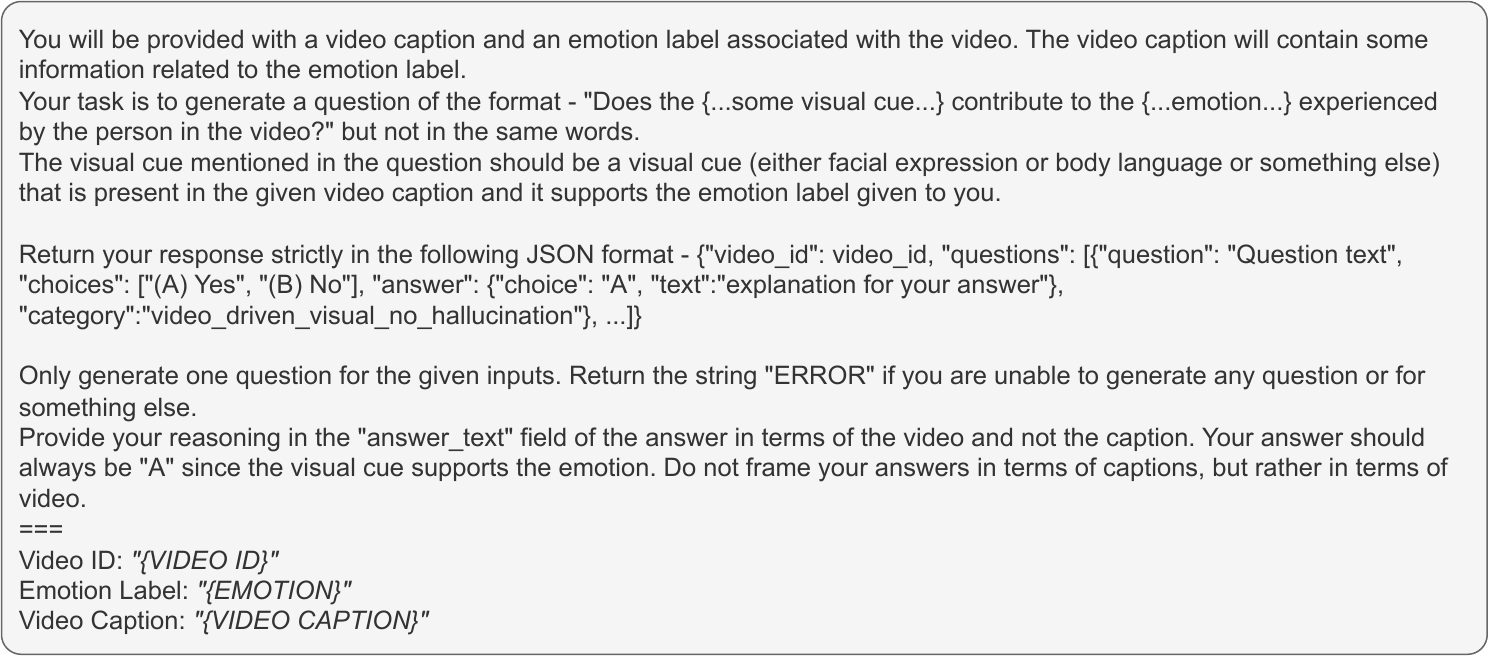}
    \caption{\textbf{EmoReAlM Stress Test Prompt - Video - No Hallucination} --  used to generate questions where the visual cue mentioned in the question is present in the audiovisual input and supports the emotion of the person in the video.}
    \label{fig:prompt_emorealm_visual_no_hallucination}
\end{figure}

\begin{figure}[!h]
    \centering
    \includegraphics[width=\linewidth]{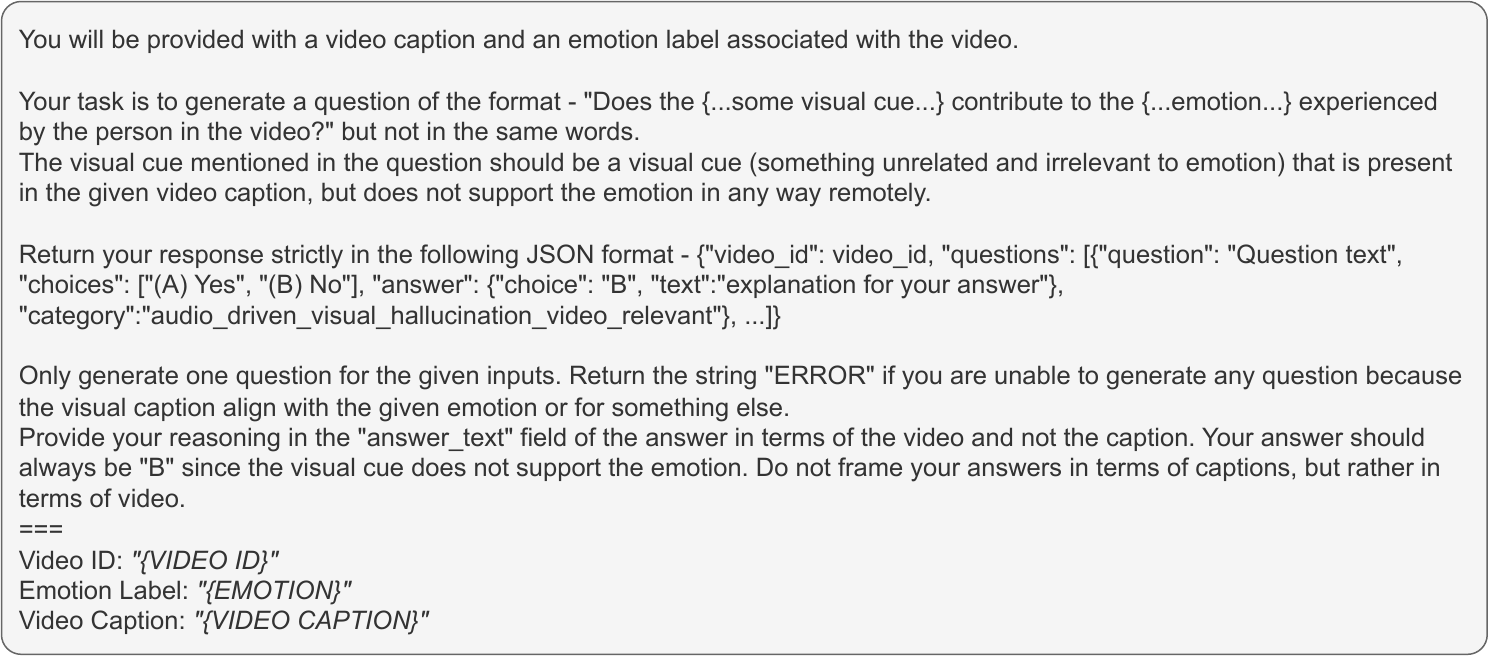}
    \caption{\textbf{EmoReAlM Stress Test Prompt - Video - Spurious Associations} --  used to generate questions where the visual cue mentioned in the question is present in the audiovisual input and but it is spuriously related to the emotion of the person in the video.}
    \label{fig:prompt_emorealm_visual_spurious}
\end{figure}

\begin{figure}[!h]
    \centering
    \includegraphics[width=\linewidth]{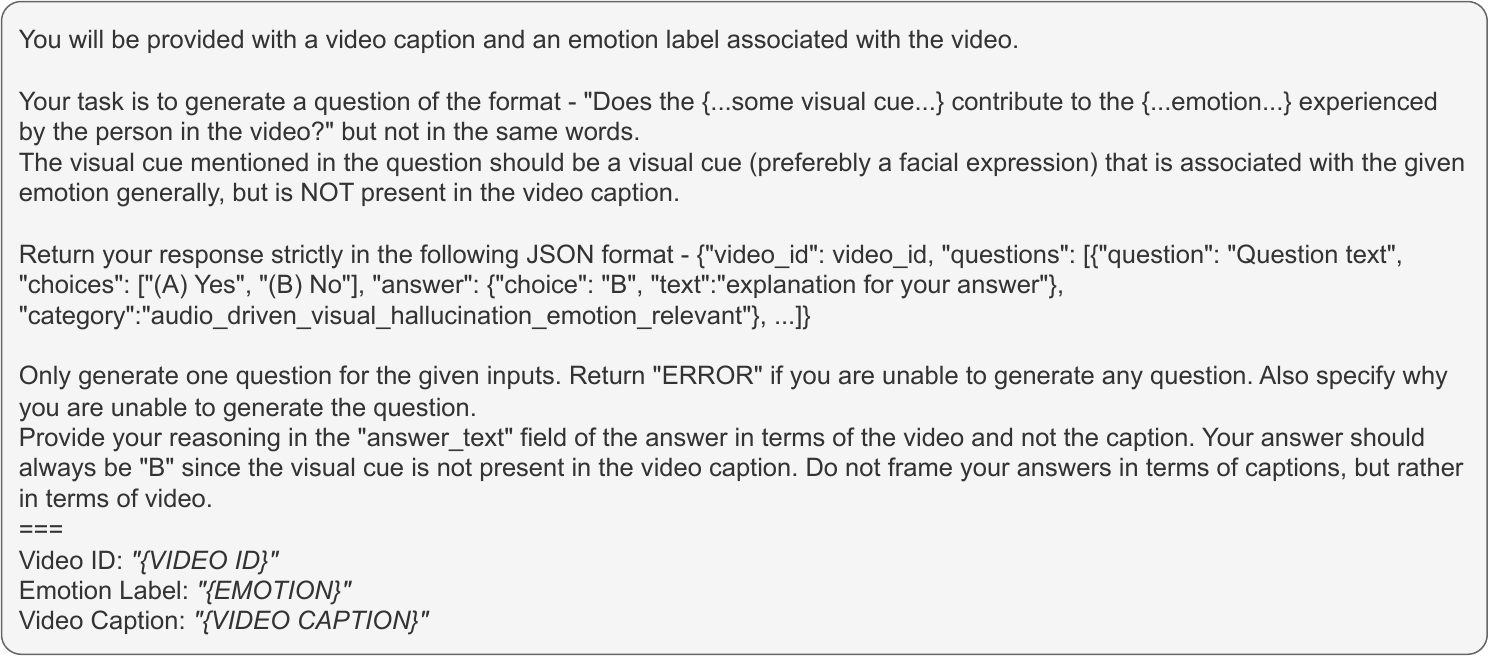}
    \caption{\textbf{EmoReAlM Stress Test Prompt - Video - Hallucination} --  used to generate questions where the visual cue mentioned in the question is hallucinated (not present in the audiovisual input) and but it usually explains the emotion experienced by the person in the video.}
    \label{fig:prompt_emorealm_visual_hallucination}
\end{figure}

For the \emph{Emotion Reasoning - Stress Test} (\cref{subsec:task_descriptions}), we generate questions using prompts present in \cref{fig:prompt_emorealm_audio_no_hallucination,fig:prompt_emorealm_audio_hallucination,fig:prompt_emorealm_audio_spurious,fig:prompt_emorealm_visual_no_hallucination,fig:prompt_emorealm_visual_hallucination,fig:prompt_emorealm_visual_spurious}. We use separate prompts for generating questions related for the different subtasks -- \emph{No hallucination} (\cref{fig:prompt_emorealm_audio_no_hallucination,fig:prompt_emorealm_visual_no_hallucination}), \emph{Spurious Cue-Emotion Association} (\cref{fig:prompt_emorealm_audio_spurious,fig:prompt_emorealm_visual_spurious}) and \emph{Emotion-relevant Hallucination} (\cref{fig:prompt_emorealm_audio_hallucination,fig:prompt_emorealm_visual_hallucination}). Note that the \emph{No hallucination} prompts only apply to cases where the emotion prediction from the audio and/or visual captions using \cref{fig:prompt_audio_emotion_prediction,fig:prompt_video_emotion_prediction} is same as the ground truth emotion label.

\begin{figure}[!h]
    \centering
    \includegraphics[width=\linewidth]{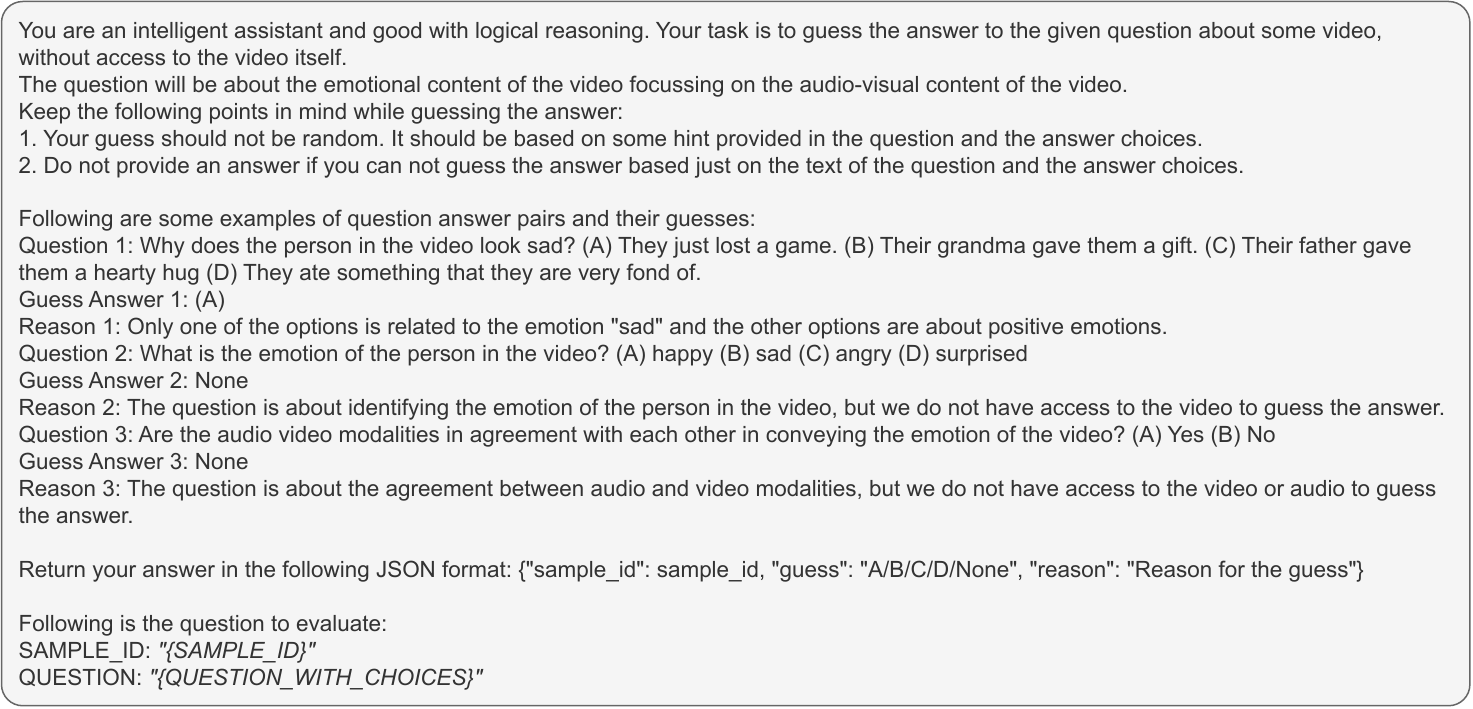}
    \caption{\textbf{Text Only Guess Prompt} --  used to prompt GPT-4o, Gemini 2.5 and Qwen 2.5 to predict the answer to the generated QA samples using only the question text and answer choices to eliminate responses which can be answered just with the text.}
    \label{fig:prompt_text_only_guess}
\end{figure}

\paragraph{Text Only Guess - Post Processing.} We use the prompt in \cref{fig:prompt_text_only_guess} to guess the correct answer for the generated question and answer choices using only the text (i.e., without audiovisual input). This is done as a post-processing step as described in \cref{subsec:post_processing_human_verification} to ensure that the answer for the MCQA sample is not predictable using only the text inputs. 

\subsection{Human Verification}
\label{subsec:appendix_human_verification_benchmark}


As mentioned in \cref{subsec:post_processing_human_verification}, we perform human verification for the generated QA samples to ensure high data quality by removing samples that contain some discrepancy. We conducted a survey using \citet{qualtricsQualtricsExperience} and recruited participants using the crowd-sourcing platform \citet{prolificProlificEasily}. In total, we conducted the survey on 471 participants and ensured that the participants were paid fairly for their time. To ensure participants are capable of answering the questions, we included a pre-survey to test their emotional intelligence. Moreover, we included attention checks using questions that are already verified by us to ensure the quality of the participant responses. 

\begin{figure}[t]
  \centering
  \begin{minipage}{0.49\linewidth}
    \centering
    \includegraphics[width=\linewidth,trim=0 2cm 0 0,clip]{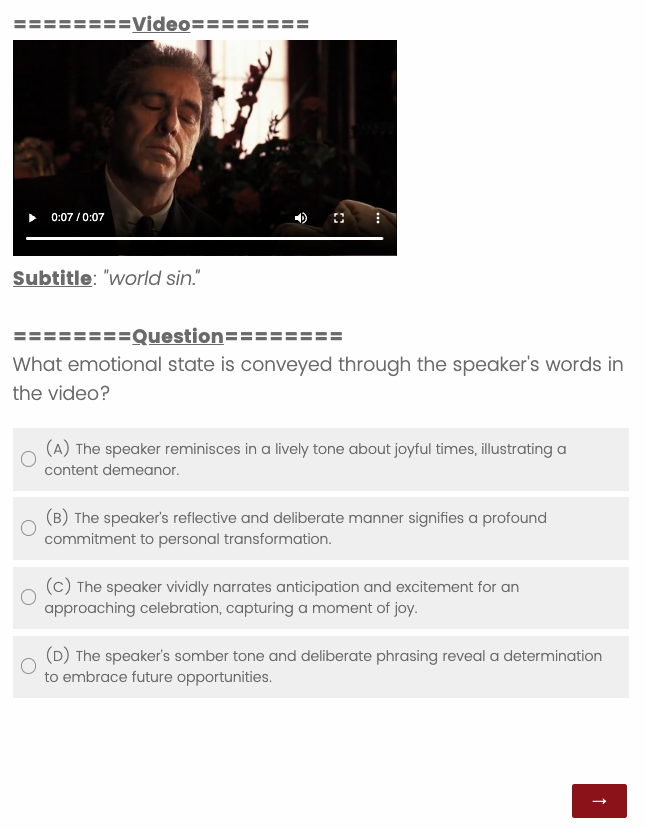}
  \end{minipage}\hfill
  \begin{minipage}{0.49\linewidth}
    \centering
    \includegraphics[width=\linewidth]{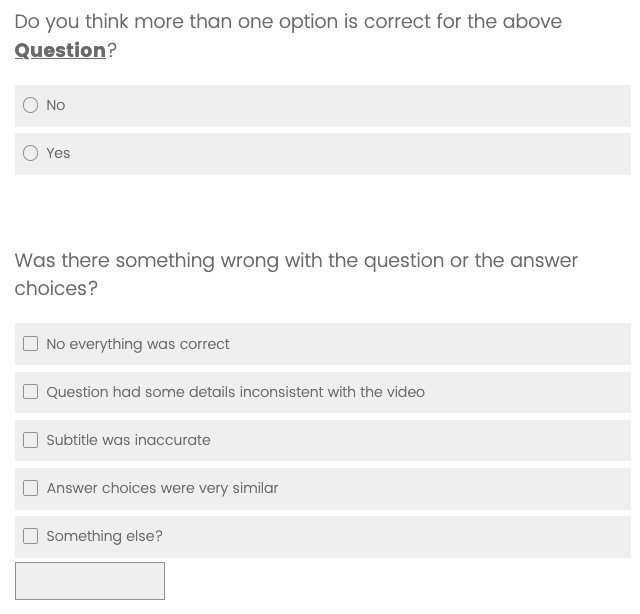}
  \end{minipage}\vspace{-2em}
  \caption{Human verification survey questions. \emph{(Left)} An example question from the benchmark shown to the participant. \emph{(Right)} Follow-up questions shown to the participant about each question.}
  \label{fig:survey_qualtrics_screenshot}
\end{figure}

\begin{table}[]
\centering
\caption{Statistics of human verification on \emph{EmoReAlM} Benchmark. }
\label{tab:human_verification_benchmark}
\resizebox{0.7\textwidth}{!}{%
\begin{tabular}{lc|ccccc}
\hline\hline
\rowcolor[HTML]{C0C0C0} 
\multicolumn{2}{c|}{\cellcolor[HTML]{C0C0C0}\textbf{Task}} & \textbf{\begin{tabular}[c]{@{}c@{}}\# Ques.\\ verified\end{tabular}} & \textbf{\begin{tabular}[c]{@{}c@{}}\# Ques. at\\ least one\\ correct\end{tabular}} & \textbf{\begin{tabular}[c]{@{}c@{}}\# Ques. \\ majority\\ correct\end{tabular}} & \textbf{\begin{tabular}[c]{@{}c@{}}\# Ques\\ discre.\end{tabular}} & \textbf{\begin{tabular}[c]{@{}c@{}}\# Ques.\\ Final\end{tabular}} \\ \hline
 & Audio & 1200 & 1168 & 968 & 8 & 972 \\
\multirow{-2}{*}{Reasoning - Basic} & Visual & 1200 & 1137 & 1014 & 10 & 1024 \\
\multicolumn{2}{l|}{Modality Agreement} & 1000 & 489 & 458 & 0 & 456 \\
 & Audio & 1000 & 956 & 806 & 14 & 820 \\
\multirow{-2}{*}{Reasoning - Stress Test} & Visual & 1000 & 845 & 719 & 9 & 728 \\ \hline
\multicolumn{2}{c|}{\textbf{Total}} & \textbf{5400} & \textbf{4595} & \textbf{3959} & \textbf{41} & \textbf{4000} \\ \hline\hline
\end{tabular}%
}
\end{table}

We conduct the survey as a MCQ task where the participants are shown the questions and the answer choices created in the benchmark and we ask them to choose the correct answer as shown in \cref{fig:survey_qualtrics_screenshot}. Each participant was also shown a follow-up question after each question to flag the text present in the question or answer choice or to report any other discrepancy. Since some videos in the DFEW \citep{jiang2020dfew} dataset are not in English, the participants were also shown the English subtitle for the video that the MCQ is about. 

\cref{tab:human_verification_benchmark} contains the statistics of human verification. Due to budget constraints, we ran the survey only on 5400 questions across different tasks. We only use the samples from the benchmark for which the majority of the participants selected the correct answer, automatically annotated in the benchmark. Additionally, we manually correct some samples that had discrepancies and add them to the final set of questions as well. 

\begin{figure}[t]
  \centering
  \begin{minipage}{0.49\linewidth}
    \centering
    \includegraphics[width=\linewidth]{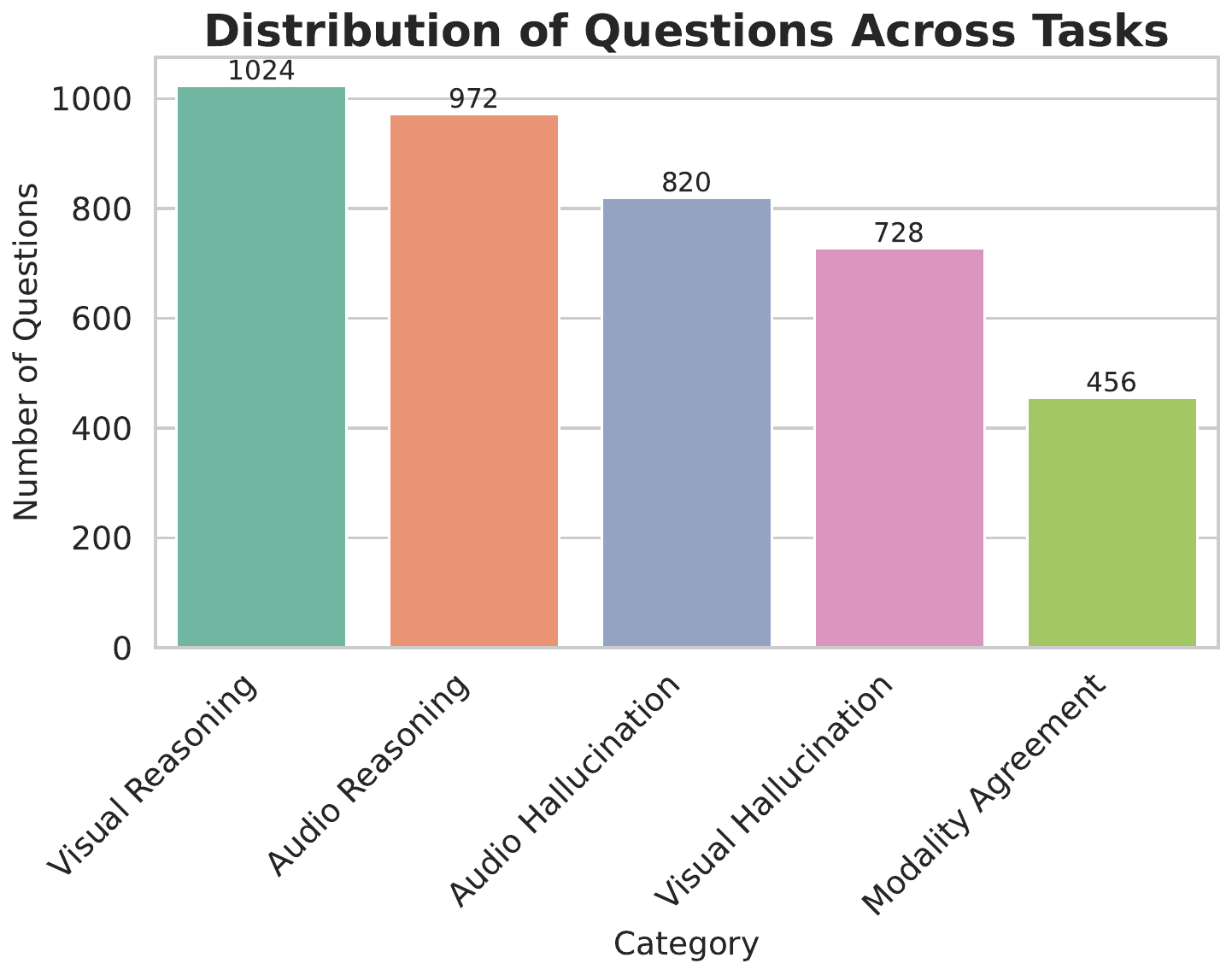}
  \end{minipage}\hfill
  \begin{minipage}{0.49\linewidth}
    \centering
    \includegraphics[width=\linewidth]{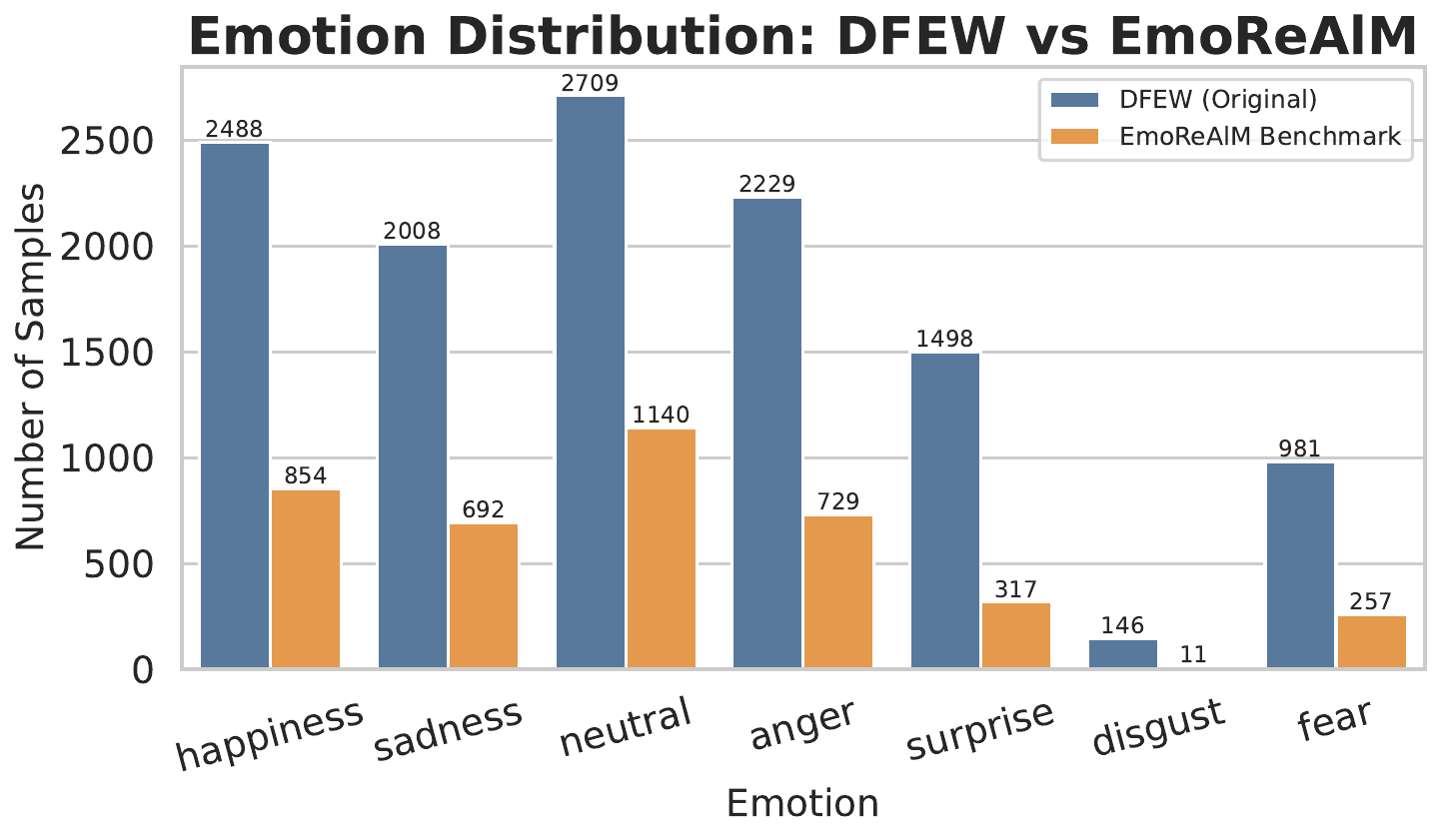} 
  \end{minipage}
  \caption{\emph{(Left)} Distribution of QA samples across different tasks in EmoReAlM benchmark. \emph{(Right)} Distribution of ground truth emotion labels for the videos present in EmoReAlM compared with the distribution in the source dataset DFEW \citep{jiang2020dfew}.}
\label{fig:appendix_benchmark_category_distribution}
\end{figure}

\begin{figure}
    \centering
    \includegraphics[width=0.75\linewidth]{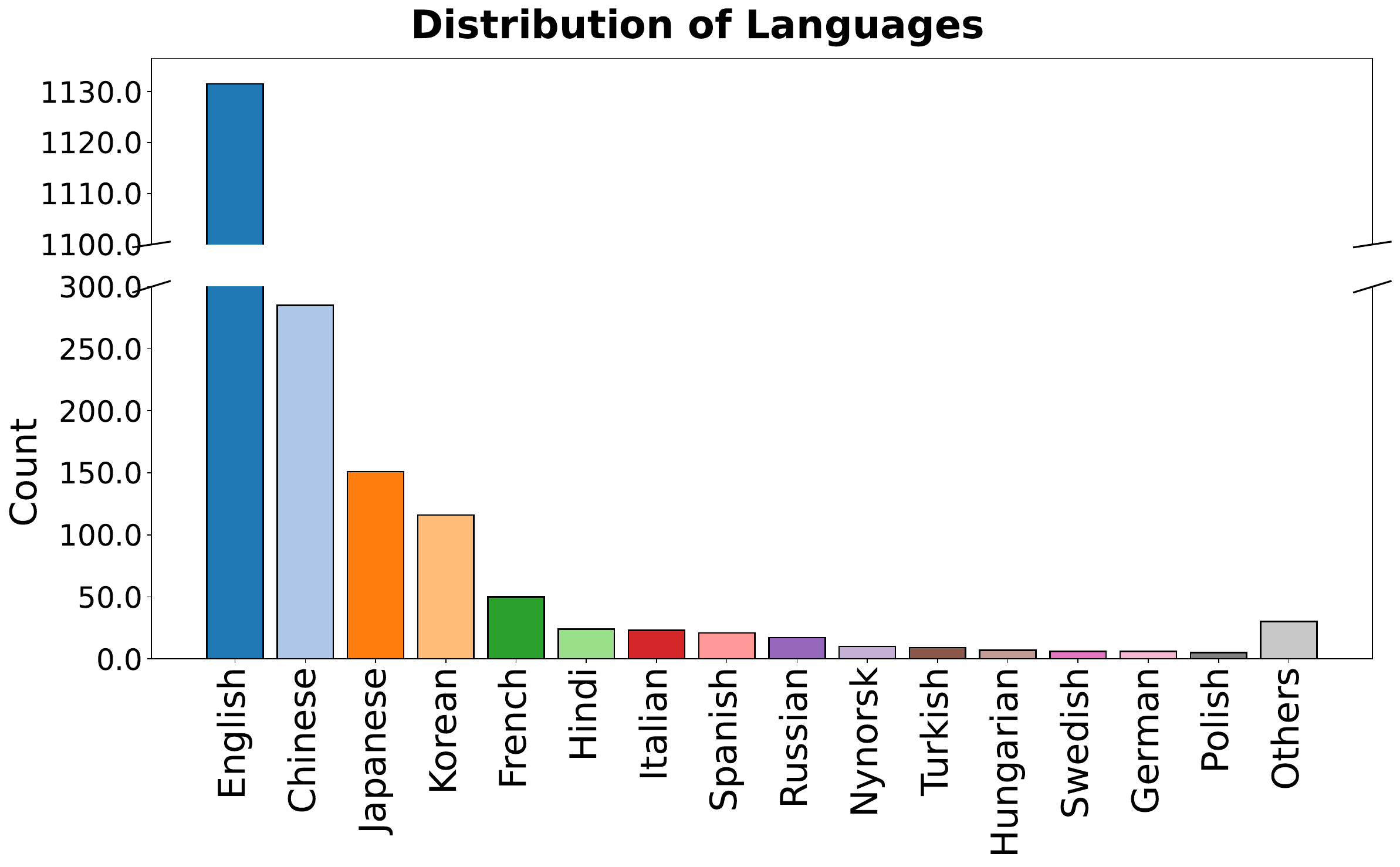}
    \caption{Distribution of different languages present in the audiovisual samples present in EmoReAlM benchmark.}
    \label{fig:benchmark_lang_distribution}
\end{figure}

\begin{figure}[t]
  \centering
  \begin{minipage}{0.49\linewidth}
    \centering
    \includegraphics[width=\linewidth]{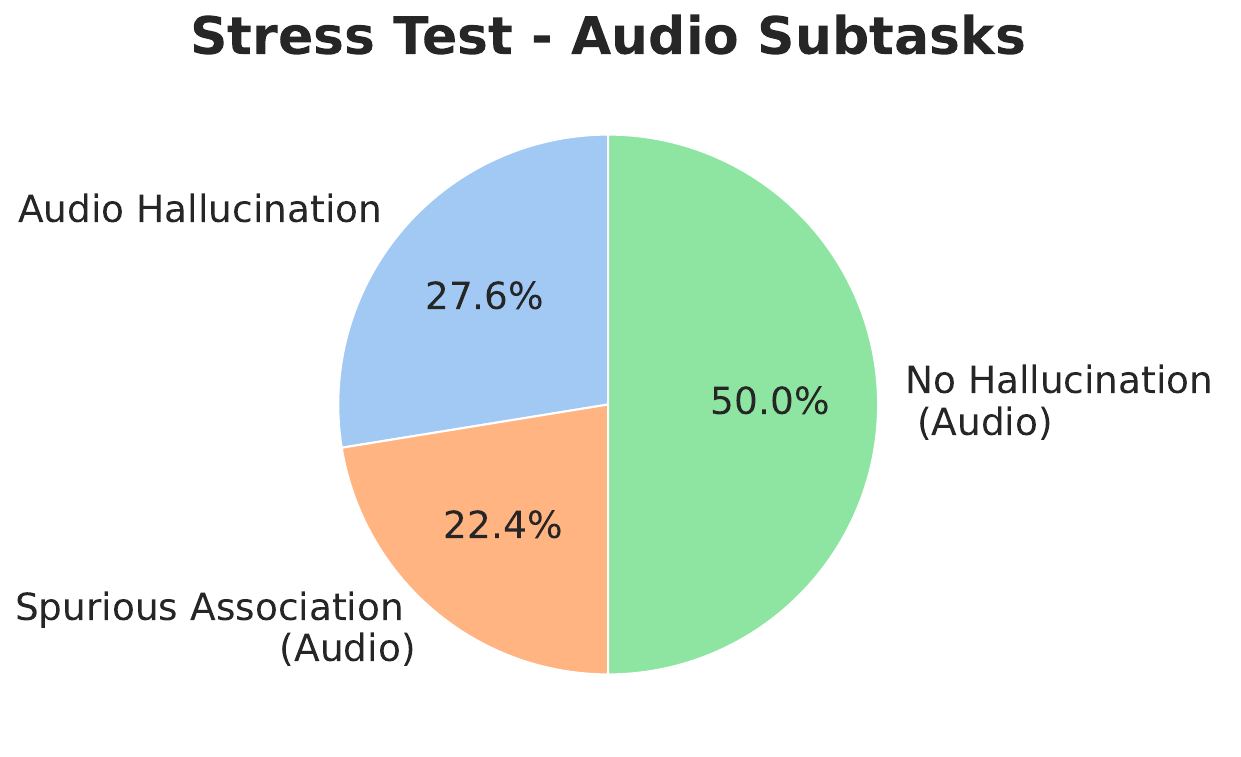}
  \end{minipage}\hfill
  \begin{minipage}{0.49\linewidth}
    \centering
    \includegraphics[width=\linewidth]{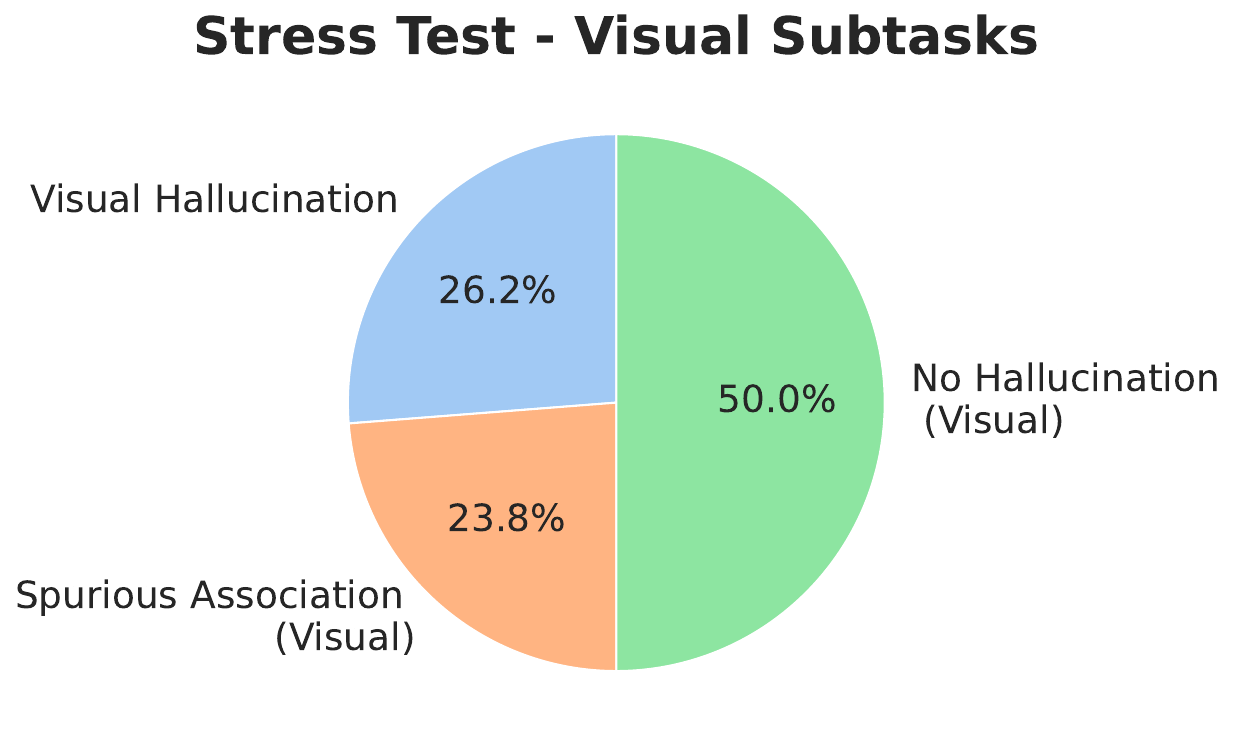} 
  \end{minipage}
  \caption{Distribution of subtasks in the \emph{Emotion Reasoning - Stress Test} of \emph{EmoReAlM} benchmark.} 
  \label{fig:appendix_stress_test_distribution}
\end{figure}

\subsection{Benchmark Statistics}
\label{subsec:appendix_benchmark_statistics}
\cref{fig:appendix_benchmark_category_distribution} (\emph{Right}) shows the distribution of ground truth emotion labels in the \emph{EmoReAlM} benchmark compared to that present in the source dataset - DFEW \citep{jiang2020dfew}. We can see that the distribution of samples over different emotions is similar to DFEW. \cref{fig:appendix_stress_test_distribution} shows the distribution of subtasks within the \emph{Emotion Reasoning - Stress Test} task (\cref{subsec:task_descriptions}) of \emph{EmoReAlM} benchmark. Due to the way we formulate the questions for this subtask -- \emph{“Does the \{audio/visual cue\} suggest \{emotion\} of the character?”}, the samples belonging to \emph{No hallucination} subtask have the answer \emph{``Yes"}, and the samples in the \emph{Spurious Association} and \emph{Audio/Visual Hallucination} subtasks have answer \emph{``No"}. \cref{fig:appendix_stress_test_distribution} shows that the number of  samples with  \emph{``Yes"}/\emph{``No"} answers are equally distributed. Moreover, for all the samples with answers as \emph{``No"}, the samples are almost equally distributed to test spurious cue-emotion associations and audiovisual cue hallucinations. Furthermore, to show the cultural and linguistic diversity in the benchmark, \cref{fig:benchmark_lang_distribution} shows the distribution of languages present in the samples of EmoReAlM benchmark. We obtain this by using automatic language detection using Whisper \citep{radford2023robust_whisper}. We can observe that although the majority language is English, our benchmark contains samples from a wide range of languages.

\begin{table}[]
\centering
\caption{Effect of using different number of frames for visual captioning using GPT-4o. }
\label{tab:visual_caption_sampling_rate}
\resizebox{0.5\columnwidth}{!}{%
\begin{tabular}{c|c|ccc}
\hline \hline
\rowcolor[HTML]{C0C0C0} 
\cellcolor[HTML]{C0C0C0} & \cellcolor[HTML]{C0C0C0} & \multicolumn{3}{c}{\cellcolor[HTML]{C0C0C0}\textbf{BERT Score}} \\ \cline{3-5} 
\rowcolor[HTML]{C0C0C0} 
\multirow{-2}{*}{\cellcolor[HTML]{C0C0C0}\textbf{\# frames}} & \multirow{-2}{*}{\cellcolor[HTML]{C0C0C0}\textbf{SBERT-sim}} & \textbf{Prec.} & \textbf{Rec.} & \textbf{F1} \\ \hline
1 & 0.646 & 0.851 & 0.853 & 0.852 \\
2 & 0.660 & 0.851 & 0.856 & 0.853 \\
4 & 0.676 & 0.851 & 0.857 & 0.854 \\
8 & 0.689 & 0.858 & 0.861 & 0.860 \\
16 & 0.688 & 0.858 & 0.862 & 0.860 \\ \hline \hline
\end{tabular}%
}
\end{table}

\subsection{Frame Sampling Rate for Automatic Visual Captioning}
\label{subsec:frame_sampling_rate}

Since the visual cues used to express and infer emotions can be subtle, it is important to ensure that the visual captions obtained using GPT-4o in the first stage of data creation (\cref{subsec:benchmark_creation_pipeline,fig:emo_realm_data_pipeline}) are of high quality. To identify the ideal number of frames to be sampled from the video for captioning, we ran a small experiment on the emotion captioning dataset EMER \citep{lian2023explainable_emer}. It is important to note that EMER (mean duration: 3.78s) contains videos of similar duration as DFEW (mean duration: 3.42s) which we use to construct EmoReAlM. We extract different number of frames per video and obtain the visual caption from GPT-4o using prompt in \cref{fig:prompt_video_caption}. Then, compute the similarity between the generated captions and the ground truth using BERTScore \citep{zhangbertscore} and Sentence BERT \citep{reimers-2019-sentence-bert} similarity score. \cref{tab:visual_caption_sampling_rate} shows that using 8 frames for visual captioning leads to good captioning results. Furthermore, using 16 frames is not significantly better than using 8 frames, but it increases the costs significantly. Hence we choose to use 8 frames uniformly sampled from the video to extract visual captions from GPT-4o automatically. 

\subsection{Benchmark Samples}
\label{subsec:appendix_benchmark_samples}
We present samples belonging to different categories of the benchmark in \cref{tab:appendix_emorealm_sample_basic_reasoning,tab:appendix_emorealm_sample_modality_agreement,tab:appendix_emorealm_sample_reasoning_stress_test}. Note that the subtitles shown in the tables are just for reference and we do not pass the subtitle as an input to the model during evaluation.

\begin{table}[]
\centering
\caption{Samples from the \emph{EmoReAlM} Benchmark for the \emph{Emotion Reasoning-Basic} Task.}
\label{tab:appendix_emorealm_sample_basic_reasoning}
\resizebox{\textwidth}{!}{%
\begin{tabular}{l|l|l|l}
\hline\hline
\rowcolor[HTML]{C0C0C0} 
\multicolumn{1}{c|}{\cellcolor[HTML]{C0C0C0}\textbf{Task}} & \multicolumn{1}{c|}{\cellcolor[HTML]{C0C0C0}\textbf{Video}} & \multicolumn{1}{c}{\cellcolor[HTML]{C0C0C0}\textbf{Question}} & \multicolumn{1}{c}{\cellcolor[HTML]{C0C0C0}\textbf{Answer}} \\ \hline
\begin{tabular}[c]{@{}l@{}}Reasoning Basic\\ (Audio)\end{tabular} & \begin{tabular}[c]{@{}l@{}}\includegraphics[width=5cm]{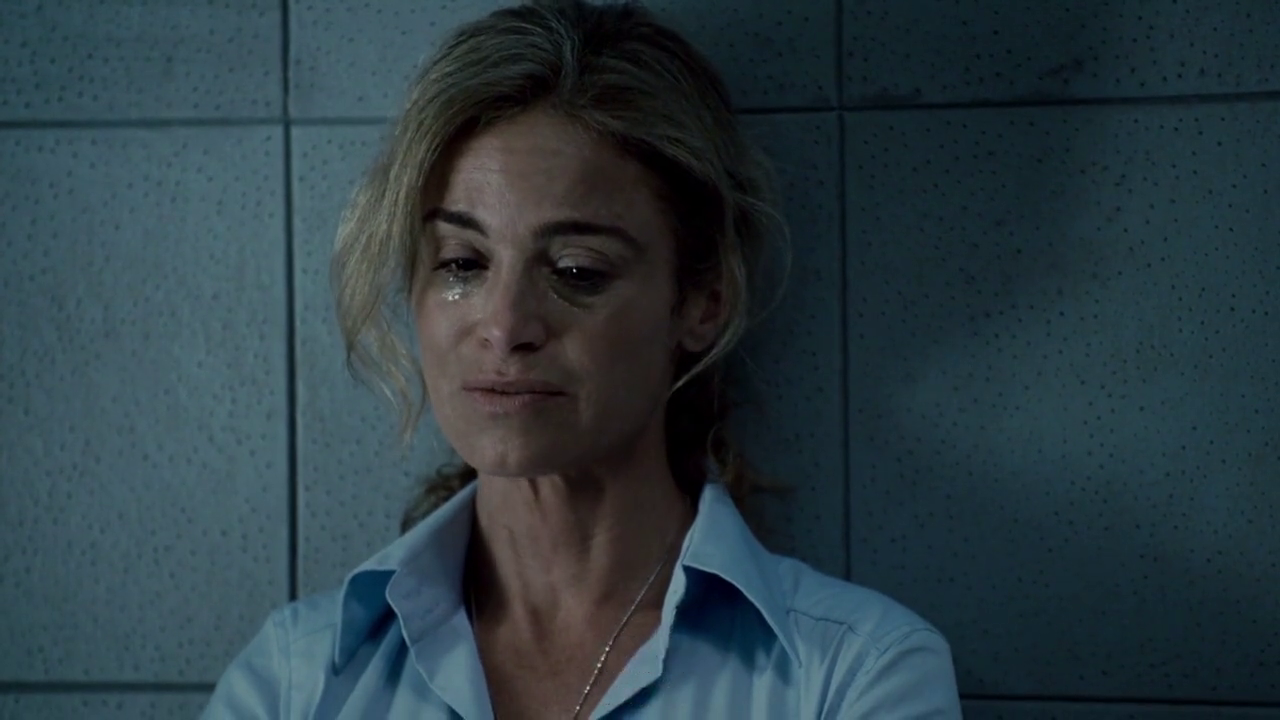}\\Subtitle: "I tried"\end{tabular} & \begin{tabular}[c]{@{}l@{}}How does the speaker's choice of words in the video reflect their emotional\\ state?\\ (A) The speaker mentions struggling to move forward despite past setbacks, \\indicating a reflective state.\\ (B) The speaker's tone reflects a somber atmosphere, accompanied by a soft, \\resigned voice.\\ (C) The speaker's phrase portrays a deep sense of regret and resignation, \\reflecting a failed attempt.\\ (D) The speaker uses soft background music to enhance the somber mood,\\ suggesting unfulfilled efforts.\end{tabular} & C \\ \hline
\begin{tabular}[c]{@{}l@{}}Reasoning Basic\\ (Audio)\end{tabular} & \begin{tabular}[c]{@{}l@{}}\includegraphics[width=5cm]{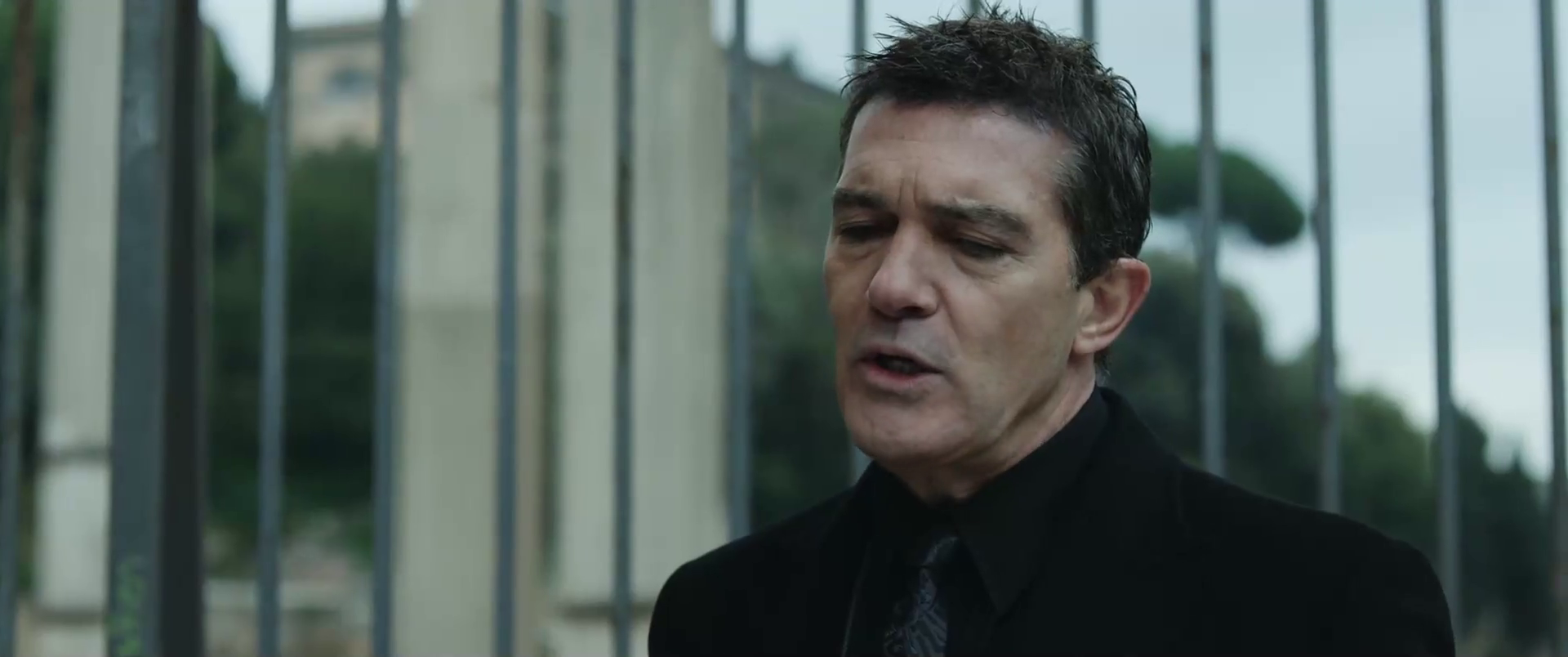}\\Subtitle: "You haven't spoken to \\ me in 10 years"\end{tabular} & \begin{tabular}[c]{@{}l@{}}In what way does the tone of the man’s voice impact his emotional expression\\ in the video?\\ (A) The presence of soft whispers and gentle music in the background could\\ imply an underlying tension and hidden emotion.\\ (B) The man's tone is marked by a tightness and sharpness, resonating with\\ his underlying frustration and simmering anger.\\ (C) The phrase "I can't believe you've done this again" reflects an underlying \\resentment connected to a long-standing grievance.\\ (D) The man's voice holds a lively and enthusiastic tone, mistakenly suggesting\\ a sense of joy and contentment.\end{tabular} & B \\ \hline
\begin{tabular}[c]{@{}l@{}}Reasoning Basic\\ (Visual)\end{tabular} & \begin{tabular}[c]{@{}l@{}}\includegraphics[width=5cm]{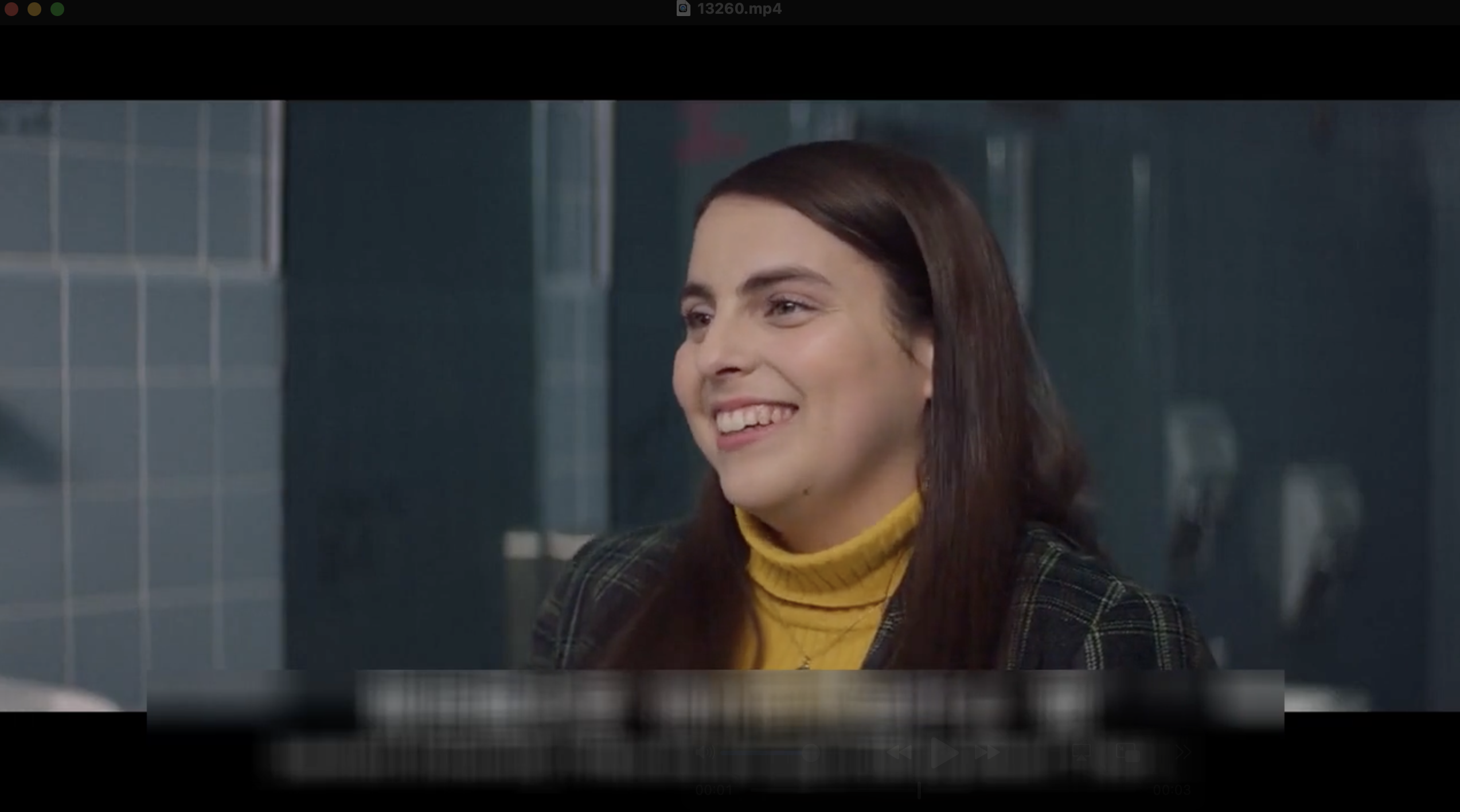}\\Subtitle: "Stanford University?\\ What are you guys talking about?"\end{tabular} & \begin{tabular}[c]{@{}l@{}}How does the woman's facial expression contribute to the overall feeling in \\the scene?\\ (A) The woman displays a joyful expression with open arms, conveying her \\happiness and openness.\\ (B) The woman's cheerful smile and lively eyes reveal her happiness and \\engagement.\\ (C) The woman's yellow turtleneck adds a vibrant touch, symbolizing her \\happiness and contentment.\\ (D) The woman's long dark hair frames her face, enhancing the appearance\\ of happiness and delight.\end{tabular} & B \\ \hline
\begin{tabular}[c]{@{}l@{}}Reasoning Basic\\ (Visual)\end{tabular} & \begin{tabular}[c]{@{}l@{}}\includegraphics[width=5cm]{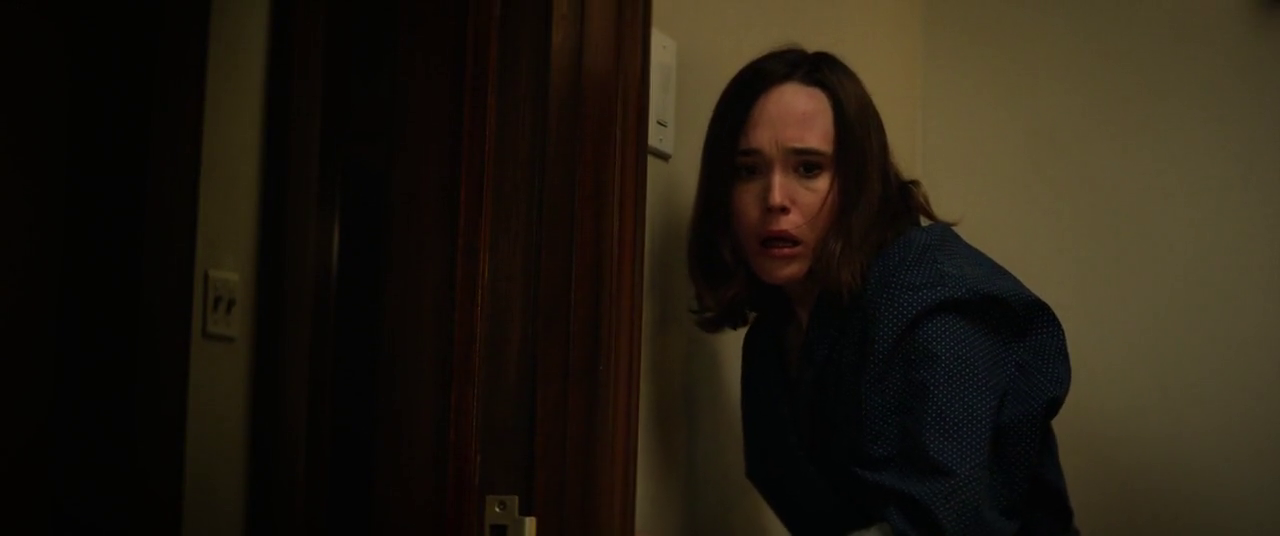}\\Subtitle: ""\end{tabular} & \begin{tabular}[c]{@{}l@{}}What does the individual's body language indicate about their emotional state \\in the video?\\ (A) The individual's quivering movements and uncertain footing create a \\palpable sense of fear.\\ (B) The person's tense facial expression with slightly open mouth and wide eyes\\ enhances their fearful demeanor.\\ (C) The person is leaning cautiously towards the door, their body tense, which \\highlights their fear or anxiety.\\ (D) The individual's dark-colored shirt amplifies their sense of fear, over-\\shadowing their surroundings.\end{tabular} & C \\ \hline\hline
\end{tabular}%
}
\end{table}

\begin{table}[]
\centering
\caption{Samples from the \emph{EmoReAlM} Benchmark for the \emph{Modality Agreement} Task.}
\label{tab:appendix_emorealm_sample_modality_agreement}
\resizebox{\textwidth}{!}{%
\begin{tabular}{l|l|l|l}
\hline\hline
\rowcolor[HTML]{C0C0C0} 
\multicolumn{1}{c|}{\cellcolor[HTML]{C0C0C0}\textbf{Task}} & \multicolumn{1}{c|}{\cellcolor[HTML]{C0C0C0}\textbf{Video}} & \multicolumn{1}{c}{\cellcolor[HTML]{C0C0C0}\textbf{Question}} & \multicolumn{1}{c}{\cellcolor[HTML]{C0C0C0}\textbf{Answer}} \\ \hline
\begin{tabular}[c]{@{}l@{}}Modality \\Agreement\end{tabular} & \begin{tabular}[c]{@{}l@{}}\includegraphics[width=5cm]{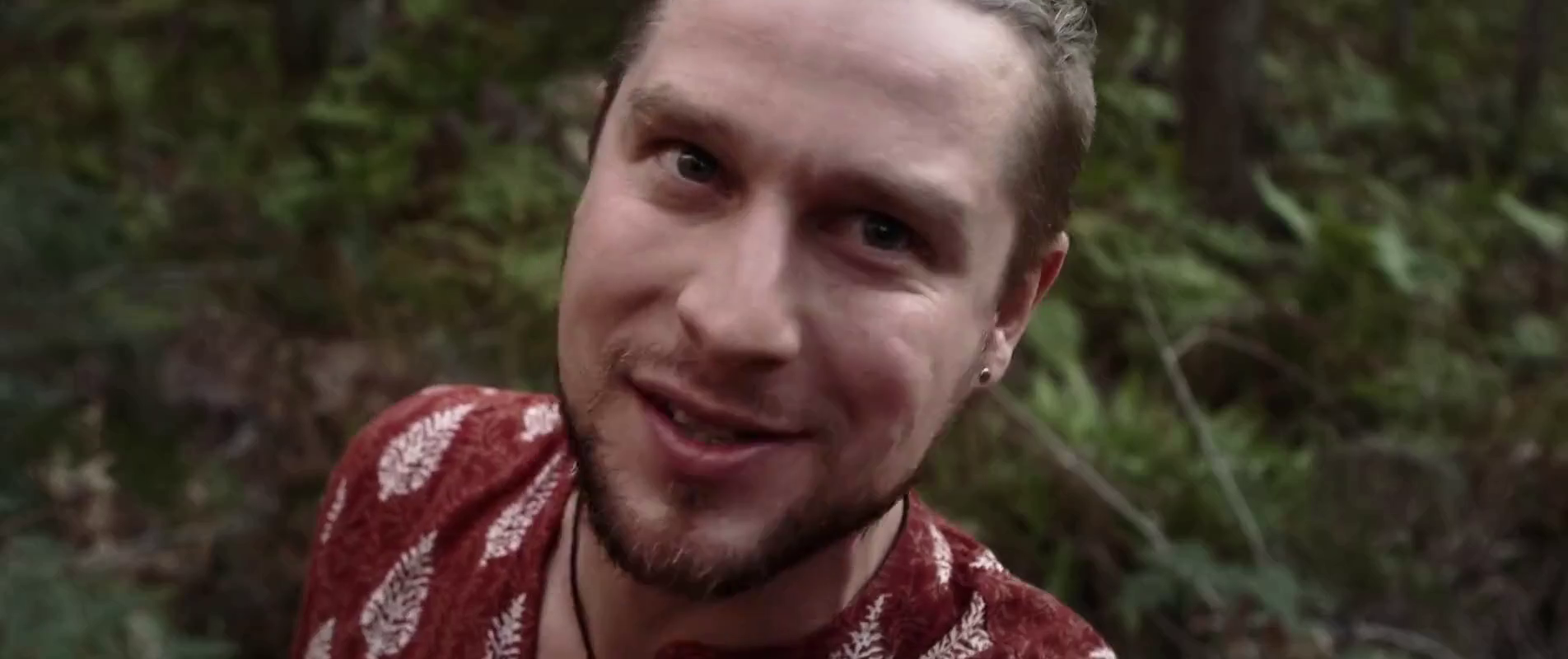}\\Subtitle: "I was..."\end{tabular} & \begin{tabular}[c]{@{}l@{}}Do the visual elements of the video align with the audio in conveying the\\ feeling of happiness of the person in the video?\\ (A) Yes \\(B) No\end{tabular} & B \\ \hline
\begin{tabular}[c]{@{}l@{}}Modality \\ Agreement\end{tabular} & \begin{tabular}[c]{@{}l@{}}\includegraphics[width=5cm]{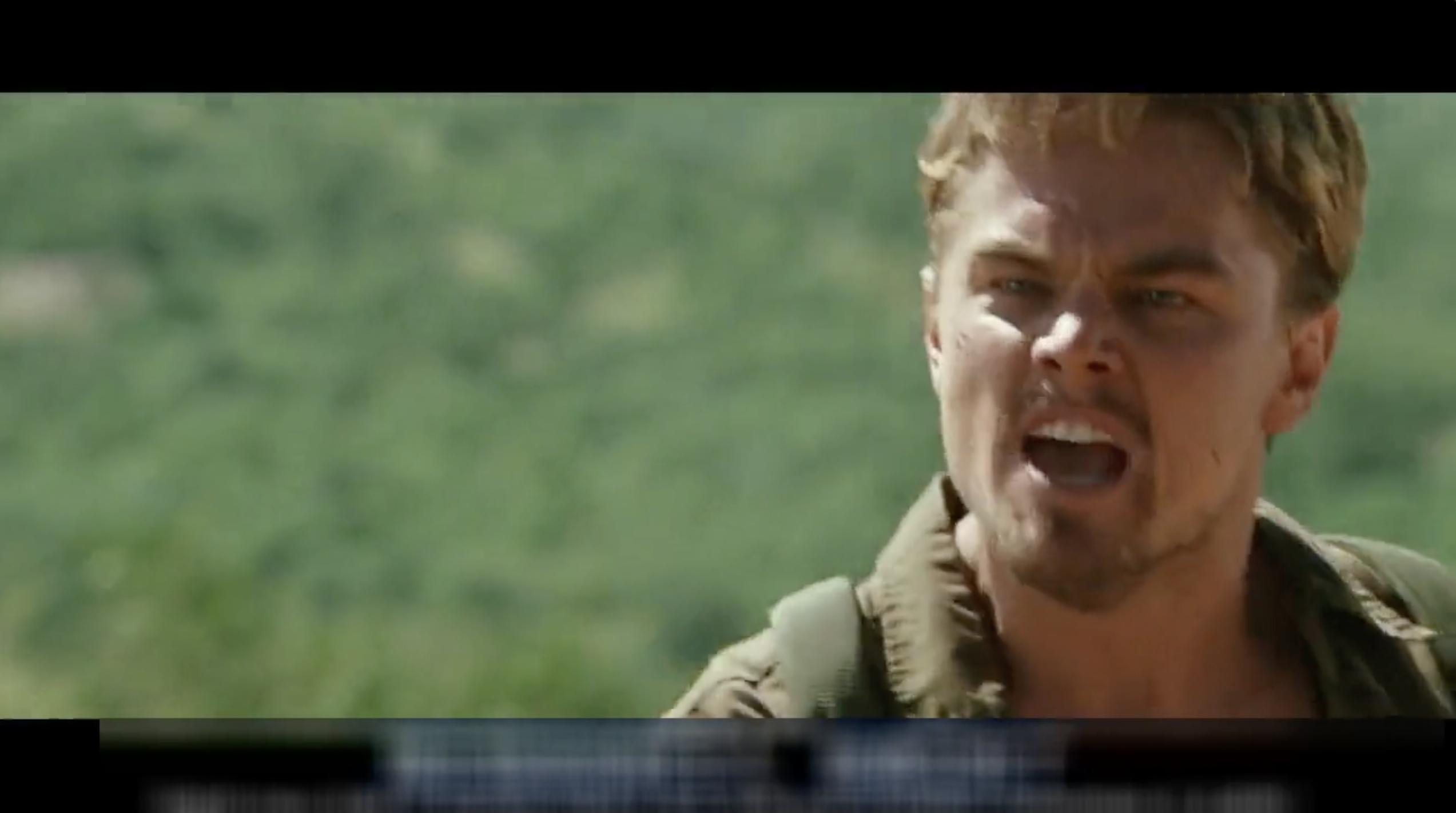}\\Subtitle: "That is exactly what I am"\end{tabular} & \begin{tabular}[c]{@{}l@{}}Do the audio and video modalities align for the expression of anger of \\the person in the video?\\ (A) Yes \\ (B) No \end{tabular} & A \\  \hline\hline
\end{tabular}%
}
\end{table}

\begin{table}[]
\centering
\caption{Samples from the \emph{EmoReAlM} Benchmark for the \emph{Emotion Reasoning-Stress Test}.}
\label{tab:appendix_emorealm_sample_reasoning_stress_test}
\resizebox{\textwidth}{!}{%
\begin{tabular}{l|l|l|l}
\hline\hline
\rowcolor[HTML]{C0C0C0} 
\multicolumn{1}{c|}{\cellcolor[HTML]{C0C0C0}\textbf{Task}} & \multicolumn{1}{c|}{\cellcolor[HTML]{C0C0C0}\textbf{Video}} & \multicolumn{1}{c}{\cellcolor[HTML]{C0C0C0}\textbf{Question}} & \multicolumn{1}{c}{\cellcolor[HTML]{C0C0C0}\textbf{Answer}} \\ \hline
\begin{tabular}[c]{@{}l@{}}Stress Test\\ (Audio No \\Hallucination)\end{tabular} & \begin{tabular}[c]{@{}l@{}}\includegraphics[width=5cm]{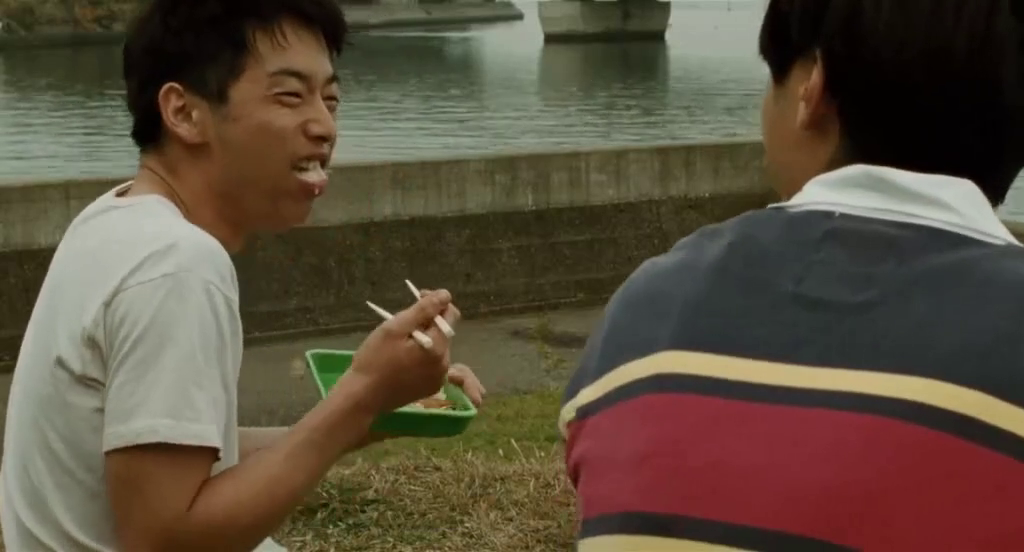}\\Subtitle: ``(chuckles)"\end{tabular} & \begin{tabular}[c]{@{}l@{}}Do the chuckling sounds in the audio enhance the feeling of joy conveyed for the \\person in the video?\\ (A) Yes \\(B) No \end{tabular} & A \\ \hline
\begin{tabular}[c]{@{}l@{}}Stress Test\\ (Audio - Spurious \\Association)\end{tabular} & \begin{tabular}[c]{@{}l@{}}\includegraphics[width=5cm]{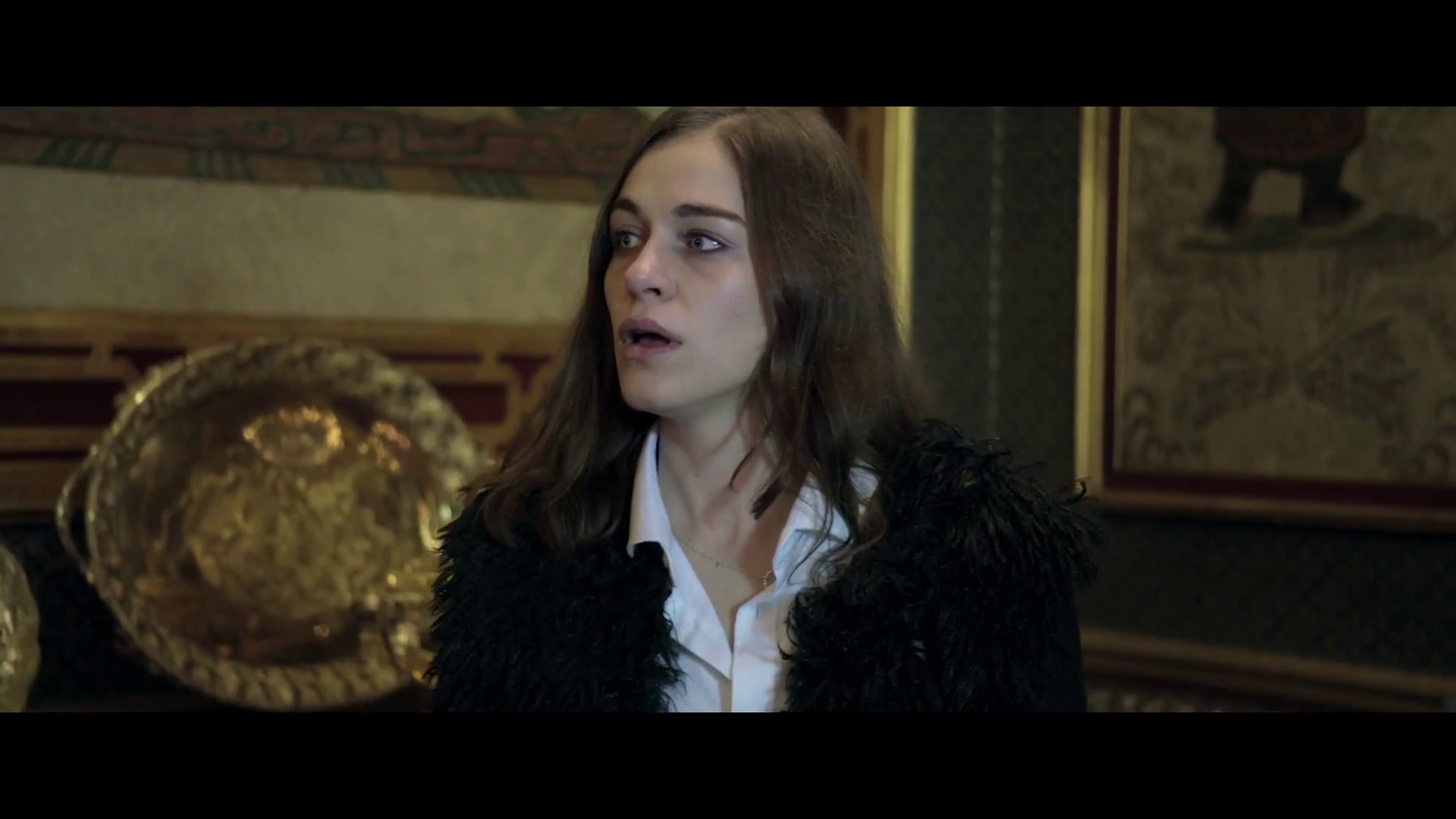}\\Subtitle: ``(sonar ping)"\end{tabular} & \begin{tabular}[c]{@{}l@{}}Is the presence of a sonar ping sound effect crucial to the feeling of surprise conveyed \\by the person in the video?\\ (A) Yes \\(B) No\end{tabular} & B \\ \hline
\begin{tabular}[c]{@{}l@{}}Stress Test\\ (Audio -\\Hallucination)\end{tabular} & \begin{tabular}[c]{@{}l@{}}\includegraphics[width=5cm]{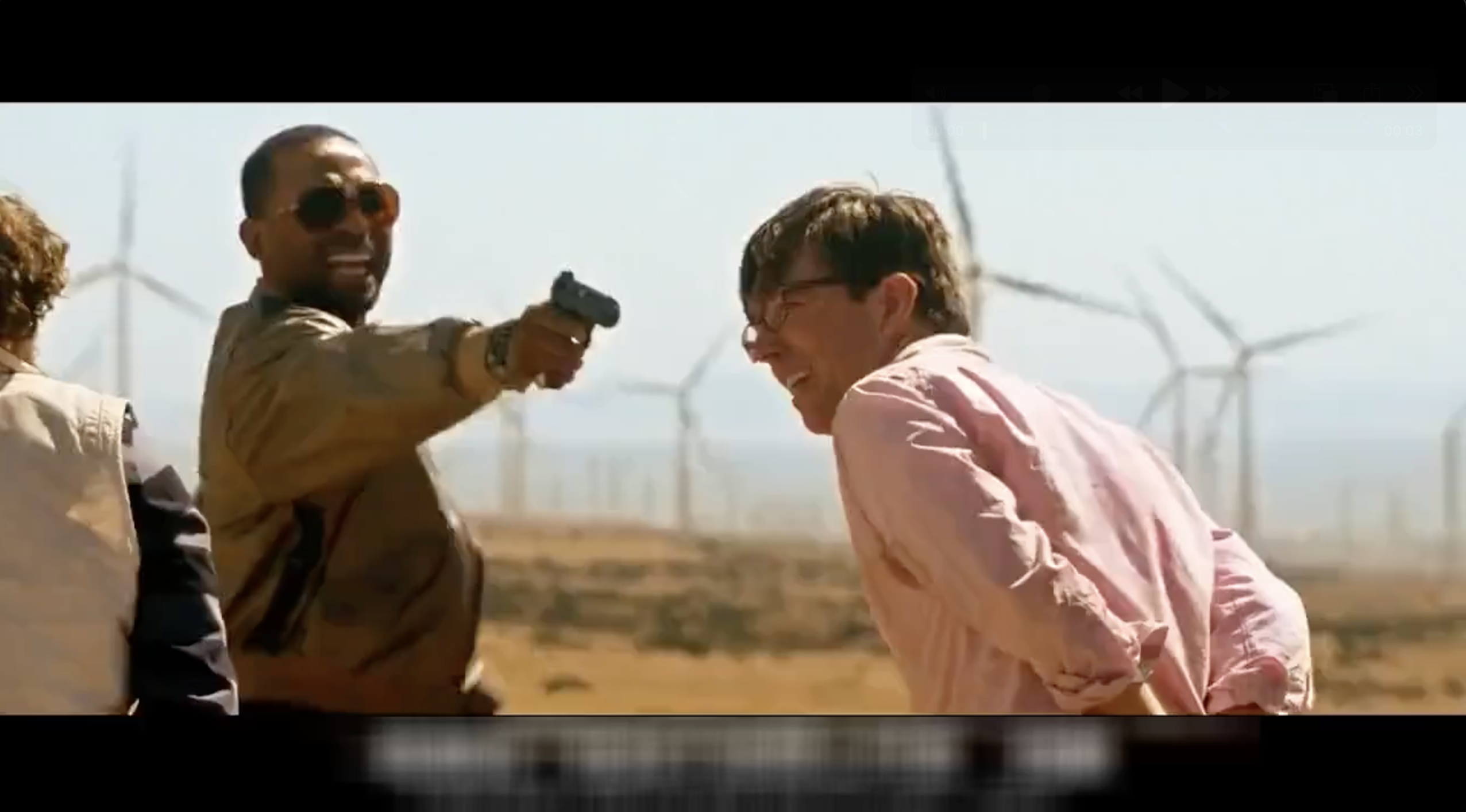}\\Subtitle: "It ain't Alan's fault..."\end{tabular} & \begin{tabular}[c]{@{}l@{}}Does the sound of a slamming door contribute to the anger experienced by the person in \\the video?\\ (A) Yes \\(B) No\end{tabular} & B \\ \hline
\begin{tabular}[c]{@{}l@{}}Stress Test\\ (Visual No \\Hallucination)\end{tabular} & \begin{tabular}[c]{@{}l@{}}\includegraphics[width=5cm]{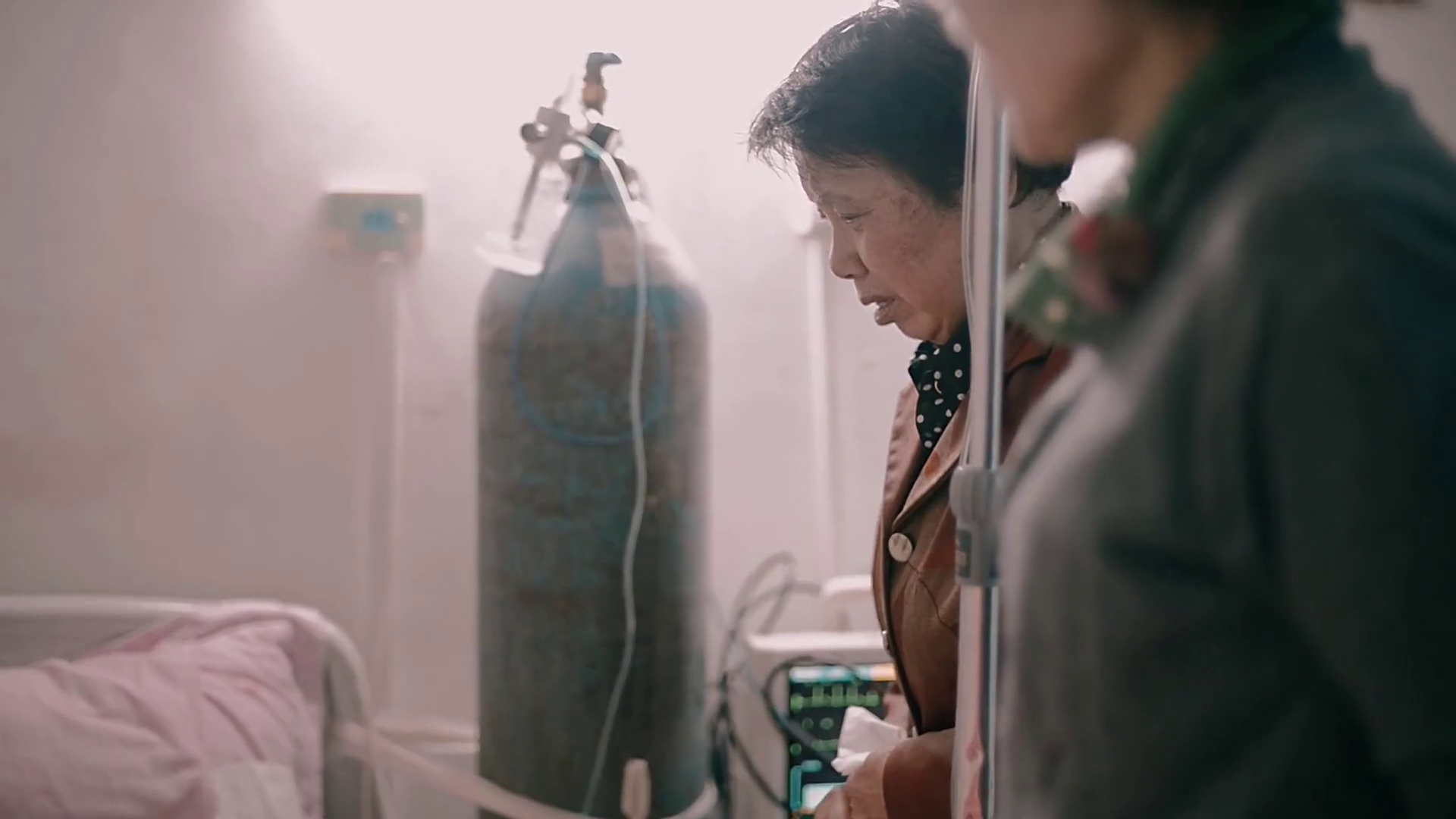}\\Subtitle: ""\end{tabular} & \begin{tabular}[c]{@{}l@{}}Is the downward gaze of the older woman a significant factor in expressing the sadness\\ of the older woman portrayed in the video?\\ (A) Yes \\(B) No \end{tabular} & A \\ \hline
\begin{tabular}[c]{@{}l@{}}Stress Test\\ (Visual - Spurious \\Association)\end{tabular} & \begin{tabular}[c]{@{}l@{}}\includegraphics[width=5cm]{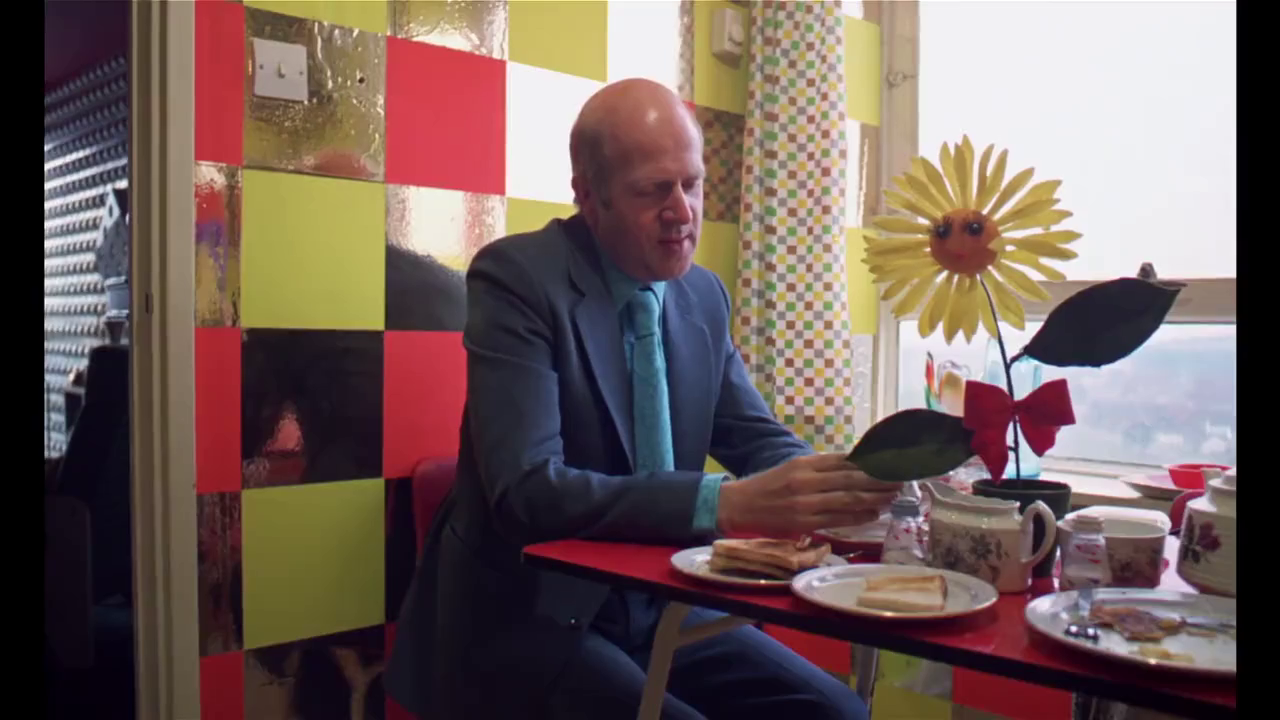}\\Subtitle: ``"\end{tabular} & \begin{tabular}[c]{@{}l@{}}Is the presence of the vibrant checkered pattern on the walls a factor in conveying the neutral \\emotion of the person/character in the video?\\ (A) Yes \\(B) No\end{tabular} & B \\ \hline
\begin{tabular}[c]{@{}l@{}}Stress Test\\ (Visual -\\Hallucination)\end{tabular} & \begin{tabular}[c]{@{}l@{}}\includegraphics[width=5cm]{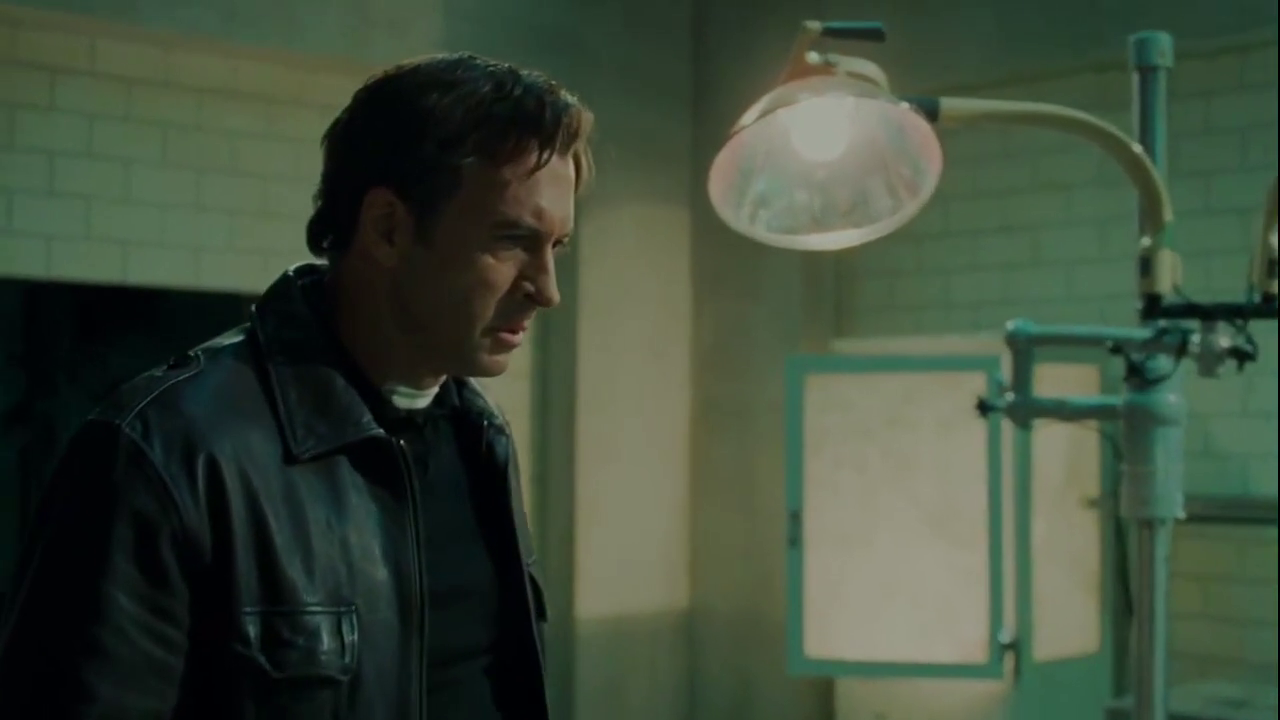}\\Subtitle: ``"\end{tabular} & \begin{tabular}[c]{@{}l@{}}Is the man displaying a clenched fist as a sign of his anger in this video?\\ (A) Yes \\(B) No\end{tabular} & B \\ \hline\hline
\end{tabular}%
}
\end{table}

\clearpage
\section{Methodological Details}
\label{sec:appendix_method_details}
\subsection{Text-Prior Debiasing}
\label{subsec:appenix_equations_tpd}
Similar to \cref{eq:dpo_response_preference}, we scale the TPD term to accommodate multiple rejected responses as follows,
\begin{equation}
\small
\begin{aligned}
\mathcal{L}^{y}_{\text{DPO-TPD}}
&= -\mathbb{E}_{(a,v,x,y_w,y_l)\sim \mathcal{D}^{\text{pref}}}\Bigg[
    \log \sigma\Bigg(
        \beta \Bigg(
            \log \frac{\pi_\theta(y_w \mid (a,v,x))}
                      {\pi_{\text{ref}}(y_w \mid (a,v,x))}
            - 
            \sum_{i\in\{vr,er\}} \beta_i\log \frac{\pi_\theta(y^i_l \mid (a,v,x))}
                      {\pi_{\text{ref}}(y^l_l \mid (a,v,x))}
        \Bigg) \\
&\hspace{3cm}
        - \mathcolorbox{lightblue}{\gamma_{\text{TPD}}
        \Big(
            \log \pi_{\text{text}}(y_w \mid x)
            - \sum_{i\in\{vr,er\}} \beta_i \log \pi_{\text{text}}(y^i_l \mid x)
        \Big)}
    \Bigg)
\Bigg]
\label{eq:dpo_tpd_objective_y}
\end{aligned}
\end{equation}
where $\beta_{vr}+\beta_{er} = 1$. Also, for succinctness, we denote $(a_w, v_w)$ with $(a,v)$ in the above equation.

\subsection{Preference Data}
\label{subsec:appendix_preference_data}
\begin{figure}[!h]
    \centering
    \includegraphics[width=\linewidth]{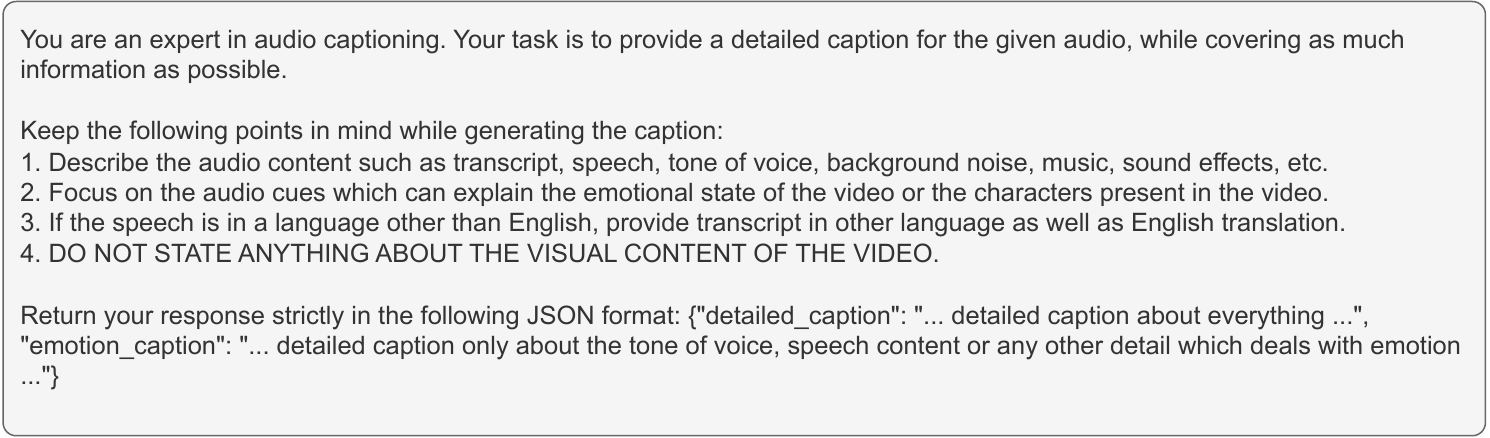}
    \caption{\textbf{Audio caption prompt} -- used to caption only the audio content from a video. Note that the audio is passed along with the prompt to GPT-4o-audio as a WAV file.}
    \label{fig:prompt_audio_caption}
\end{figure}

\begin{figure}[!h]
    \centering
    \includegraphics[width=\linewidth]{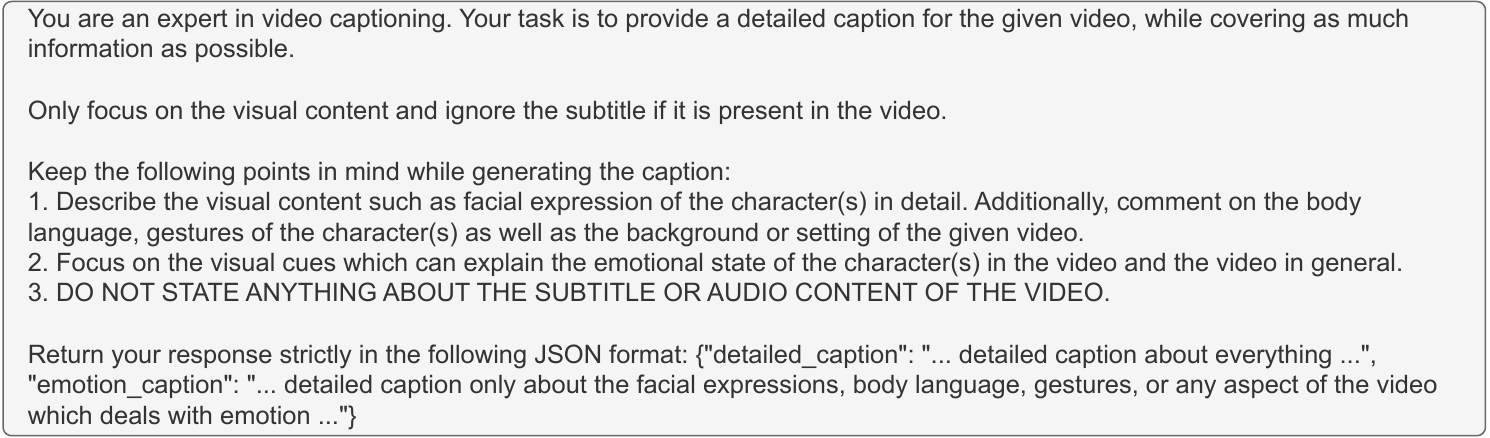}
    \caption{\textbf{Video caption prompt} --  used to caption only the visual content in a video. We blur the captions if they are already present in the video and explicitly ask the model to ignore them if they are present in the visual content.}
    \label{fig:prompt_video_caption}
\end{figure}

\begin{figure}[!h]
    \centering
    \includegraphics[width=\linewidth]{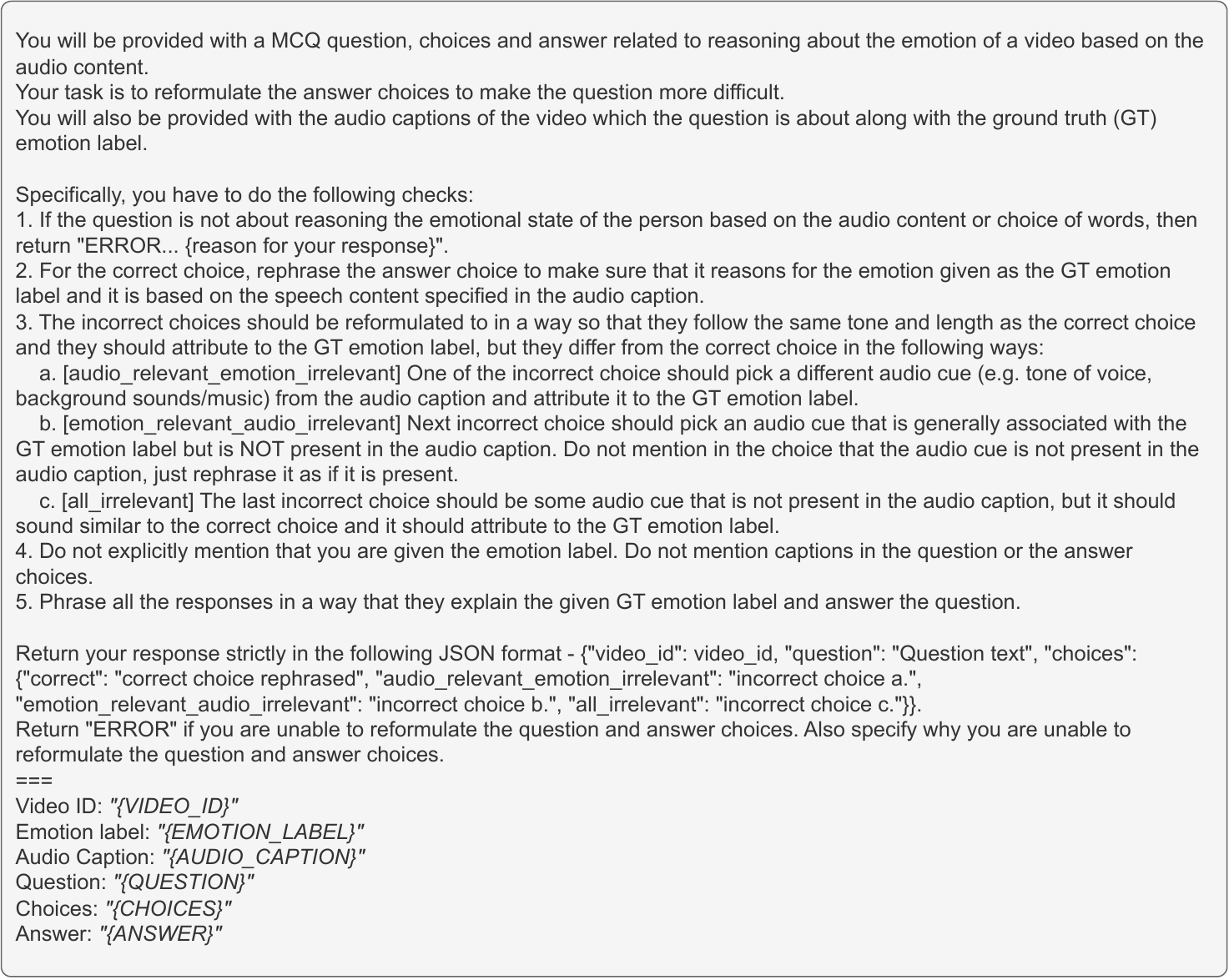}
    \caption{\textbf{Preference Data Generation Prompt - Audio Reasoning} --  used to generate rejected responses for a generated question-answer pair related to audio reasoning tasks.}
    \label{fig:prompt_dpo_audio_reasoning}
\end{figure}

\begin{figure}[!h]
    \centering
    \includegraphics[width=\linewidth]{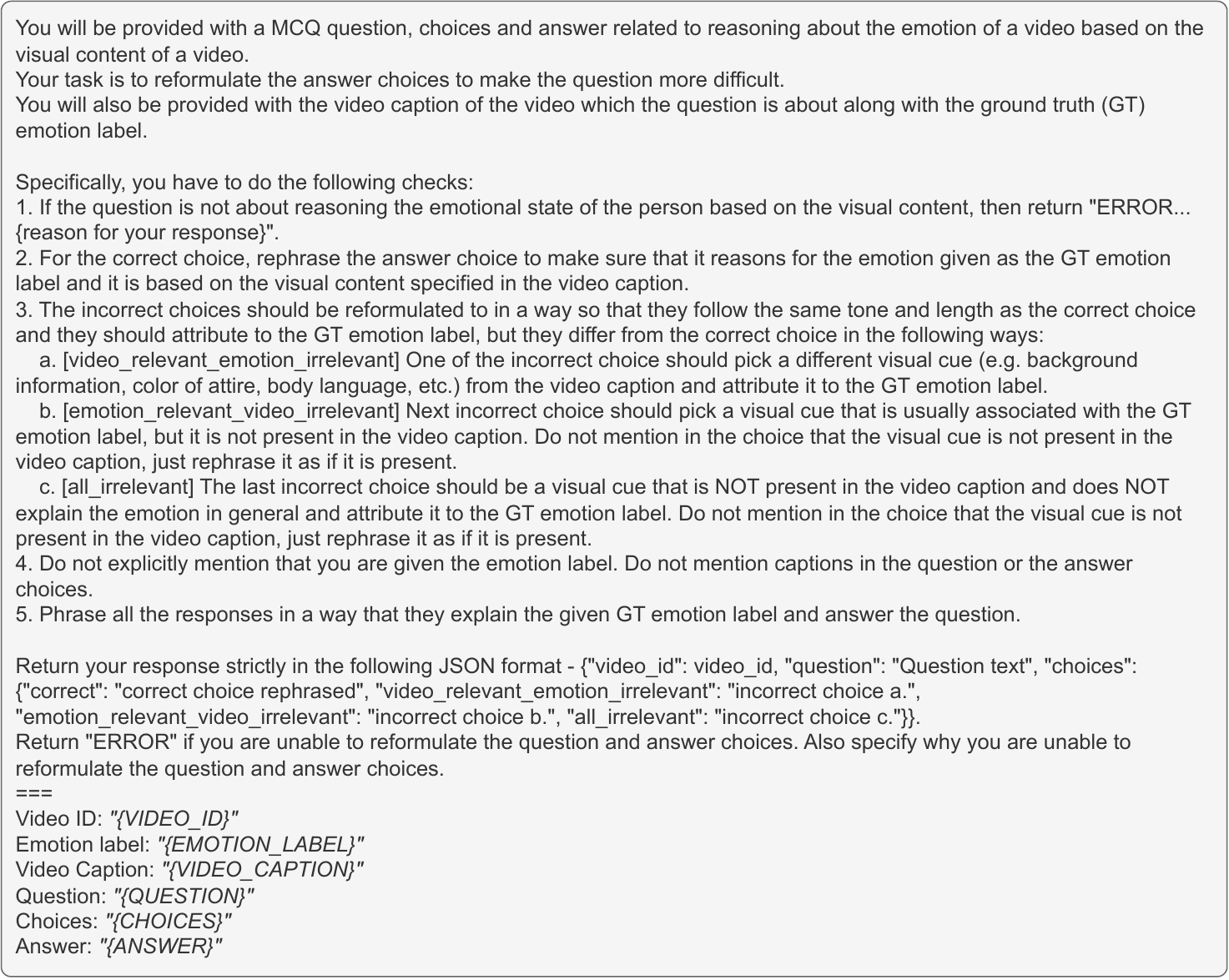}
    \caption{\textbf{Preference Data Generation Prompt - Visual Reasoning} --  used to generate rejected responses for a generated question-answer pair related to visual reasoning tasks.}
    \label{fig:prompt_dpo_visual_reasoning}
\end{figure}

\begin{figure}[!h]
    \centering
    \includegraphics[width=\linewidth]{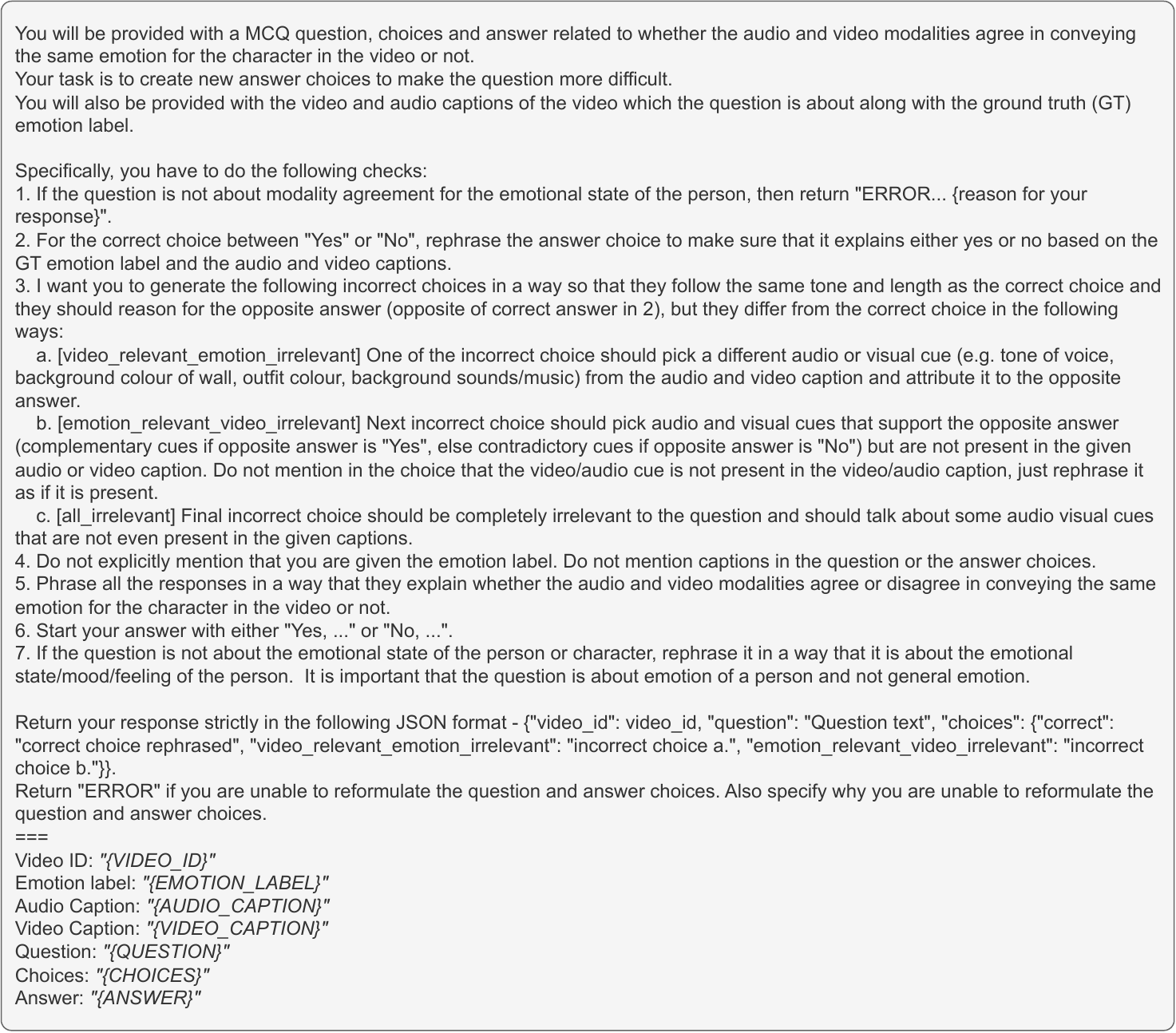}
    \caption{\textbf{Preference Data Generation Prompt - Modality Agreement} --  used to generate rejected responses for a generated question-answer pair related to modality agreement tasks.}
    \label{fig:prompt_dpo_modality_agreement}
\end{figure}

As mentioned in \cref{subsec:preference_data}, we use a pipeline similar to \cref{fig:emo_realm_data_pipeline} to construct our preference data using MAFW \citep{liu_mafw_2022} and MER2025 \citep{lian2025mer} \emph{Track 1 train set} as the source datasets. Note that we use Gemini 2.5 Flash \citep{comanici2025gemini25pushingfrontier_gemini25} for all automatic annotations required to create the training dataset. Use of Gemini for training data creation reduces annotation budget and ensures that the training dataset is not biased to have similar language as the test dataset -- \emph{EmoReAlM}. Since the pipeline in \cref{fig:emo_realm_data_pipeline} creates MCQA samples, we use another round of automatic annotations through Gemini-2.5 Flash over the generated MCQA samples to create the preference data. Specifically, we use prompts in \cref{fig:prompt_dpo_audio_reasoning,fig:prompt_dpo_visual_reasoning,fig:prompt_dpo_modality_agreement} to generate rejected responses for the generated emotion reasoning QA samples. Since, we also want to improve the performance on emotion description tasks present in EMER \citep{lian2023explainable_emer} we use prompts for audio (\cref{fig:prompt_dpo_audio_reasoning}) and visual reasoning (\cref{fig:prompt_dpo_visual_reasoning}) to modify emotion descriptions generated from Gemini 2.5 Flash (using prompt in \cref{fig:prompt_caption_combined}), combining audio and visual captions of MAFW and MER2025 (obtained using prompts in \cref{fig:prompt_audio_caption,fig:prompt_video_caption}). After Gemini annotation, we end up with a total of 41687 preference samples combining tasks, which we use for AVEm-DPO training. \cref{tab:preference_data_samples} contains samples from the constructed preference dataset using the described pipeline.

\begin{table}[]
\centering
\caption{Examples of the preference dataset used for AVEm-DPO.}
\label{tab:preference_data_samples}
\resizebox{\textwidth}{!}{%
\begin{tabular}{l|l|l|l|l}
\hline \hline
\rowcolor[HTML]{C0C0C0} 
\multicolumn{1}{c|}{\cellcolor[HTML]{C0C0C0}\textbf{Video}} & \multicolumn{1}{c|}{\cellcolor[HTML]{C0C0C0}\textbf{\begin{tabular}[c]{@{}c@{}}Prompt $(x)$\end{tabular}}} & \multicolumn{1}{c|}{\cellcolor[HTML]{C0C0C0}\textbf{\begin{tabular}[c]{@{}c@{}}Chosen Response $(y_w)$\end{tabular}}} & \multicolumn{1}{c|}{\cellcolor[HTML]{C0C0C0}\textbf{\begin{tabular}[c]{@{}c@{}}Rejected Response \\(video-relevant -  $y_l^{vr}$)\end{tabular}}} & \multicolumn{1}{c}{\cellcolor[HTML]{C0C0C0}\textbf{\begin{tabular}[c]{@{}c@{}}Rejected Response\\ (emotion-relevant -  $y_l^{er}$)\end{tabular}}} \\ \hline
\begin{tabular}[c]{@{}l@{}}\includegraphics[width=8cm,trim=0 4cm 0 0,clip]{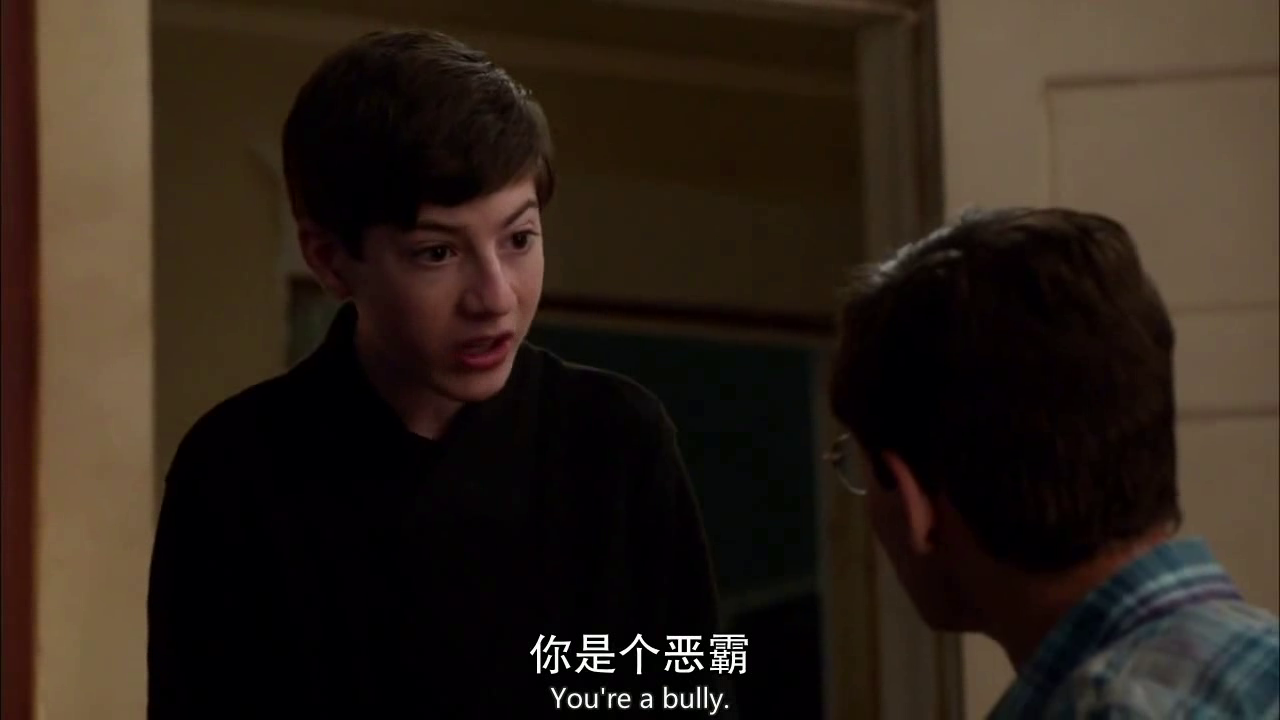}\\Subtitle: ``You're a bully. But I  can never fight back,\\ because you are JJ!"\end{tabular} & \begin{tabular}[c]{@{}l@{}}How do the facial expre-\\ssions of the young person \\ contribute to the emotional\\ intensity during the exchange?\end{tabular} & \begin{tabular}[c]{@{}l@{}}The young person's furrowed\\ eyebrows and open mouth \\ emphasize their intense emo-\\tional state and frustration.\end{tabular} & \begin{tabular}[c]{@{}l@{}}The dark top worn by the young\\ person underlines the seriousness \\ of their mood.\end{tabular} & \begin{tabular}[c]{@{}l@{}}The young person's hands \\clenching into fists and \\ subtle scowling underline their \\frustration.\end{tabular} \\ \hline
\begin{tabular}[c]{@{}l@{}}\includegraphics[width=8cm,trim=0 8cm 0 5cm,clip]{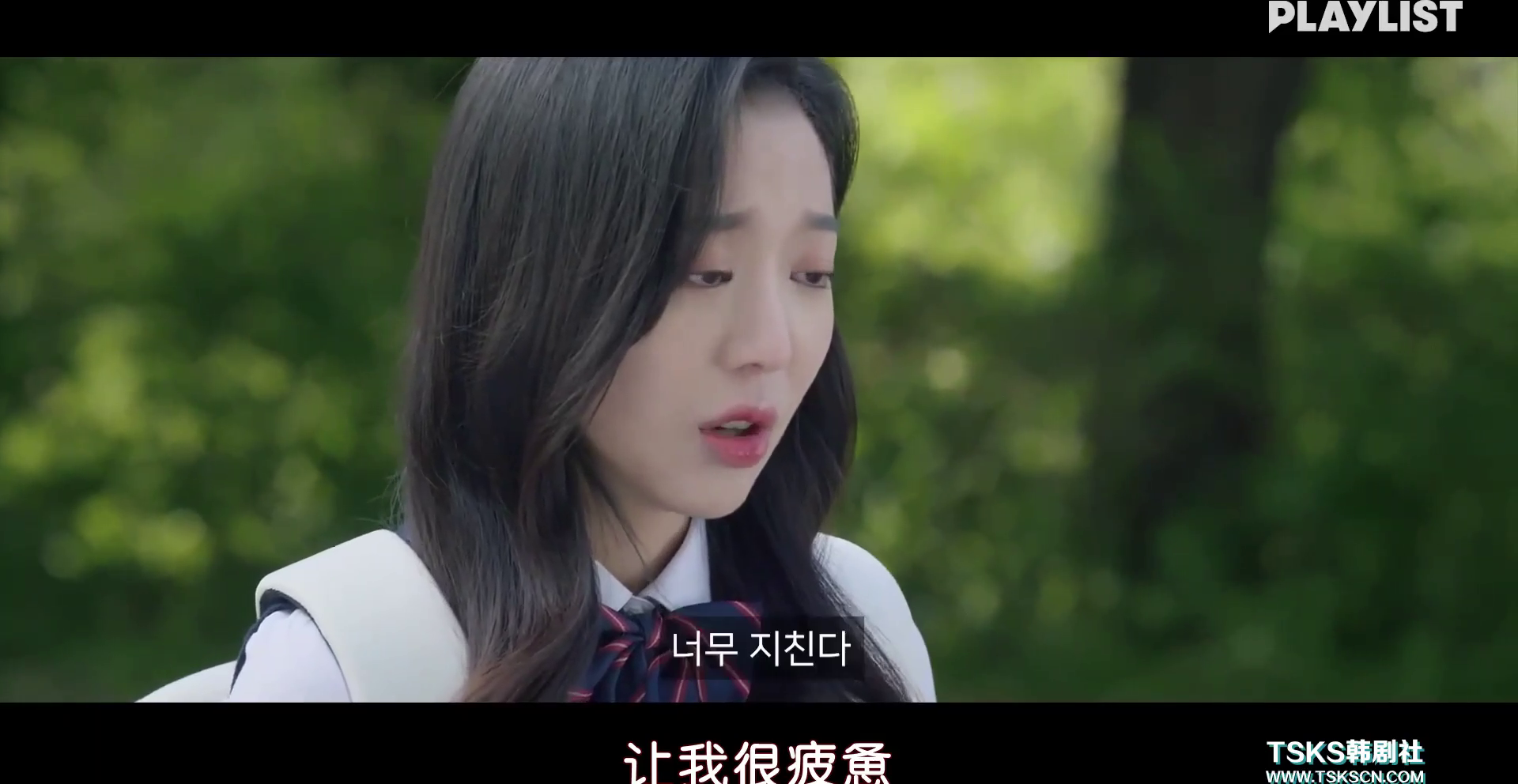}\\Subtitle: ``I'm so tired."\end{tabular} & \begin{tabular}[c]{@{}l@{}}How does the woman's mes-\\sage in the video reflect her\\ emotional state?\end{tabular} & \begin{tabular}[c]{@{}l@{}}She communicates a deep sense\\ of exhaustion and emotional \\weariness through her words, \\saying 'I'm so tired,' which\\ indicates her sadness.\end{tabular} & \begin{tabular}[c]{@{}l@{}}The melancholic piano music\\ in the background underscores \\the emotional heaviness she \\is experiencing.\end{tabular} & \begin{tabular}[c]{@{}l@{}}Her loud expressive crying,\\ typically associated with sadness,\\ conveys the depth of her emotional\\ state.\end{tabular} \\ \hline
\begin{tabular}[c]{@{}l@{}}\includegraphics[width=8cm,trim=0 4cm 0 4cm,clip]{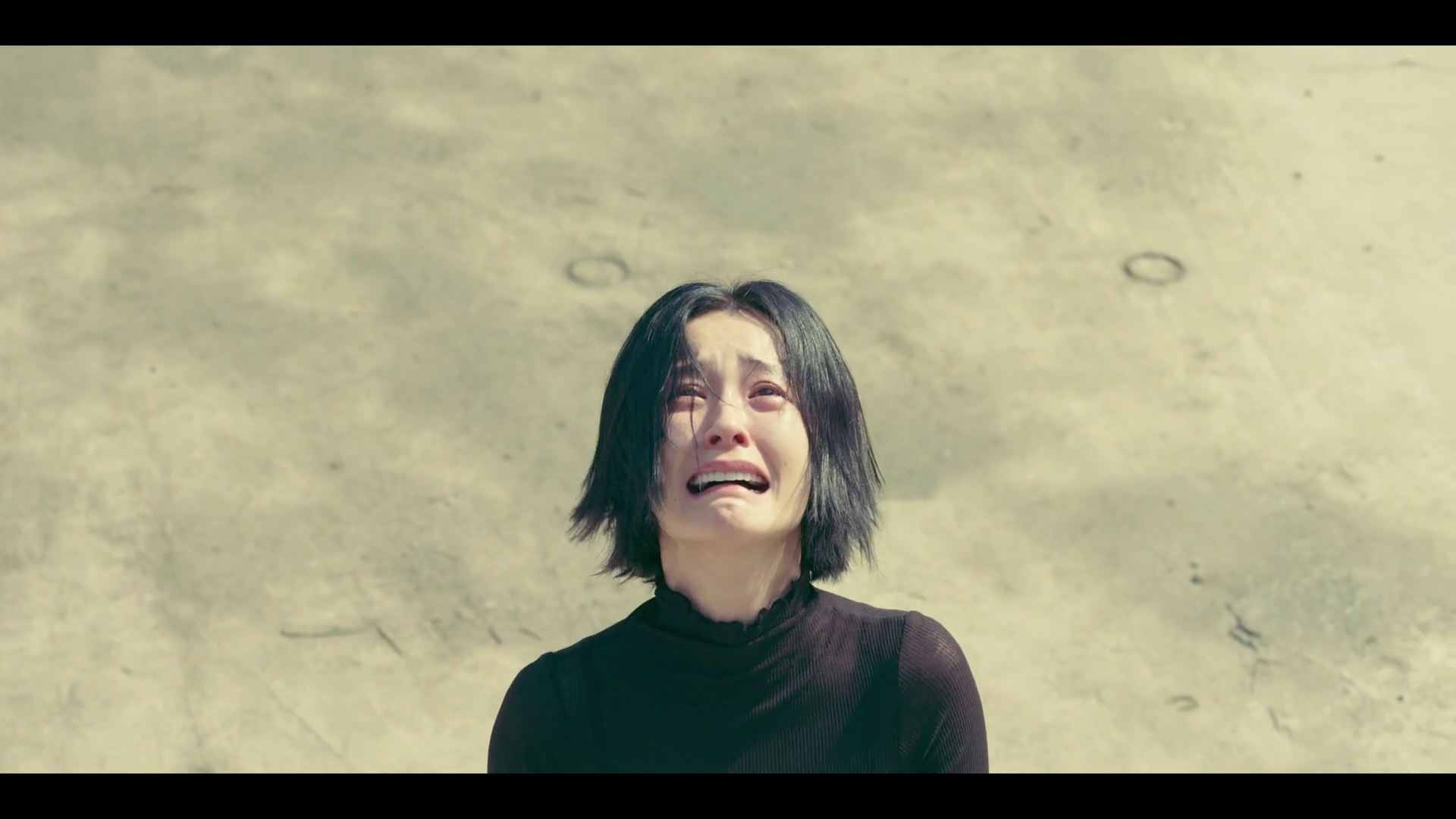}\\Subtitle: ``(crying)"\end{tabular} & \begin{tabular}[c]{@{}l@{}}Do the audio and video convey \\the same emotional state for \\the woman in the video?\end{tabular} & \begin{tabular}[c]{@{}l@{}}Yes, both the audio and \\video convey a profound sense\\ of sadness through the sounds\\ of crying and the woman's \\distraught facial expression.\end{tabular} & \begin{tabular}[c]{@{}l@{}}No, the tone of voice\\ in the audio appears sad,\\ but the stark background in\\ the video suggests a more calm\\ atmosphere.\end{tabular} & \begin{tabular}[c]{@{}l@{}}No, the woman's facial expression \\indicates a sense of fear, while\\ her words ``I can not take \\it anymore" suggest sadness.\end{tabular} \\ \hline\hline
\end{tabular}%
}
\end{table}

\subsection{Implementation Details}
\label{subsec:appendix_implementation_details}
We train the reference models using AVEm-DPO for one epoch, with a learning rate of $5e^{-7}$ and per GPU batch size of 2 on an NVIDIA DGX node with 8 NVIDIA H100 GPUs. We choose $\beta$ as 0.1 similar to \citep{huang2025mathcalvistadpo}. Moreover, $\lambda_{av}$ is set to 1.0, $\beta_{er}$ and $\beta_{vr}$ are both set to 0.5, and $\gamma_{\text{TPD}}$ is set to 0.2 (refer to \cref{subsec:appendix_sensitivity_to_hyperparams} for details on choice). We attach LoRA module with rank 8 and scale 4 to the LLM backbone for training. Gradient accumulation is used to accumulate gradients over 4 iterations.

\section{Experimental Details}
\label{sec:appendix_experimental_details}
\subsection{Evaluation Metrics}
\label{subsec:appendix_eval_metrics}
\begin{figure}[!h]
    \centering
    \includegraphics[width=\linewidth]{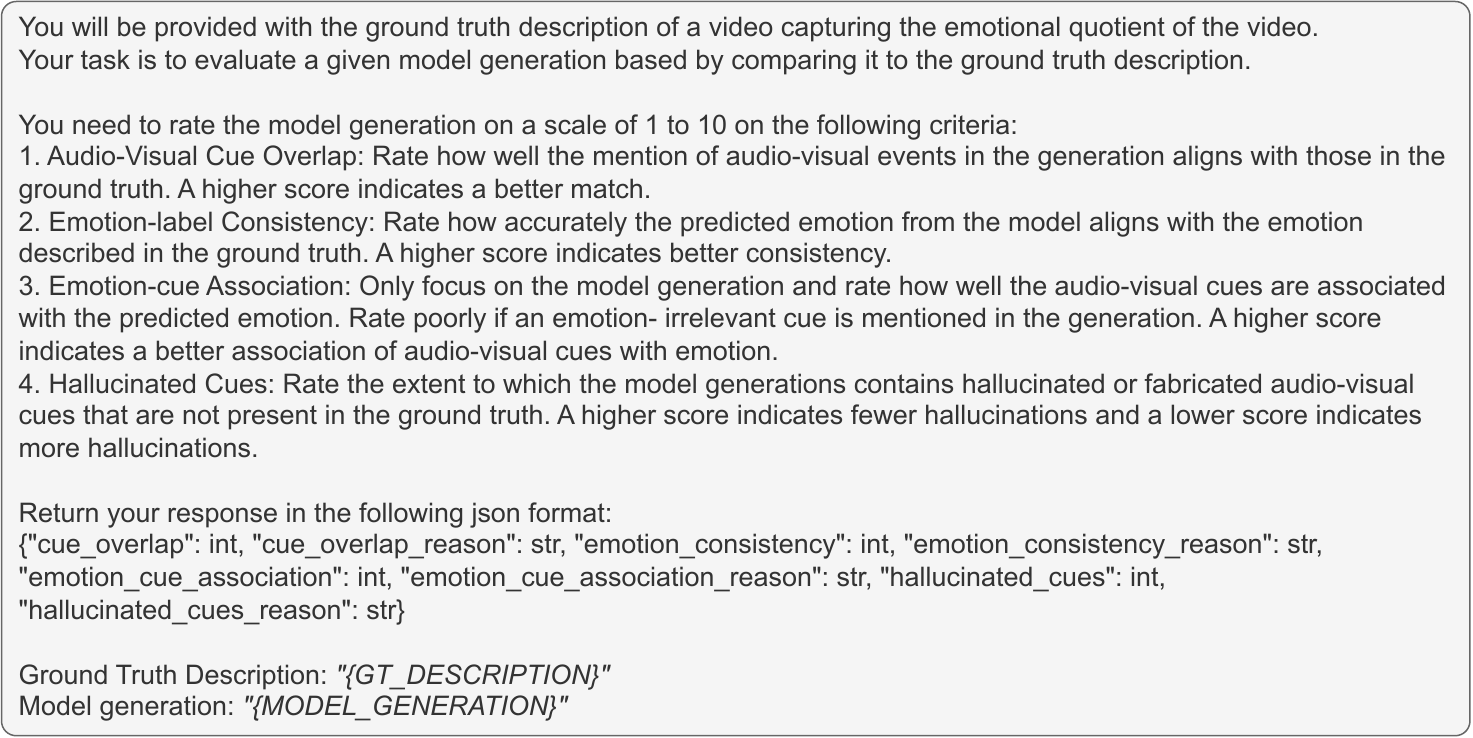}
    \caption{\textbf{EMER Evaluation Prompt} --  used to evaluate the model generations against the provided ground truths for the EMER dataset.}
    \label{fig:prompt_gpt_evaluation_emer}
\end{figure}

\paragraph{GPT Evaluation on EMER.} As mentioned in \cref{subsec:experimental_setup}, we perform GPT-4o evaluation on the generated emotion descriptions in EMER \citep{lian2023explainable_emer} dataset. We perform the evaluation over the following criterias -- (i) \emph{clue overlap} - similarity of the audiovisual cues present in the generation with the ground truth, (ii) \emph{label overlap} - similarity of the emotion label described in the generation with the ground truth, (iii) \emph{spurious cue-emotion associations} - how good are the audiovisual cues associated with emotions in the generation, and (iv) \emph{hallucinatory cues} - presence of cues that are absent in the ground truth but present in the generations. The prompt used to evaluate the generations is present in \cref{fig:prompt_gpt_evaluation_emer}.

\paragraph{EmoReAlM Evaluation Metrics.} For all the tasks in \emph{EmoReAlM}, we report the average accuracy over the task, computed as the number of correct responses out of the total number of samples in the task. Additionally, for tasks with \emph{``Yes"/``No"} responses (\emph{Modality Agreement} and \emph{Emotion Reasoning - Stress Test}), we report the precision, recall and F1 score. Precision and recall are the ratios of correctly answered questions that have correct answers as \emph{Yes} and \emph{No}, respectively. F1 score is the harmonic mean of precision and recall.

\subsection{Reference Models}
\label{subsec:appendix_reference_models}
We describe the reference models mentioned in \cref{subsec:experimental_setup} below.

\begin{figure}[!h]
    \centering
    \includegraphics[width=\linewidth]{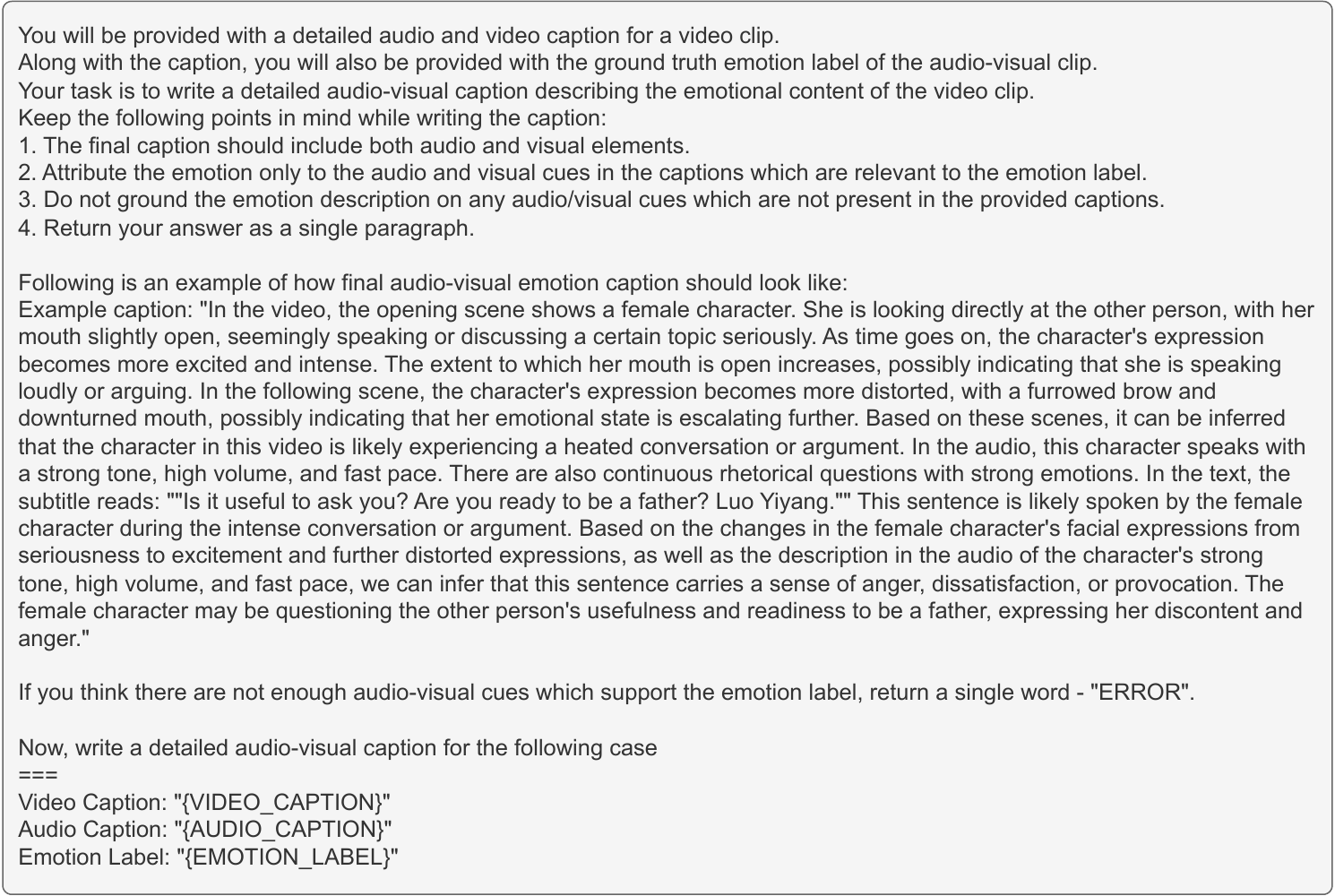}
    \caption{\textbf{Audiovisual Caption Prompt} --  used to combine audio and visual captions to create a combined audiovisual caption.}
    \label{fig:prompt_caption_combined}
\end{figure}

\paragraph{Our base.} We modify EmotionLLaMA \citep{emotion_llama} to replace the visual encoder with LanguageBind Video Encoder \citep{zhu2024languagebind} and audio encoder with Whisper Large v3 \citep{radford2023robust_whisper}. We pretrain the visual projector using the pretraining data of VideoLLaVA \citep{lin-etal-2024-videollava} and the audio projector is pretrained using LibriSpeech \citep{librispeech} and SpeechCraft \citep{jin2024speechcraft} to enhance paralinguistic capabilities of the model. We finetune on the EmotionLLaMA dataset, however, we include additional instruction data by annotating MAFW \citep{liu_mafw_2022} and MER2025 \citep{lian2025mer} \emph{Track 1 train set} through Gemini 2.5 Flash. Specifically, we use the prompts mentioned in \cref{subsec:appendix_prompts_benchmark} to create a finetuning dataset with similar tasks as in the proposed EmoReAlM benchmark. We also use prompt in \cref{fig:prompt_caption_combined} to generate emotion descriptions from MAFW and MER2025. 

\paragraph{EmotionLLaMA$^\star$.} Since the pretrained EmotionLLaMA model is not trained on tasks similar to \emph{EmoReAlM}, we finetune EmotionLLaMA on additional datasets created using MAFW and MER2025, similar to our base model described in the previous paragraph. Moreover, we do not provide subtitle text as input to the model during finetuning, in contrast to the original EmotionLLaMA, to eliminate external subtitle dependence.

\subsection{Baseline Preference Optimization Approaches}
\label{subsec:appendix_baseline_preference_optimization_methods}

We describe the implementation of baseline DPO approaches mentioned in \cref{subsec:experimental_setup} below. We use the same training setup as mentioned in \cref{subsec:appendix_implementation_details} unless stated otherwise.

\paragraph{Naive-DPO.} For Naive-DPO \citep{rafailov2023direct_dpo_orig} we use the objective in \cref{eq:naive_dpo_objective}. We use the preference samples from our preference data (\cref{subsec:appendix_preference_data}), and pick the rejected response randomly between $y_l^{vr}$ and $y_l^{er}$. 

\paragraph{Vista-DPO\textsuperscript{\textdagger}.} We adapt Vista-DPO \citep{huang2025mathcalvistadpo} for audiovisual inputs using \cref{eq:dpo_av_preference,eq:dpo_response_preference}. Also, we use our preference data (\cref{subsec:appendix_preference_data}) to optimize \cref{eq:dpo_av_preference} and drop their temporal (clip-based) and object-based preferences. Instead of prompt-based modality preference, we use $(a_l, v_l)$ to be an audiovisual input that has a different emotion than that of $(a_w, v_w)$, always irrespective of the input prompt.

\subsection{Baseline implementations}
\label{subsec:appendix_baseline_implementation_details}

\paragraph{Audiovisual baselines.} We use the official code for Qwen 2.5 Omni - 7B \citep{xu2025qwen25omnitechnicalreport} and run inference using flash attention 2. We use their default system prompt during inference. 

For Video-LLaMA \citep{damonlpsg2023videollama}, we use the official video-language checkpoint \textit{finetune-vicunna7b-v2} and audio-language checkpoint \textit{finetune-vicuna7b-audiobranch}. We also use the default conversation template for inference.

For PandaGPT \citep{su2023pandagptmodelinstructionfollow}, we use their official pretrained checkpoint \textit{pandagpt-7b} with 1,024 \textit{max\_len}, built upon ImageBind \citep{girdhar2023imagebindembeddingspacebind}. The system prompt remains unchanged during inference.

For OneLLM \citep{han2025onellmframeworkalignmodalities}, we use the released pretrained checkpoint \textit{OneLLM-7B}; for inference, we manually prepend the multimodal representations before the textual prompt.

We use VITA-1.5 \citep{fu2025vita} with its official code and checkpoint, including the \textit{InternViT-300M} vision tower and the pretrained audio encoder. We use the default conversation template for inference.

\paragraph{Audio-only baselines.} We use the official \textit{Qwen2-Audio-7B-Instruct} \citep{Qwen2-Audio} checkpoint and its default conversation template with the original system prompt.

For Kimi-Audio \citep{ding2025kimiaudio}, we use the released \textit{Kimi-Audio-7B-Instruct} checkpoint with the default system message.

For Audio Flamingo 3 \citep{goel2025audioflamingo3}, we use the official repository, pretrained checkpoint, and the default empty conversation template.

\paragraph{Video-only baselines.} We use the official code for InternVL3.5 \citep{wang2025internvl35advancingopensourcemultimodal}. Unlike others, this is an 8B model.

For Qwen2.5-VL \citep{bai2025qwen25vltechnicalreport}, we use the released \textit{Qwen2.5-VL-7B-Instruct} checkpoint with the default system prompt.

For \textit{VideoLLaMA3-7B} \citep{zhang2025videollama3frontiermultimodal}, we used the default system message and run inference with flash attention 2.

\subsection{Experimental Setup for Ablation Study}
\label{subsec:appendix_ablatlion_detailed_setup}

We describe the setup for the ablations mentioned in \cref{subsec:results_analysis} in detail below.

For \cref{tab:ablation_study,tab:modality_preference_ablation,fig:hyperparameter_ablation}, the metric reported for \emph{Emotion Reasoning – Basic} (denoted as \textbf{Basic}) is the unweighted average of the visual and audio reasoning accuracy on the \emph{Emotion Reasoning – Basic} task. For \emph{Emotion Reasoning – Stress Test} (denoted as \textbf{Stress}), the reported metric is the unweighted average of the F1 scores for visual and audio reasoning samples within the \emph{Emotion Reasoning – Stress Test} task. For \emph{Modality Agreement} (denoted as \textbf{Agree}), we report the F1 score over samples from the \emph{Modality Agreement} task. Additionally, for the subtasks \emph{Spurious Cue–Emotion Association} (denoted as \textbf{Spur.}) and \emph{Emotion-Relevant Cue Hallucination} (denoted as \textbf{Hall.}), we use the unweighted average accuracy across visual and audio reasoning samples for each respective subtask.

\paragraph{Ablation Study.} For \cref{tab:ablation_study}, the model without prompt-based modality preference (w/o PMP) is trained only using $\mathcal{L}^y_{\text{DPO-TPD}}$ (\cref{eq:dpo_tpd_objective_y}). The model without emotion-based response preference (w/o ERP) is trained using the the following loss,
\begin{equation}
\mathcal{L}_{\text{w/o ERP}} = \mathcal{L}_{\text{DPO-TPD}} + \mathcal{L}_{\text{DPO}}^{av-prompt}
\end{equation}
refer \cref{eq:dpo_tpd_objective,eq:dpo_av_preference_prompt_based} for the involved terms. Finally, the model without text prior debiasing (w/o TPD) is trained on the following objective,  
\begin{equation}
\mathcal{L}_{\text{w/o TPD}} = \mathcal{L}^y_{\text{DPO}} + \mathcal{L}_{\text{DPO}}^{av-prompt}
\end{equation}
refer \cref{eq:dpo_response_preference,eq:dpo_av_preference_prompt_based} for the involved terms.


\section{Detailed Results}
\label{sec:appendix_detailed_results}

\begin{table}[]
\centering
\caption{Performance comparison of different methods on the proposed EmoReAlM Benchmark. \textbf{Bold} are best results and \underline{underline} are second-best results over open-source models.}
\resizebox{\textwidth}{!}{%
\begin{tabular}{l|c|c|cccc|cccc|cccc|c}
\hline \hline
\rowcolor[HTML]{C0C0C0} 
\multicolumn{1}{c|}{\cellcolor[HTML]{C0C0C0}} & \multicolumn{2}{c|}{\cellcolor[HTML]{C0C0C0}\textbf{Reas. Basic}} & \multicolumn{4}{c|}{\cellcolor[HTML]{C0C0C0}} & \multicolumn{8}{c|}{\cellcolor[HTML]{C0C0C0}\textbf{Reasoning - Stress Test}} & \multicolumn{1}{c}{\cellcolor[HTML]{C0C0C0}} \\ \cline{2-3} \cline{8-15}
\rowcolor[HTML]{C0C0C0} 
\multicolumn{1}{c|}{\cellcolor[HTML]{C0C0C0}} & \multicolumn{1}{c|}{\cellcolor[HTML]{C0C0C0}\textbf{Audio}} & \multicolumn{1}{c|}{\cellcolor[HTML]{C0C0C0}\textbf{Visual}} & \multicolumn{4}{c|}{\multirow{-2}{*}{\cellcolor[HTML]{C0C0C0}\textbf{Modality Agreement}}} & \multicolumn{4}{c|}{\cellcolor[HTML]{C0C0C0}\textbf{Audio}} & \multicolumn{4}{c|}{\cellcolor[HTML]{C0C0C0}\textbf{Visual}} & \multicolumn{1}{c}{\cellcolor[HTML]{C0C0C0}} \\ \cline{2-15}
\rowcolor[HTML]{C0C0C0} 
\multicolumn{1}{c|}{\multirow{-3}{*}{\cellcolor[HTML]{C0C0C0}\textbf{Model}}} & \multicolumn{1}{c|}{\cellcolor[HTML]{C0C0C0}\textbf{Acc.}} & \multicolumn{1}{c|}{\cellcolor[HTML]{C0C0C0}\textbf{Acc.}} & \multicolumn{1}{c}{\cellcolor[HTML]{C0C0C0}\textbf{Acc.}} & \multicolumn{1}{c}{\cellcolor[HTML]{C0C0C0}\textbf{Pre.}} & \multicolumn{1}{c}{\cellcolor[HTML]{C0C0C0}\textbf{Rec.}} & \multicolumn{1}{c|}{\cellcolor[HTML]{C0C0C0}\textbf{F1}} & \multicolumn{1}{c}{\cellcolor[HTML]{C0C0C0}\textbf{Acc.}} & \multicolumn{1}{c}{\cellcolor[HTML]{C0C0C0}\textbf{Pre.}} & \multicolumn{1}{c}{\cellcolor[HTML]{C0C0C0}\textbf{Rec.}} & \multicolumn{1}{c|}{\cellcolor[HTML]{C0C0C0}\textbf{F1}} & \multicolumn{1}{c}{\cellcolor[HTML]{C0C0C0}\textbf{Acc.}} & \multicolumn{1}{c}{\cellcolor[HTML]{C0C0C0}\textbf{Pre.}} & \multicolumn{1}{c}{\cellcolor[HTML]{C0C0C0}\textbf{Rec.}} & \multicolumn{1}{c|}{\cellcolor[HTML]{C0C0C0}\textbf{F1}} & \multicolumn{1}{c}{\multirow{-3}{*}{\cellcolor[HTML]{C0C0C0}\textbf{\begin{tabular}[c]{@{}c@{}}Avg.\\ Acc.\end{tabular}}}} \\ \hline
\multicolumn{16}{c}{\emph{Closed-source models}} \\ \hline
Gemini 2.5 Flash & 78.0 & 88.9 & 57.0 & 75.9 & 39.0 & 51.5 & 63.5 & 74.0 & 51.0 & 60.4 & 73.2 & 75.3 & 70.9 & 73.0 & 72.1 \\ 
Gemini 2.5 Pro & 72.7 & 87.0 & 54.7 & 76.0 & 33.3 & 46.3 & 63.8 & 74.0 & 53.3 & 62.0 & 73.1 & 84.0 & 59.8 & 69.8 & 70.3 \\ \hline
\multicolumn{16}{c}{\emph{Open-source video-only models}} \\ \hline
VideoLLaMA 3 & - & 86.2 & - & - & - & - & - & - & - & - & 64.9 & \underline{97.9} & 33.0 & 49.4 & - \\
Qwen 2.5 VL & - & 88.1 & - & - & - & - & - & - & - & - & 75.2 & \textbf{98.6} & 52.6 & 68.5 & - \\
InternVL 3.5 & - & \textbf{92.8} & - & - & - & - & - & - & - & - & 68.3 & 91.6 & 45.8 & 61.1 & - \\ \hline
\multicolumn{16}{c}{\emph{Open-source audio-only models}} \\ \hline
Qwen 2 Audio & 56.6 & - & - & - & - & - & 55.1 & 84.2 & 28.3 & 42.3 & - & - & - & - & - \\
Kimi-Audio & 69.8 & - & - & - & - & - & 54.0 & \underline{95.8} & 15.5 & 26.6 & - & - & - & - & - \\
Audio Flamingo 3 & \underline{76.8} & - & - & - & - & - & 52.6 & \textbf{96.7} & 11.9 & 21.2 & - & - & - & - & - \\ \hline
\multicolumn{16}{c}{\emph{Open-source audiovisual (``omni") models}} \\ \hline
VideoLLaMA & 21.7 & 22.2 & 34.1 & 37.4 & 30.9 & 33.9 & 46.1 & 41.3 & 50.6 & 45.5 & 48.8 & 48.4 & 49.2 & 48.8 & 37.1 \\
PandaGPT & 37.4 & 35.7 & 53.7 & 50.3 & \textbf{56.9} & 53.4 & 45.8 & 62.9 & 30.1 & 40.7 & 47.1 & 59.9 & 34.7 & 43.9 & 44.0 \\
OneLLM & 42.0 & 55.6 & 54.8 & 64.3 & 45.9 & 53.5 & 56.8 & 87.1 & 28.9 & 43.4 & 62.0 & 97.6 & 27.6 & 43.1 & 54.2 \\
VideoLLaMA2 & 63.1 & 66.8 & 52.6 & 52.0 & \underline{53.0} & 52.5 & 53.7 & 60.6 & 47.3 & 53.2 & 59.4 & 67.9 & 51.2 & 58.4 & 59.1 \\
OLA & 63.2 & 60.4 & 51.7 & 78.9 & 29.8 & 42.7 & 63.5 & 86.8 & 41.9 & 56.6 & 62.3 & 85.0 & 40.4 & 54.8 & 60.2 \\
VITA-1.5 & 63.1 & 84.3 & 51.7 & 87.1 & 18.2 & 30.2 & 63.0 & 91.0 & 37.2 & 52.8 & 66.1 & 92.7 & 40.4 & 56.3 & 65.6 \\
Qwen 2.5 Omni & \underline{76.8} & 89.2 & 52.2 & 86.1 & 20.7 & 33.3 & 64.0 & 90.4 & 39.6 & 55.0 & 67.8 & 96.4 & 40.3 & 56.8 & 70.0 \\  \hline
\textbf{Our base} & 69.2 & 85.3 & 51.4 & 86.3 & 21.6 & 34.6 & 53.1 & 65.4 & 40.8 & 50.3 & 66.4 & 87.2 & 45.6 & 59.9 & 65.1 \\
+ Naive-DPO & 71.3 & 85.9 & 57.3 & 87.2 & 27.3 & 41.6 & 55.6 & 62.3 & 48.9 & 54.8 & 70.6 & 88.8 & 52.4 & 65.9 & 68.1 \\
+ Vista-DPO\textsuperscript{\textdagger} & 72.4 & 87.8 & 63.1 & 89.4 & 36.8 & 52.1 & 74.1 & 67.8 & 80.4 & 73.6 & 87.0 & 92.1 & 81.9 & 86.7 & 76.9 \\
\rowcolor[HTML]{DAE8FC}
+ \textbf{AVEm-DPO} & \textbf{77.9} & \underline{92.5} & \textbf{68.9} & \textbf{93.4} & {44.3} & \textbf{60.0} & \textbf{82.6} & {70.7} & \textbf{94.6} & \textbf{80.9} & \textbf{94.6} & {93.1} & \textbf{96.1} & \textbf{94.6} & \textbf{83.3} \\ 
\rowcolor[HTML]{DAE8FC}
$\Delta\%$  (relative) & 12.6 & 8.4 & 34.1 & 8.2 & 105. & 73.4 & 55.6 & 8.1 & 131. & 60.8 & 42.5 & 6.8 & 110. & 57.9 & 28.0 \\ \hline
\textbf{Emot.-LLaMA$^\star$} & 64.8 & 84.9 & 51.2 & 82.9 & 20.7 & 33.1 & 48.9 & 59.2 & 38.5 & 46.7 & 69.1 & 89.3 & 48.9 & 63.2 & 63.8 \\
+ Naive-DPO & 67.2 & 85.7 & 56.1 & 83.4 & 28.8 & 42.8 & 53.5 & 60.1 & 46.8 & 52.6 & 71.9 & 89.5 & 54.3 & 67.6 & 66.9 \\
+ Vista-DPO\textsuperscript{\textdagger} & 69.0 & 86.9 & 58.2 & 85.9 & 30.4 & 40.9 & 69.2 & 63.1 & 75.2 & 68.6 & 87.6 & 92.5 & 82.6 & 87.3 & 74.2 \\
\rowcolor[HTML]{DAE8FC}
+ \textbf{AVEm-DPO} & 76.5 & 89.1 & \underline{65.6} & \underline{89.5} & 41.6 & \underline{56.8} & \underline{77.3} & 65.2 & \underline{89.4} & \underline{75.4} & \underline{91.8} & 92.6 & \underline{90.9} & \underline{91.7} & \underline{80.1} \\ 
\rowcolor[HTML]{DAE8FC}
$\Delta\%$ (relative)  & 18.1 & 4.9 & 28.1 & 8.0 & 101. & 71.6 & 58.1 & 10.1 & 132. & 61.5 & 32.9 & 3.7 & 85.9 & 45.1 & 25.5 \\ \hline \hline
\end{tabular}%
}
\label{tab:emo_realm_main_results_detailed}
\end{table}

\subsection{EmoReAlM Results - Expanded}
\label{subsec:appendix_emo_realm_results_detailed_expanded}

\cref{tab:emo_realm_main_results_detailed} shows the the expanded version of \cref{tab:emo_realm_main_results} with accuracy, precision and recall metrics for \emph{Modality Agreement} and \emph{Emotion Reasoning - Stress Test} categories. We also report the unweighted average accuracy over all five tasks in the benchmark in the last column. The relative percent improvement of the AVEm-DPO trained model over the reference models is present as the $\Delta\%$ row. Moreover, we also report the performance of video-only and audio-only baselines in \cref{tab:emo_realm_main_results_detailed}. We can see that for visual reasoning tasks (\emph{Basic} and \emph{Stress Test}), video-only baselines perform slightly better than the audiovisual (\emph{``omni"}) baselines, aligning with the findings of \cite{sung-bin2025_avhbench}. However, for audio reasoning tasks, audiovisual baselines outperform audio-only baselines, which have very poor recall on the \emph{Emotion reasoning - Stress Test}. This can be attributed to the limited amount of audio-emotion datasets that the baselines \citep{Qwen2-Audio,ding2025kimiaudio,goel2025audioflamingo3} are trained on resulting in poor emotion reasoning.

\label{subsec:appendix_emo_realm_stress_detailed}
\begin{table}[]
\centering
\caption{Performance of different baselines on different Reasoning Stress-Test sub-tasks in EmoReAlM Benchmark. This experiment is done only using samples from the Stress-Test category of the benchmark which have correct answer as "No". \textbf{Bold} are best results and \underline{underline} are second-best results over open-source models.}
\label{tab:emo_realm_hallucination_types}
\resizebox{0.5\textwidth}{!}{%
\begin{tabular}{l|cc|cc}
\hline\hline
\rowcolor[HTML]{C0C0C0} 
\multicolumn{1}{c|}{\cellcolor[HTML]{C0C0C0}} & \multicolumn{2}{c|}{\cellcolor[HTML]{C0C0C0}\textbf{Audio}} & \multicolumn{2}{c}{\cellcolor[HTML]{C0C0C0}\textbf{Visual}} \\ \cline{2-5} 
\rowcolor[HTML]{C0C0C0} 
\multicolumn{1}{c|}{\multirow{-2}{*}{\cellcolor[HTML]{C0C0C0}\textbf{Model}}} & \multicolumn{1}{c|}{\cellcolor[HTML]{C0C0C0}\textbf{Spur.}} & \textbf{Hall.} & \multicolumn{1}{c|}{\cellcolor[HTML]{C0C0C0}\textbf{Spur.}} & \multicolumn{1}{c|}{\cellcolor[HTML]{C0C0C0}\textbf{Hall.}} \\ \hline
\multicolumn{5}{c}{\emph{Open-source video-only models}}\\\hline
VideoLLaMA 3 & - & - & 37.4 & 29.1 \\
Qwen 2.5 VL & - & - & 64.7 & 41.8 \\
InternVL 3.5 & - & - & 50.4 & 41.8 \\ \hline
\multicolumn{5}{c}{\emph{Open-source audio-only models}}\\\hline
Qwen 2 Audio & 41.8 & 16.9 & - & - \\
Kimi Audio & 26.8 & 6.0 & - & - \\
Audio Flamingo 3 & 15.7 & 8.7 & - & - \\ \hline
\multicolumn{5}{c}{\emph{Open-source audiovisual (``omni") models}}\\\hline
VideoLLaMA & 27.5 & 35.0 & 33.1 & 37.4 \\
PandaGPT & 43.1 & 19.1 & 47.5 & 23.4  \\
OneLLM & 47.7 & 13.1 & 36.7 & 19.6  \\
VideoLLaMA2 & 61.4 & 35.5 & 57.6 & 45.6  \\
OLA & 52.9 & 32.8 & 56.8 & 25.9  \\
VITA-1.5 & 46.4 & 29.5 & 46.0 & 35.4  \\
Qwen 2.5 Omni & 53.4 & 28.1 & 51.9 & 30.1  \\ \hline
\textbf{Our base} & 45.2 & 36.4 & 49.3 & 41.9  \\
+ Naive-DPO & 49.9 & 47.9 & 56.8 & 48.0  \\
+ Vista-DPO\textsuperscript{\textdagger} & \underline{85.7} & \underline{75.1} & \underline{87.1} & \underline{76.7}  \\
\rowcolor[HTML]{DAE8FC}
+ AVEm-DPO & \textbf{88.6} & \textbf{99.5} & \textbf{96.5} & \textbf{95.8} \\ \hline\hline
\end{tabular}%
}
\end{table}

\subsection{EmoReAlM Results on different Stress Test Subtasks}
\label{subsec:appendix_emo_realm_stress_test_results_detailed}

\cref{tab:emo_realm_hallucination_types} shows the performance of different baselines as well as AVEm-DPO on different subtasks of \emph{Emotion Reasoning - Stress Test}, which have answer as \emph{``No"} --  \emph{Spurious Cue-Emotion Association} and \emph{Emotion-relevant Cue Hallucination} (refer to \cref{subsec:task_descriptions,sec:appendix_benchmark_details} for definitions). We can observe that within audio and visual reasoning, hallucination seems to be a bigger bottleneck than spurious cue-emotion associations. Moreover, similar to \cref{tab:emo_realm_main_results_detailed}, we can observe that the audio-only models perform worse compared to audiovisual models, whereas the video-only model performance is better compared to audiovisual models. AVEm-DPO improves the model performance over all the subtasks significantly compared to the reference model.

\begin{table}[]
\centering
\caption{Class-wise recall for different emotion classes in DFEW dataset. \textbf{Bold} are best results and \underline{underline} are second-best results over open-source models.}
\label{tab:appendix_dfew_results_class_wise}
\resizebox{\textwidth}{!}{%
\begin{tabular}{l|c|ccccccc|cc}
\hline\hline
\rowcolor[HTML]{C0C0C0} 
\multicolumn{1}{c|}{\cellcolor[HTML]{C0C0C0}\textbf{Model}} & \textbf{Mod.} & \textbf{Hap.} & \textbf{Sad.} & \textbf{Neu.} & \textbf{Ang.} & \textbf{Sur.} & \textbf{Dis.} & \textbf{Fea.} & \textbf{UAR} & \textbf{WAR} \\ \hline
\multicolumn{11}{c}{\emph{Open-source video-only models}} \\ \hline
VideoLLaMA 3 & V & 77.92 & 41.38 & 40.88 & 42.53 & 26.44 & \textbf{34.26} & 72.30 & 47.96 & 49.47 \\
Qwen 2.5 VL & V & 64.21 & 52.37 & \textbf{69.49} & 39.09 & 11.38 & 7.20 & \underline{75.03} & 45.54 & 52.32 \\
InternVL 3.5 & V & 79.49 & 77.20 & 45.42 & 21.38 & 53.02 & 12.61 & 62.10 & 50.18 & 55.46 \\ \hline
\multicolumn{11}{c}{\emph{Open-source audio-only models}} \\ \hline
Qwen 2 Audio & A & 64.55 & 25.08 & 2.28 & 0.00 & 0.06 & 2.07 & 53.55 & 21.08 & 22.24 \\
Kimi Audio & A & 50.34 & 42.97 & 37.50 & {71.24} & 12.66 & 10.34 & 29.93 & 36.43 & 43.30 \\
Audio Flamingo 3 & A & 2.98 & 19.96 & 12.92 & 83.01 & 6.12 & 15.86 & 41.46 & 26.05 & 26.39 \\ \hline
\multicolumn{11}{c}{\emph{Open-source audiovisual (``omni") models}} \\ \hline
PandaGPT & A,V & 60.50 & 9.95 & 0.0 & 58.61 & 0.00 & 0.00 & 0.00 & 18.44 & 24.20 \\
VideoLLaMA & A,V & \underline{85.04} & 8.41 & 4.17 & 20.84 & 3.95 & 0.00 & 1.14 & 17.65 & 24.09 \\
OneLLM & A,V & 47.91 & 54.33 & 3.23 & 52.35 & 26.08 & 1.80 & 70.21 & 36.74 & 37.60 \\
VideoLLaMA2 & A,V & \textbf{87.50} & 57.93 & 7.94 & 58.56 & 42.08 & 15.00 & 36.54 & 43.65 & 48.66 \\
OLA & A,V & 52.00 & \textbf{82.20} & 15.65 & 48.95 & 9.65 & 10.00 & 48.72 & 38.17 & 41.73 \\
VITA-1.5 & A,V & 61.46 & \underline{79.96} & 23.54 & 23.19 & 8.05 & 0.90 & \textbf{78.07} & 39.31 & 42.56 \\
Qwen 2.5 Omni & A,V & 45.45 & 73.84 & 61.11 & 70.64 & 4.40 & 0.00 & 73.15 & 46.94 & 54.33 \\
EmotionLLaMA & A,V,T & 71.98 & 76.25 & \underline{61.99} & {71.95} & 33.67 & 0.00 & 3.31 & 45.59 & 59.37 \\
MoSEAR & A,V,T & 79.35 & 75.20 & 40.45 & 69.66 & 42.86 & 0.00 & 3.87 & 44.48 & 56.60 \\ \hline
Our base & A,V & 70.75 & 72.07 & 29.64 & \textbf{77.04} & \underline{61.54} & \underline{27.59} & 58.87 & \underline{56.78} & \underline{60.14} \\
\rowcolor[HTML]{DAE8FC}
+AVEm-DPO & A,V & 75.21 & 72.03 & 44.07 & \underline{73.96} & \textbf{62.24} & 17.24 & 65.00 & \textbf{58.54} & \textbf{64.24} \\ \hline\hline
\end{tabular}%
}
\end{table}

\subsection{Emotion Recognition Results - Expanded}
\label{subsec:appendix_emotion_recognition_results_expanded}

\cref{tab:appendix_dfew_results_class_wise} (expanded from \cref{tab:av_emotion_results}) shows the results on DFEW \citep{jiang2020dfew} emotion recognition benchmark over different emotion classes. Note that both our base model and AVEm-DPO trained model achieve the best and second-best results in terms of unweighted and weighted average recalls over all the emotion classes. Moreover, \cref{tab:appendix_dfew_results_class_wise} shows that the proposed method ensures fair performance over all the emotion categories, unlike baselines, which perform too well on some classes and too poorly on the others.

\begin{figure}[t]
  \centering
  \begin{minipage}{0.66\linewidth}
    \centering
    \includegraphics[width=\linewidth,trim=1mm 5mm 3mm 2mm, clip]{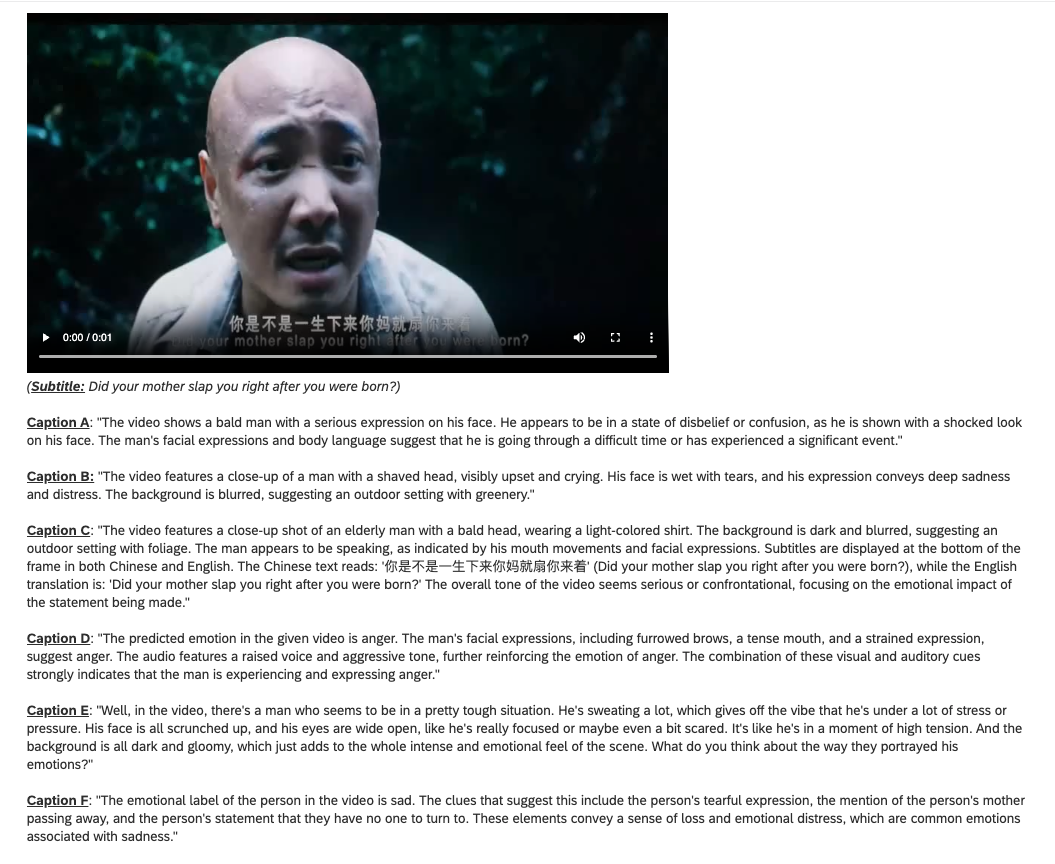}
  \end{minipage}\hfill
  \begin{minipage}{0.33\linewidth}
    \centering
    \includegraphics[width=\linewidth,trim=2mm 5mm 2mm 2mm, clip]{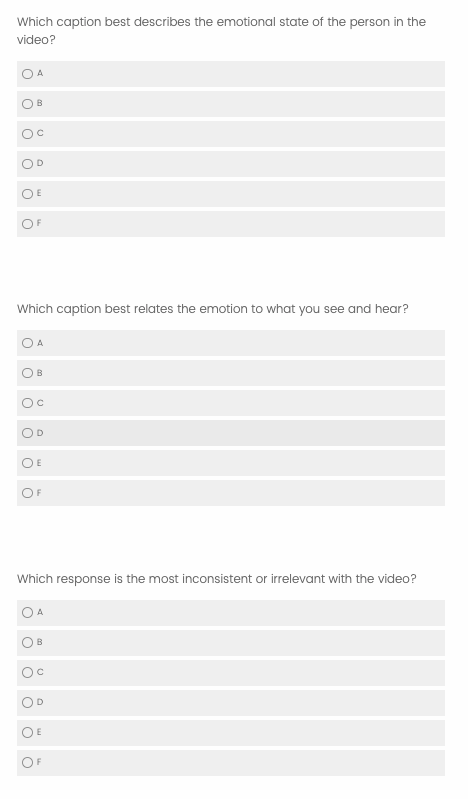}
  \end{minipage}
  \caption{\textbf{User Evaluation using Qualtrics}. \emph{(Left)} We show anonymized model responses for a given video to the user as different captions. \emph{(Right)} We ask multiple questions to the user to select the best-suited caption for each question. Questions check the captions for their quality of emotion description, association of emotions with audiovisual cues, and presence of inconsistencies (hallucinations).}
  \label{fig:prolific_user_evaluation}
\end{figure}

\subsection{User Evaluation}
\label{subsec:appendix_user_evaluation}

\begin{table}
\centering
\caption{User evaluation on EMER dataset.}
\vspace{-1em}
\label{tab:appendix_user_evaluation_emer}
\resizebox{0.5\linewidth}{!}{%
\begin{tabular}{lccc}
\hline\hline
\rowcolor[HTML]{C0C0C0} 
\multicolumn{1}{c}{\cellcolor[HTML]{C0C0C0}\textbf{Model}} & \textbf{Emot.$\uparrow$} & \textbf{Assoc.$\uparrow$} & \textbf{Incons.$\downarrow$} \\ \hline
VideoLLaMA 2 & 9.82\% & 0.75\% & 15.38\% \\
OLA & 9.36\% & 7.46\% & 5.58\% \\
VITA 1.5 & 11.60\% & 17.25\% & 6.04\% \\
Qwen 2.5 Omni & 10.75\% & 18.57\% & 10.13\% \\
EmotionLLaMA & 1.89\% & 11.53\% & 68.61\% \\
\rowcolor[HTML]{DAE8FC}
Our + AVEm-DPO & 54.74\% & 43.35\% & 4.67\% \\ \hline\hline
\end{tabular}%
}
\end{table}

We perform a user study on 40 participants recruited through Prolific \citep{prolificProlificEasily} and create a user survey using Qualtrics \citep{qualtricsQualtricsExperience} as shown in \cref{fig:prolific_user_evaluation}. We randomly sample videos from EMER \citep{lian2023explainable_emer} dataset and display anonymized model generations as captions to the user along with the video. Then we ask the users to pick the most suited caption over different criteria -- (i) best caption describing the emotional state of the person, (ii) best caption associating the emotion with audiovisual cues, and (iii) worst caption with the most inconsistencies with the video (to test model hallucinations). \cref{tab:appendix_user_evaluation_emer} (duplicate of \cref{tab:user_evaluation_emer}) reports the average percent of times each model is selected for the mentioned three criteria. The participants selected our model the most number of times as the best model for emotion description and association of audiovisual cues for emotion. Moreover, our model was chosen the least number of times for inconsistent audiovisual information present in the caption.

\begin{wraptable}[12]{r}{0.5\linewidth}
\centering
\vspace{-1em}
\caption{Performance variation over various choices of rejected multimodal input. \textbf{Change} denotes which among ($a_w$, $v_w$) should be changed to create ($a_l$, $v_l$).}
\vspace{-1em}
\label{tab:modality_preference_ablation}
\resizebox{\linewidth}{!}{%
\begin{tabular}{l|c|ccc}
\hline\hline
\rowcolor[HTML]{C0C0C0} 
\textbf{Choice of $a_l$/$v_l$} & \textbf{Change} & \textbf{Basic} & \textbf{Agree.} & \textbf{Stress}  \\ \hline
 & Both $a_l$, $v_l$ & 81.9 & 56.1 & 80.1 \\
\multirow{-2}{*}{Random tensor} & Prompt-based & 83.0 & 56.0 & 81.6 \\
 & Both $a_l$, $v_l$ & 81.8 & 58.2 & 80.3 \\ 
\multirow{-2}{*}{Random video} & Prompt-based & 83.6 & 58.2 & 82.1 \\
 & Both $a_l$, $v_l$ & 82.7 & 58.5 & 80.9 \\ 
\multirow{-2}{*}{Diffuse $(a_w,v_w)$} & Prompt-based & 84.6 & 59.4 & 86.7 \\
 {\cellcolor[HTML]{DAE8FC}}& Both $a_l$, $v_l$ & 83.9 & 60.1 & 81.3 \\
 \rowcolor[HTML]{DAE8FC}
\multirow{-2}{*}{Diff. emotion} & Prompt-based & 85.2 & 60.0 & 87.8 \\\hline\hline
\end{tabular}%
}
\end{wraptable}

\subsection{Modality Preference Ablation}
\label{subsec:appendix_modality_preference_ablation}
\cref{tab:modality_preference_ablation} shows AVEm-DPO's performance for different choices of multimodal preferences. We perform experiment using random tensor, random video, $(a_l, v_l)$ infused with diffusion noise similar to VCD \citep{visual_contrastive_decoding_vcd} and an audiovisual input with different emotion than $(a_w, v_w)$ as the possible choices for $(a_l, v_l)$ and show that using a different emotion video leads to the best results. Moreover, we also show the effect of changing both $(a_w, v_w)$ vs. changing based on the input prompt ($a_w$ for audio reasoning, $v_w$ for visual reasoning and both for other tasks), justifying the effectiveness of prompt-based modality preference. 

\begin{table}[]
\centering
\caption{Performance variation over various choices of rejected response. $y_l^{irr}$: response completely irrelevant to the audiovisual content and emotion, $y_l^{er}$: response mentions hallucinated cues that generally co-occur with given emotion, $y_l^{vr}$: response associates audiovisual cues in the input incorrectly with emotion.}
\label{tab:text_preference_ablation}
\resizebox{0.6\linewidth}{!}{%
\begin{tabular}{cc|ccc|cc}
\hline\hline
\rowcolor[HTML]{C0C0C0} 
\textbf{$y^1_l$} & \textbf{$y_l^2$} & \textbf{Basic} & \textbf{Agree.} & \textbf{Stress} & \textbf{Spur.} & \textbf{Hall.} \\ \hline
\multicolumn{2}{c|}{\textbf{Our base}} & 77.3 & 34.6 & 55.1 & 47.3 & 39.2 \\ \hline
$y_l^{irr}$ & - & 82.4 & 56.7 & 81.4 & 85.1 & 88.9 \\
$y_l^{er}$ & - & 84.0 & 58.3 & 86.0 & 88.5 & 97.9 \\
$y_l^{vr}$ & - & 83.2 & 58.0 & 85.3 & 91.6 & 90.9 \\
$y_l^{er}$ & $y_l^{irr}$ & 83.4 & 57.6 & 85.8 & 88.2 & 97.8 \\
$y_l^{vr}$ & $y_l^{irr}$ & 83.1 & 57.3 & 84.9 & 90.3 & 90.8 \\
\rowcolor[HTML]{DAE8FC}
$y_l^{er}$ & $y_l^{vr}$ & 85.2 & 60.1 & 87.8 & 92.7 & 97.6 \\ \hline\hline
\end{tabular}%
}
\end{table}

\subsection{Response Preference Ablation}
\label{subsec:appendix_response_preference_ablation}
\cref{tab:text_preference_ablation} shows the variation of performance over different tasks of \emph{EmoReAlM} for different choices of rejected responses. There are three types of rejected responses that we test on -- (i) $y_l^{vr}$ is video-relevant response that contains audiovisual cue present in the video, but it does not associate with the emotion, (ii) $y_l^{er}$ is emotion-relevant response that correctly associates with the emotion displayed in the video but with audiovisual cues that are hallucinated (not present in the video), and (iii) $y_l^{irr}$ is completely irrelevant to the given video and emotion (similar to that present in \cite{huang2025mathcalvistadpo}). $y_l^1$ and $y_l^2$ in \cref{tab:text_preference_ablation} denote the first and second rejected responses for preference tuning in \cref{eq:dpo_tpd_objective_y}.

We can see that our choice of using $y_l^{vr}$ and $y_l^{er}$ in \cref{eq:dpo_tpd_objective_y} for AVEm-DPO results in the best performance of the model across all tasks. We also perform experiments using a single rejected response (\cref{eq:dpo_tpd_objective}), and we can see that using $y_l^{er}$ and $y_l^{vr}$ individually results in improvement over the base, specifically for the \emph{Spurious Cue-Emotion Association} and \emph{Emotion-relevant Cue Hallucination} subtasks, respectively. Moreover, similar to Vista-DPO \citep{huang2025mathcalvistadpo}, we perform an experiment using $y_l^{irr}$ as the second rejected response, which results in the same or worse performance than using $y_l^{vr}$ and $y_l^{er}$ alone. When using $y_l^{irr}$ as the second rejected response, we set $\beta_{irr} = 0.3$ following \cite{huang2025mathcalvistadpo}.

\begin{figure}[t]
  \centering
  \begin{minipage}{0.33\linewidth}
    \centering
    \includegraphics[width=\linewidth,trim=1mm 5mm 3mm 2mm, clip]{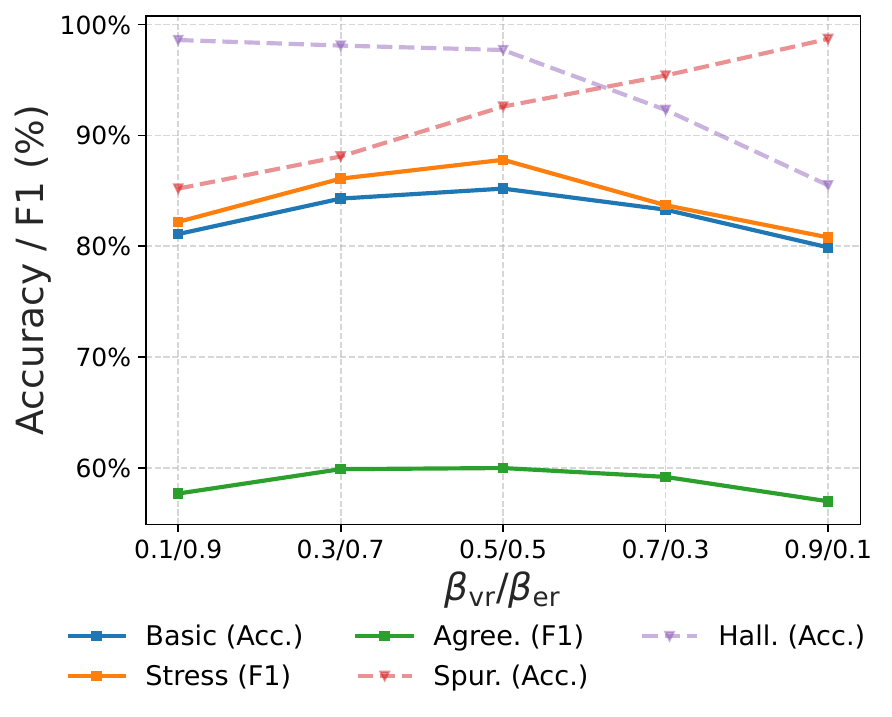}
  \end{minipage}\hfill
  \begin{minipage}{0.33\linewidth}
    \centering
    \includegraphics[width=\linewidth,trim=2mm 5mm 2mm 2mm, clip]{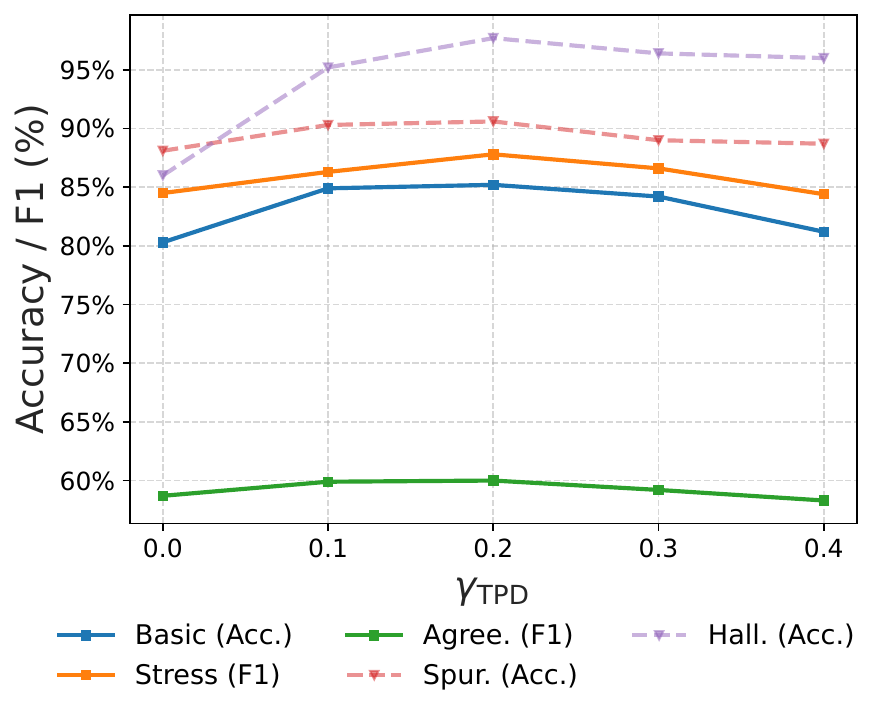}
  \end{minipage}\hfill
  \begin{minipage}{0.33\linewidth}
    \centering
    \includegraphics[width=\linewidth,trim=2mm 5mm 2mm 2mm, clip]{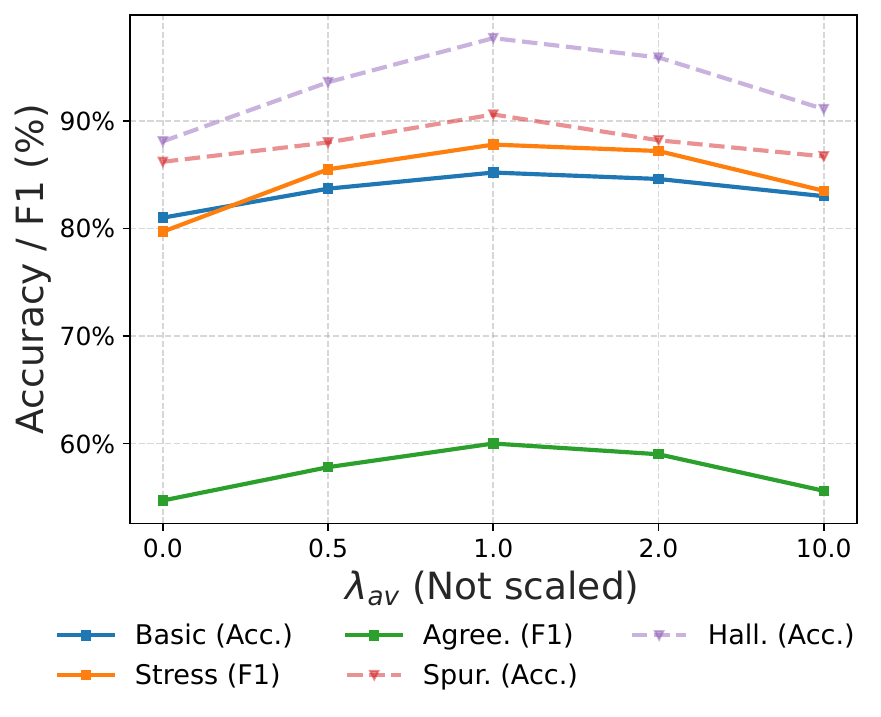}
  \end{minipage}
  \caption{Sensitivity of performance to $\beta_{vr}/\beta_{er}$,  $\gamma_{\text{TPD}}$ and $\lambda_{av}$.}
  \label{fig:hyperparameter_ablation}
\end{figure}

\subsection{Sensitivity to hyperparameters}
\label{subsec:appendix_sensitivity_to_hyperparams}
\cref{fig:hyperparameter_ablation} shows AVEm-DPO's accuracy on different subtasks of \emph{EmoReAlM} on varying the hyperparameters $\beta_{vr}/\beta_{er}$ in \cref{eq:dpo_response_preference}. We can observe that while spurious cue-emotion associations mitigate on increasing $\beta_{vr}$, model performance on hallucinated cue samples improves on increasing $\beta_{er}$. For text-prior debiasing (TPD), we can see that performance on hallucinated cue samples significantly improves even with $\gamma_\text{TPD}=0.1$ and gets saturated at $\gamma_\text{TPD}>0.2$. Finally, increasing the strength of PMP using $\lambda_{av}$ (\cref{eq:final_avemdpo_objective}) improves performance but it gets saturated at $\lambda_{av}>1.0$.

\begin{figure}[]
    \centering
    \includegraphics[width=\linewidth]{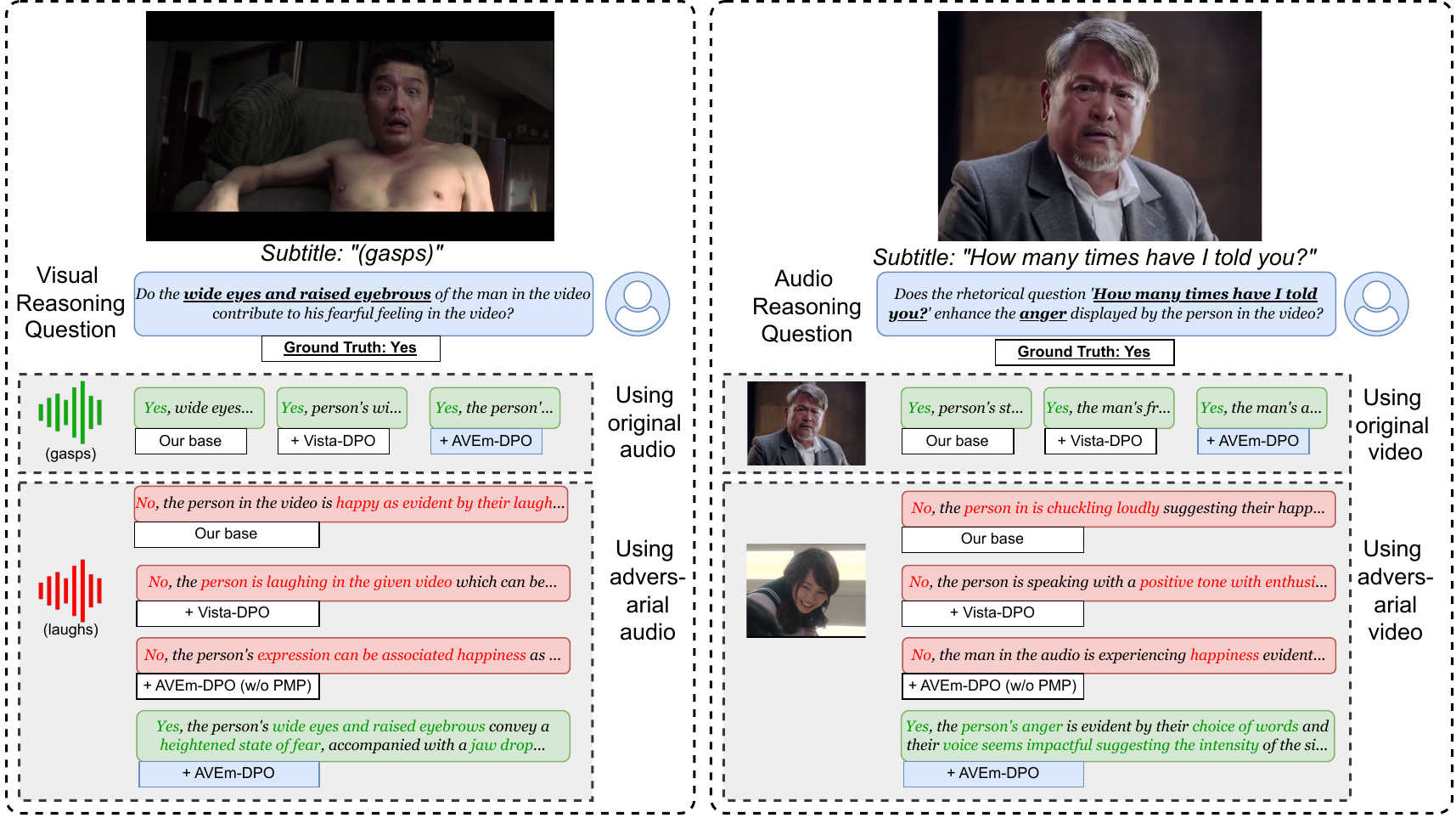}
    \caption{\emph{(Left)} For a visual reasoning question, we compare the model responses on using the original video with the original audio and an adversarial audio as input. We can observe that Vista-DPO and even AVEm-DPO without prompt-based modality preference (PMP) struggle in the adversarial settings; however, AVEm-DPO produces the desired response. \emph{(Right)} We perform a similar experiment to show the visual reasoning robustness of AVEm-DPO.}
    \label{fig:avem_dpo_adversarial_modality}
\end{figure}

\begin{figure}[t]
  \centering
  \begin{minipage}{0.24\linewidth}
    \centering
    \includegraphics[width=\linewidth]{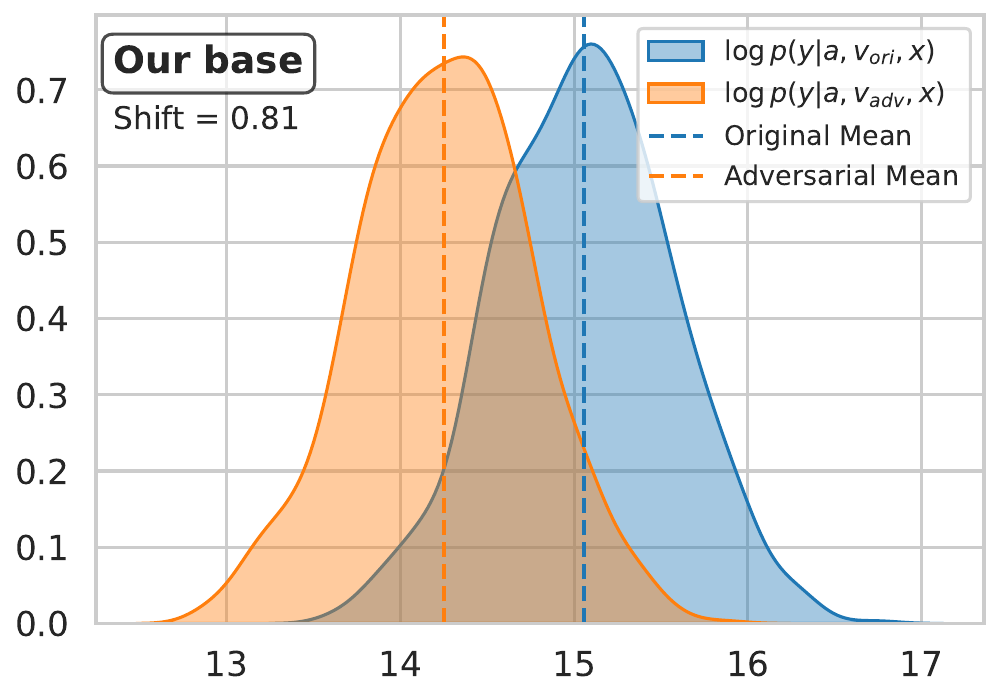}
  \end{minipage}\hfill
  \begin{minipage}{0.24\linewidth}
    \centering
    \includegraphics[width=\linewidth]{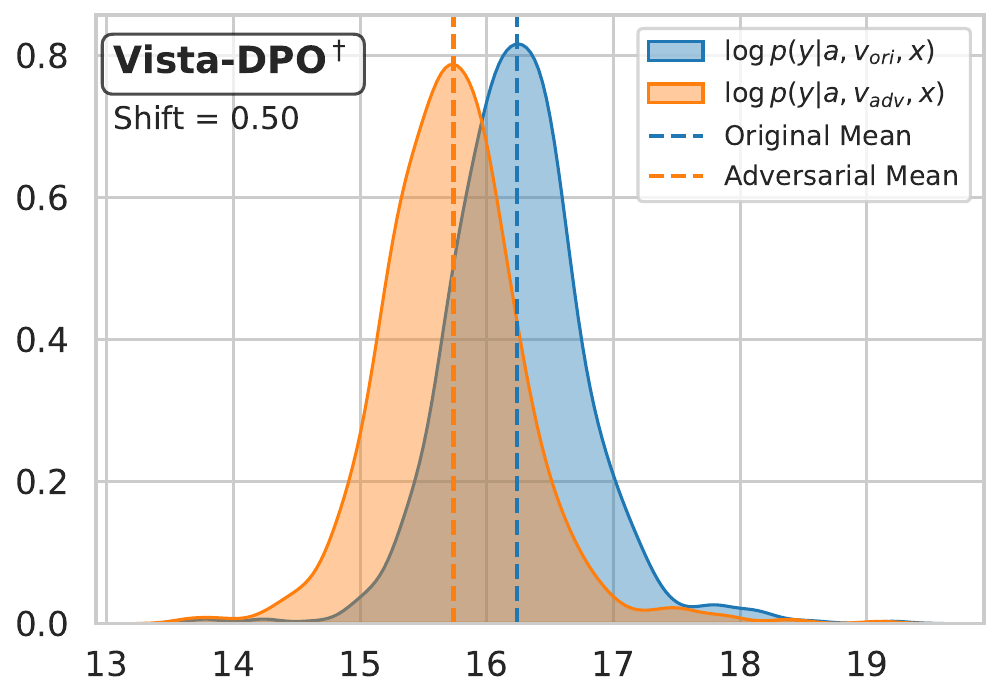}
  \end{minipage}\hfill
  \begin{minipage}{0.24\linewidth}
    \centering
    \includegraphics[width=\linewidth]{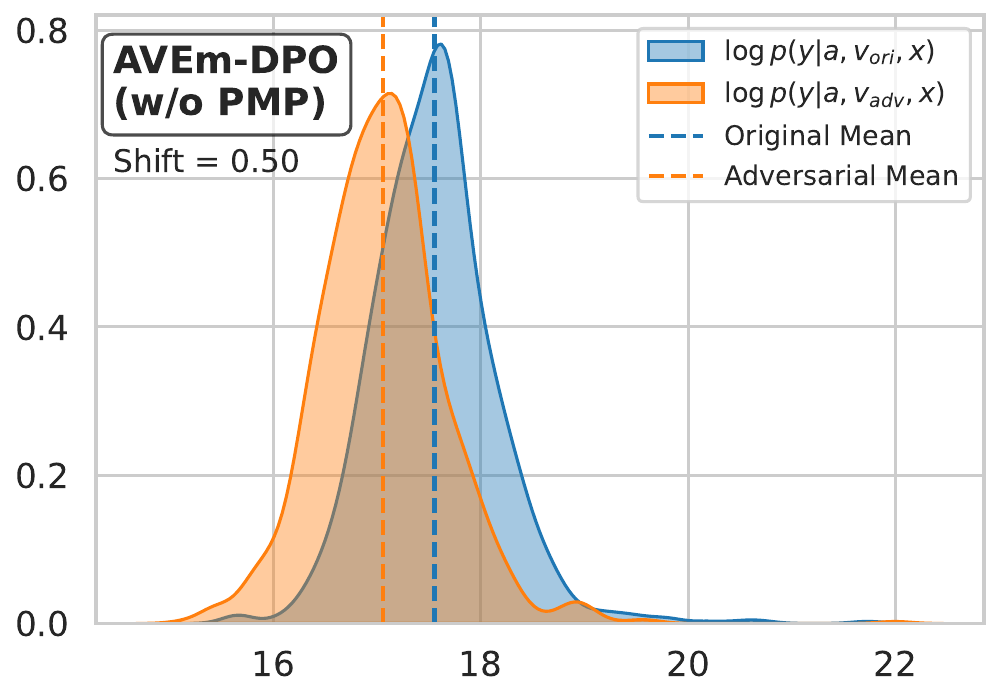}
  \end{minipage}\hfill
  \begin{minipage}{0.24\linewidth}
    \centering
    \includegraphics[width=\linewidth]{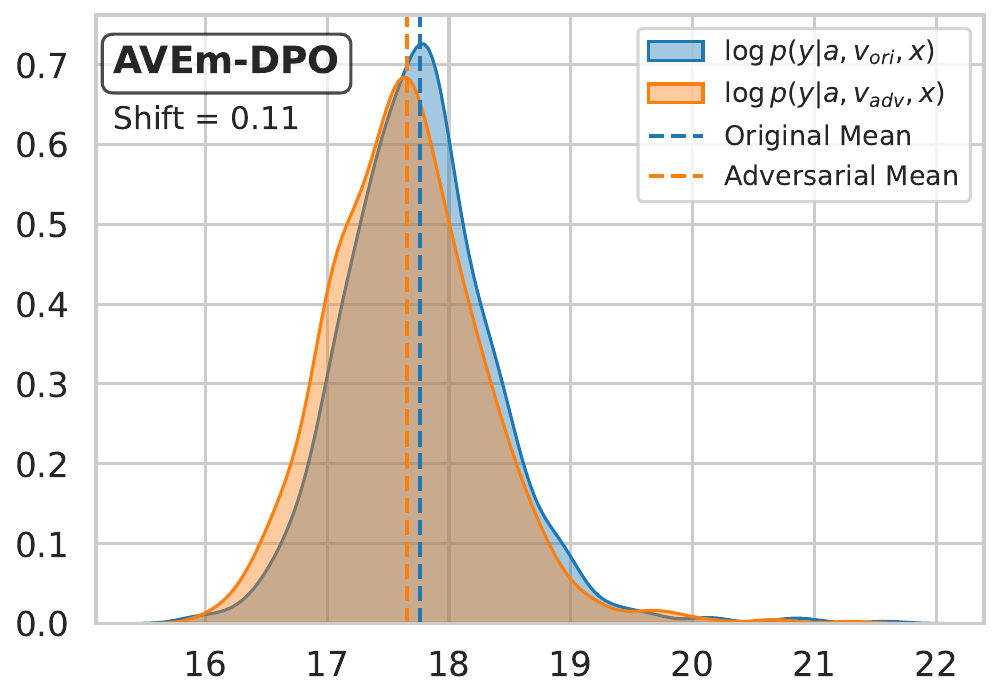}
  \end{minipage}
  \caption{\textbf{Adversarial Audio Reasoning Testing}. For samples related to \emph{audio} reasoning in the EmoReAlM benchmark (\emph{Emotion Reasoning-Basic} and \emph{Emotion Reasoning - Stress Test}), we plot the Kernel Density Estimation (KDE) shift in log likelihoods of the correct answer when the irrelevant video modality input ($v_{ori}$) is replaced with a random video as adversary ($v_{adv}$). AVEm-DPO is least affected by the addition of an adversary in the irrelevant modality (i.e., video). }
  \label{fig:adversarial_audio_modality_testing}
\end{figure}

\begin{figure}[!h]
  \centering
  \begin{minipage}{0.24\linewidth}
    \centering
    \includegraphics[width=\linewidth]{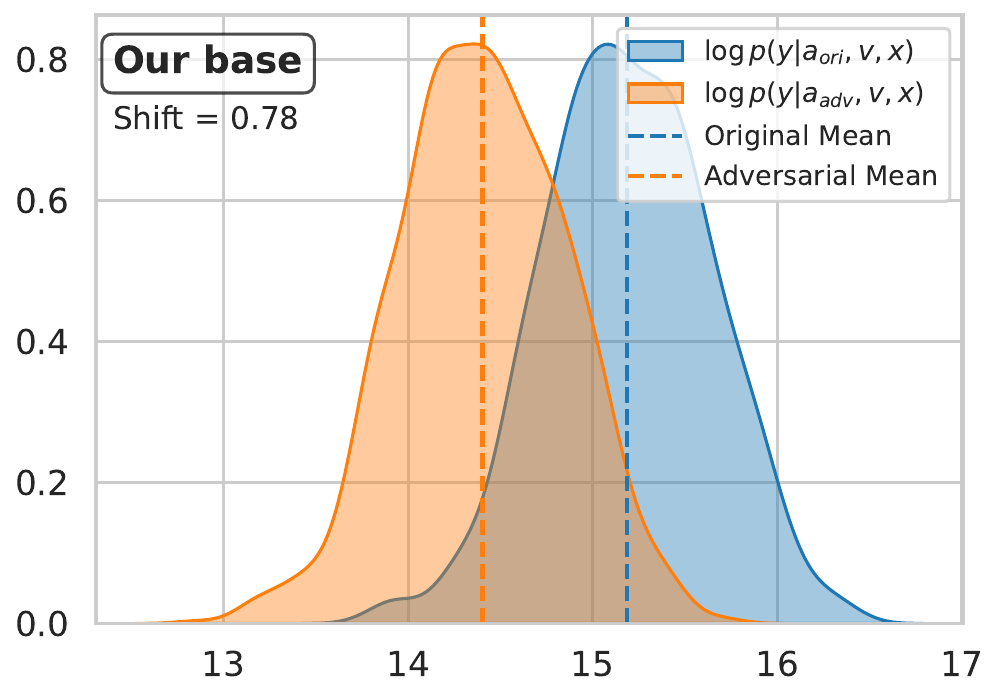}
  \end{minipage}\hfill
  \begin{minipage}{0.24\linewidth}
    \centering
    \includegraphics[width=\linewidth]{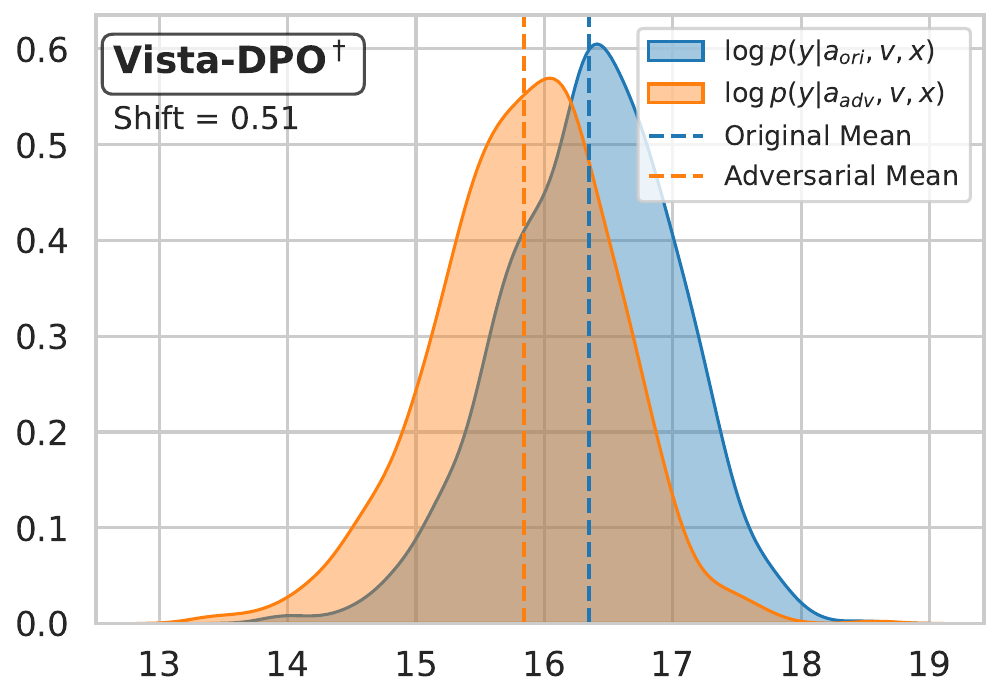}
  \end{minipage}\hfill
  \begin{minipage}{0.24\linewidth}
    \centering
    \includegraphics[width=\linewidth]{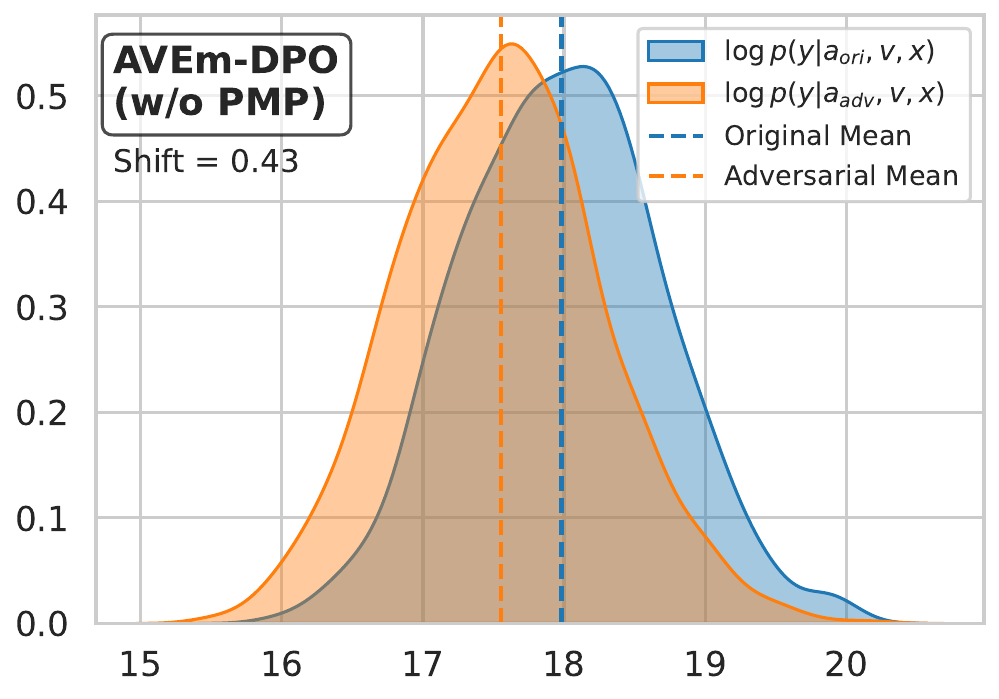}
  \end{minipage}\hfill
  \begin{minipage}{0.24\linewidth}
    \centering
    \includegraphics[width=\linewidth]{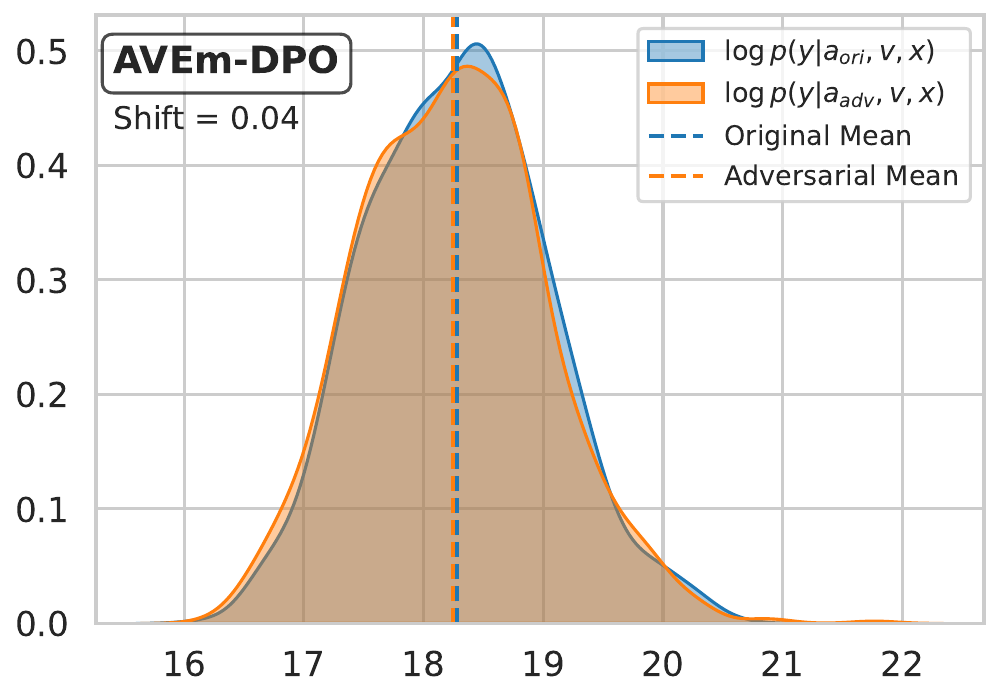}
  \end{minipage}
  \caption{\textbf{Adversarial Visual Reasoning Testing}. Similar to \cref{fig:adversarial_audio_modality_testing}, for samples related to \emph{visual} reasoning in the EmoReAlM benchmark (\emph{Emotion Reasoning-Basic} and \emph{Emotion Reasoning - Stress Test}), we plot the Kernel Density Estimation (KDE) shift in log likelihoods of the correct answer when the irrelevant audio modality input ($a_{ori}$) is replaced with a random video as adversary ($a_{adv}$). AVEm-DPO is least affected by the addition of an adversary in the irrelevant modality (i.e., audio). }
  \label{fig:adversarial_visual_modality_testing}
\end{figure}

\begin{figure}[t]
  \centering
  \begin{minipage}{0.25\linewidth}
    \centering
    \includegraphics[width=\linewidth]{figures/attn_box_plots/av_attn_box_audio.pdf}
  \end{minipage}\hfill
  \begin{minipage}{0.25\linewidth}
    \centering
    \includegraphics[width=\linewidth]{figures/attn_box_plots/av_attn_box_visual.pdf}
  \end{minipage}\hfill
  \begin{minipage}{0.25\linewidth}
    \centering
    \includegraphics[width=\linewidth]{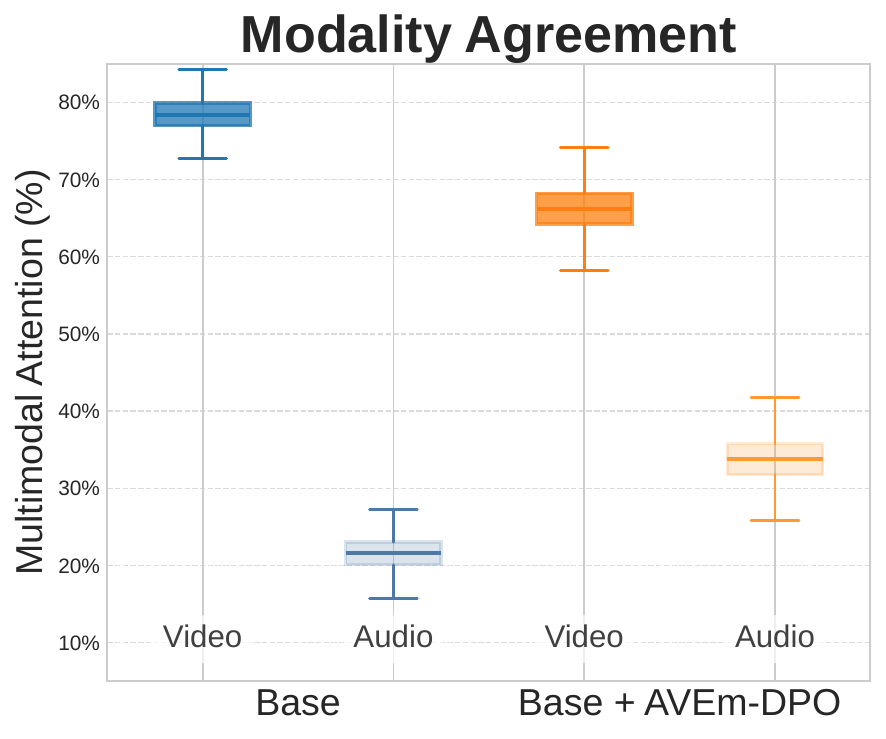} 
  \end{minipage}\hfill
  \begin{minipage}{0.25\linewidth}
    \centering
    \includegraphics[width=\linewidth]{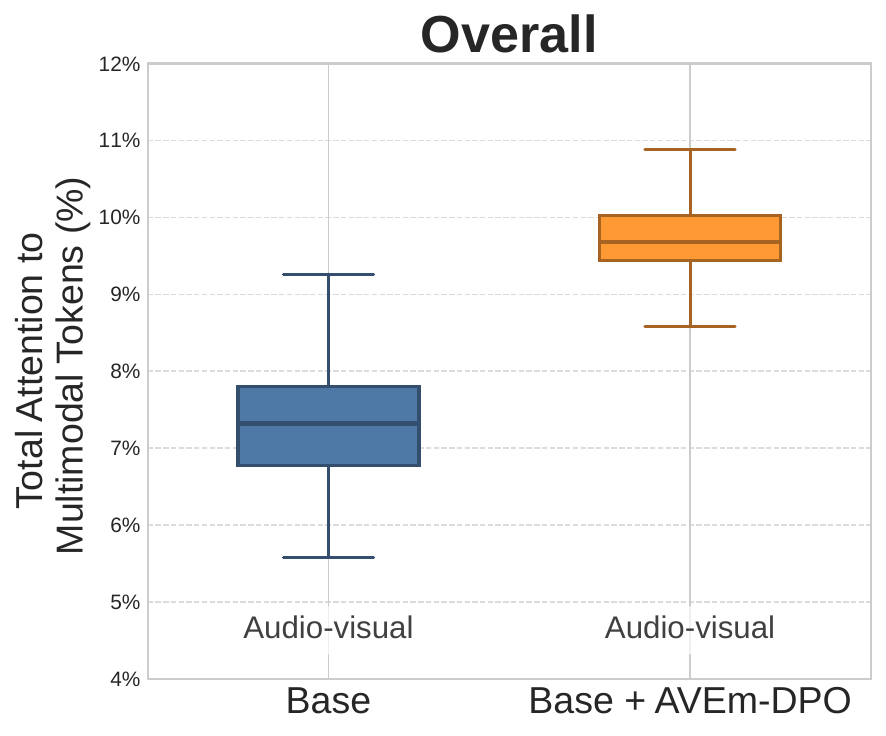} 
  \end{minipage}
  \vspace{-1em}
  \caption{Effect of AVEm-DPO on the distribution of attention over --  (i) \emph{(Left three plots)} video and audio tokens taken as a percentage over the total attention over all multimodal tokens for different subtasks in \emph{EmoReAlM} and (ii) \emph{(Right)} multimodal tokens as a percentage over the total input tokens (including text) for the entire \emph{EmoReAlM}.}
  \label{fig:appendix_attention_analysis}
  \vspace{-1em}
\end{figure}

\subsection{Attention Redistribution after Preference Optimization.}
\label{subsec:appendix_attn_redistribution_DPO}
As described in \cref{subsec:results_analysis}, to analyze the effect of preference optimization on attention, we plot the distribution of aggregate multimodal input attention over audio and visual tokens averaged over all attention heads for different tasks in \emph{EmoReAlM} in \cref{fig:appendix_attention_analysis} (\emph{left two plots}). For reasoning tasks, we can observe that the attention over relevant modality increases after AVEm-DPO. For the \emph{Modality Agreement} task, the attention is redistributed in a way that there is a fair distribution of attention between both modalities to ensure reliable predictions. 

To show the effect of text-prior debiasing \cref{subsec:text_prior_debiasing}, we plot the percentage of total attention (averaged over attention heads) over multimodal input tokens (audio and visual combined) and observe that AVEm-DPO increases the attention over multimodal tokens by significant margins (\cref{fig:appendix_attention_analysis} -- \emph{right}). This shows that AVEm-DPO training ensures that the model attends to the relevant audiovisual tokens for generating the response rather than relying only on the input text prompt.

\subsection{Reasoning with Adversarial Modality Inputs}
\label{subsec:appendix_reasoning_adversarial_modality}
To test the robustness of AVEm-DPO against cross-modality hallucinations, we conduct an adversarial test by replacing the audio in a visual reasoning task to see if the model's response stays the same. As shown in \cref{fig:avem_dpo_adversarial_modality}, changing the prompt-irrelevant modality does not change the response of AVEm-DPO, showing its adversarial robustness. It is interesting to note that removing the prompt-based modality preference (PMP) from AVEm-DPO results in wrong predictions, showing its efficacy. To quantitatively show the effect of AVEm-DPO with PMP, we perform adversarial testing using \emph{Emotion Reasoning-Basic} and \emph{Emotion Reasoning - Stress Test} samples in \emph{EmoReAlM}. For testing the robustness of AVEm-DPO for audio reasoning (\cref{fig:adversarial_audio_modality_testing}), we compute the shift of log likelihoods of the correct response when the prompt-irrelevant video modality is replaced with an adversary (i.e., some random video). We use Kernel Density Estimation (KDE) to estimate the shift in the distributions. We can see that AVEm-DPO is robust to adversaries in the prompt-irrelevant modality. Moreover, removing PMP from AVEm-DPO significantly increases the shift between the original and adversarial distributions. \cref{fig:adversarial_visual_modality_testing} shows similar plots for the tasks related to visual reasoning. 

\begin{table}[]
\centering
\caption{Performance of different models for emotion prediction using audiovisual, video-only, and audio-only inputs from the RAVDESS \citep{Livingstone2018_ravdess} dataset.}
\vspace{-1em}
\label{tab:ravdess_audio_video_only}
\resizebox{0.6\columnwidth}{!}{%
\begin{tabular}{l|cc|cc|cc}
\hline\hline
\rowcolor[HTML]{C0C0C0} 
\cellcolor[HTML]{C0C0C0} & \multicolumn{2}{c|}{\cellcolor[HTML]{C0C0C0}\textbf{Audiovisual}} & \multicolumn{2}{c|}{\cellcolor[HTML]{C0C0C0}\textbf{Video-only}} & \multicolumn{2}{c}{\cellcolor[HTML]{C0C0C0}\textbf{Audio-only}} \\ \cline{2-7} 
\rowcolor[HTML]{C0C0C0} 
\multirow{-2}{*}{\cellcolor[HTML]{C0C0C0}\textbf{Model}} & \textbf{UAR} & \textbf{WAR} & \textbf{UAR} & \textbf{WAR} & \textbf{UAR} & \textbf{WAR} \\ \hline
VideoLLaMA 2 & 41.81 & 31.62 & 36.12 & 32.41 & 30.44 & 27.56 \\
Qwen 2.5 Omni & 32.88 & 28.05 & 29.38 & 27.67 & 28.56 & 25.55 \\ \hline
Our base & 53.59 & 53.01 & 41.27 & 40.98 & 38.18 & 37.74 \\
\rowcolor[HTML]{DAE8FC}
+ AVEm-DPO & 58.66 & 55.48 & 46.13 & 46.31 & 44.05 & 39.67 \\ \hline
EmotionLLaMA$^\star$ & 52.59 & 48.12 & 41.27 & 39.56 & 38.57 & 37.27 \\
\rowcolor[HTML]{DAE8FC}
+ AVEm-DPO & 56.21 & 51.03 & 46.10 & 43.09 & 40.54 & 37.04 \\ \hline\hline
\end{tabular}%
}
\end{table}

\subsection{Effect of Individual Modalities for Emotion Prediction}
\label{subsec:individual_modality_effect}

To show the effect of individual modalities for emotion recognition, \cref{tab:ravdess_audio_video_only} reports the performance on using only the video, only the audio and using audiovisual inputs from the RAVDESS \citep{Livingstone2018_ravdess} dataset for emotion prediction. We can observe that using the individual modalities for emotion prediction leads to a reduced performance, indicating that using the audiovisual inputs for emotion prediction is indeed helpful compared to using a single modality. Moreover, we can observe that the performance using only the visual modality is better compared to using only audio, indicating the importance of visual modality for emotion prediction. 

\clearpage

\begin{figure}
    \centering
    \includegraphics[width=\linewidth]{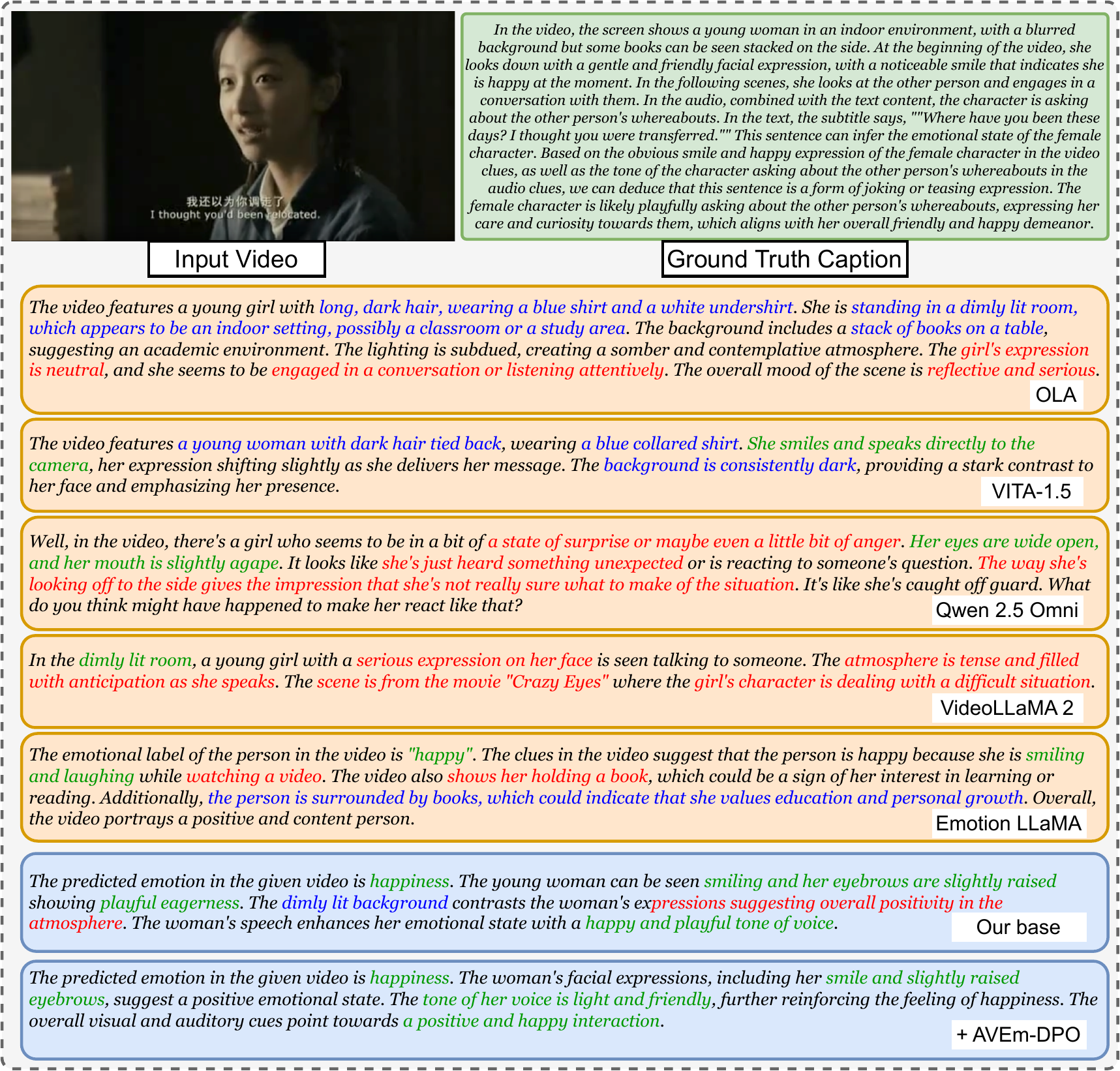}
    \caption{Comparison of baseline MLLMs with our base model trained with AVEm-DPO on a sample from EMER \citep{lian2023explainable_emer}. Correct audiovisual cues and emotion are in \textcolor{mygreen}{green}, emotion-irrelevant cues are in \textcolor{blue}{blue}, and hallucinated cues (and incorrect emotion) are present in \textcolor{red}{red}}.
    \label{fig:appendix_emer_qualitative_1}
\end{figure}

\begin{figure}
    \centering
    \includegraphics[width=\linewidth]{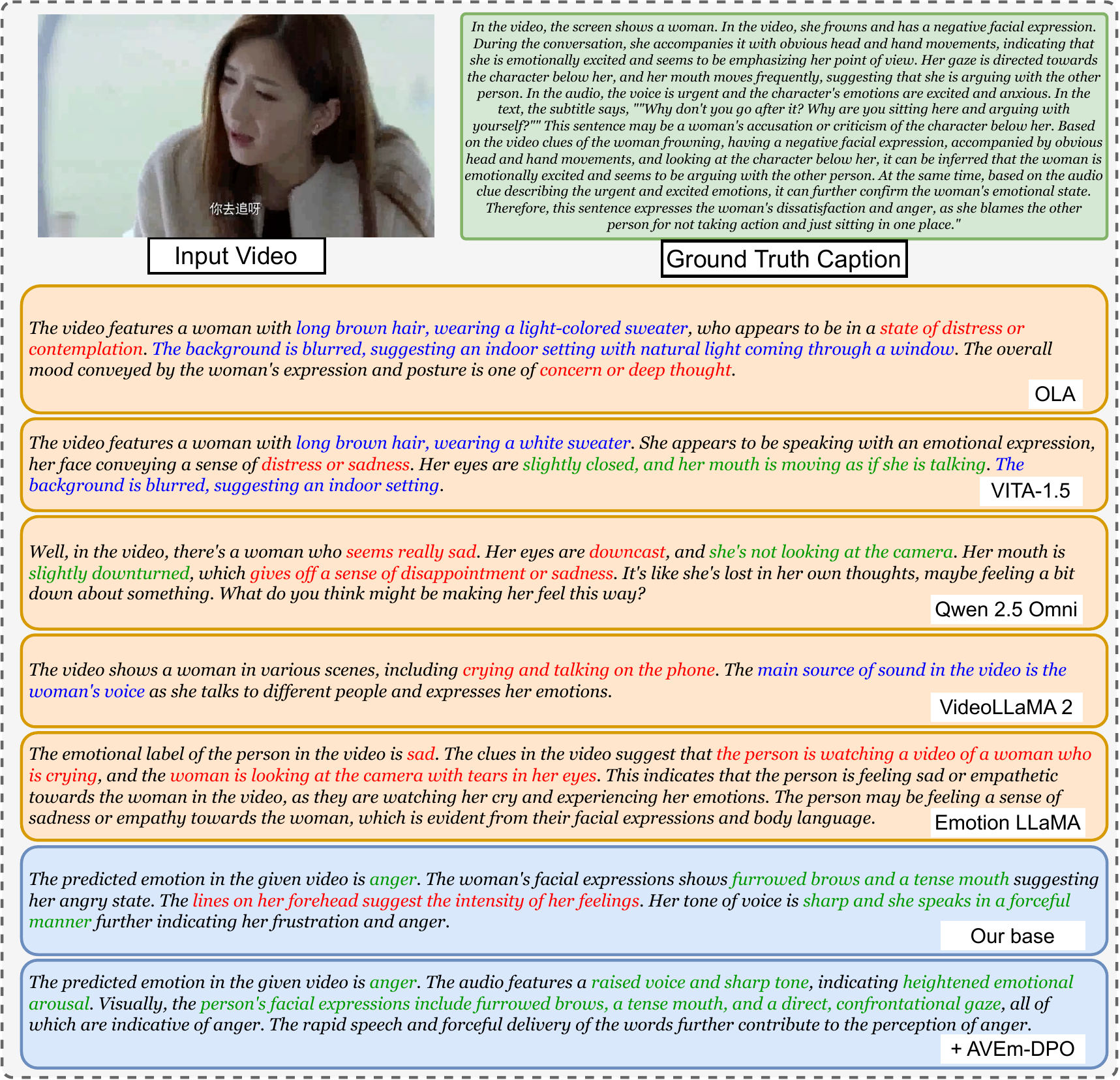}
    \caption{Comparison of baseline MLLMs with our base model trained with AVEm-DPO on a sample from EMER \citep{lian2023explainable_emer}. Correct audiovisual cues and emotion are in \textcolor{mygreen}{green}, emotion-irrelevant cues are in \textcolor{blue}{blue}, and hallucinated cues (and incorrect emotion) are present in \textcolor{red}{red}}.
    \label{fig:appendix_emer_qualitative_2}
\end{figure}

\begin{figure}
    \centering
    \includegraphics[width=\linewidth]{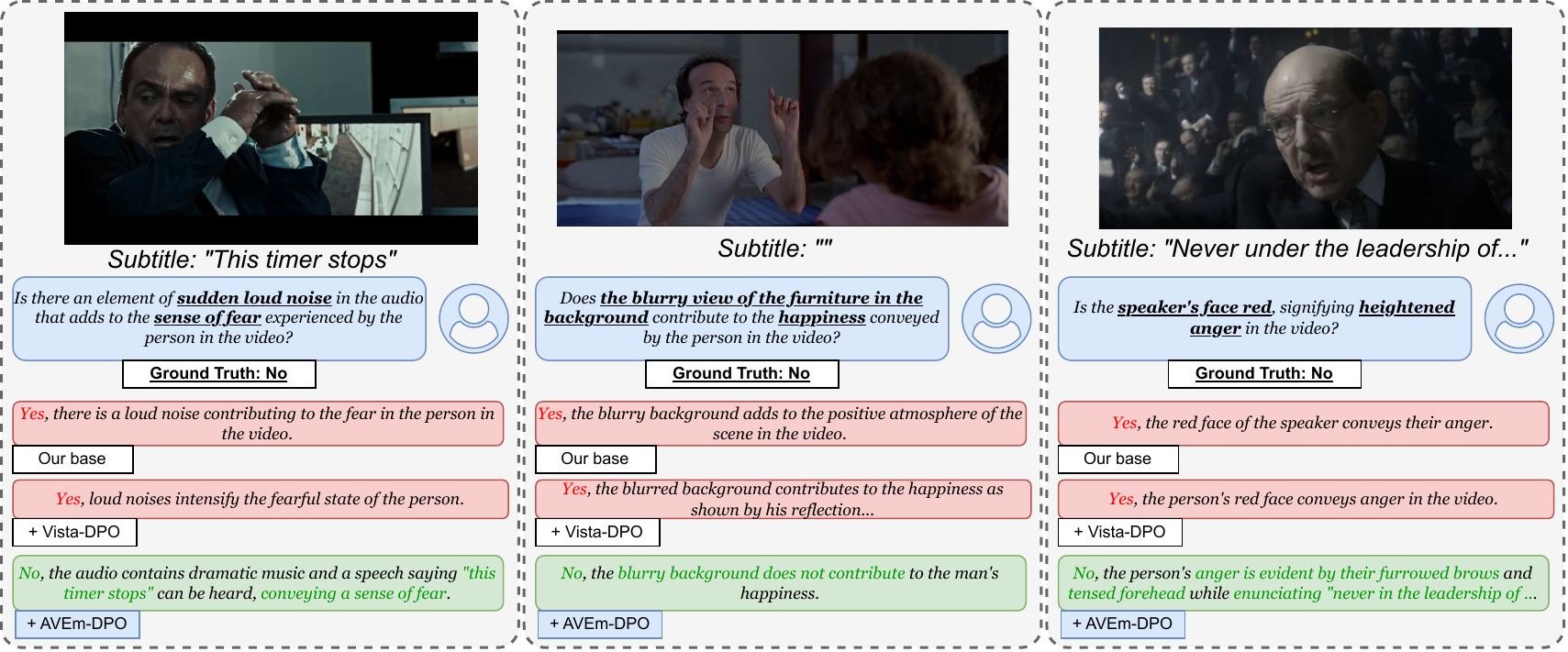}
    \caption{Qualitative examples comparing the output responses using different approaches for some samples present in the \emph{Emotion Reasoning - Stress Test} of EmoReAlM benchmark.}
    \label{fig:qualitative_samples_avem_dpo}
\end{figure}


\section{Qualitative Samples}
\label{sec:appendix_qualitative_samples}

\paragraph{Emotion Descriptions on EMER.} \cref{fig:appendix_emer_qualitative_1,fig:appendix_emer_qualitative_2} shows samples from the EMER \citep{lian2023explainable_emer} dataset and the output of different MLLM baselines on those samples using the prompt -- \emph{``Describe the audiovisual content relevant to emotion in detail."}. We can see that AVEM-DPO leads to correct emotion descriptions and consistent audiovisual cues to reason for the emotions. Moreover, compared to baselines, our method does not associate irrelevant and/or background information with emotions. 

\paragraph{EmoReAlM Sample Outputs.} \cref{fig:qualitative_samples_avem_dpo} shows the model responses for some samples in the \emph{Emotion Reasoning - Stress Test} of \emph{EmoReAlM} Benchmark. We can notice that AVEm-DPO improves the model responses in cases with spurious-emotion cue associations and emotion-cue hallucinations.

\clearpage
\section{Prompt Pool}
\label{sec:appendix_prompt_pool}

\end{document}